\documentclass[runningheads]{llncs}

\usepackage{eccv}

\usepackage{eccvabbrv}

\usepackage{graphicx}
\usepackage{booktabs}
\usepackage{xfrac}
\usepackage{adjustbox}
\usepackage{multirow}
\usepackage{subcaption}

\usepackage[accsupp]{axessibility}  

\usepackage{hyperref}

\usepackage{orcidlink}

\begin{document}

\title{Neural Graphics Texture Compression Supporting Random Access} 

\titlerunning{Neural Graphics Texture Compression Supporting Random Access}


\author{
Farzad Farhadzadeh$^{*}$\inst{1}\orcidlink{0000-0002-3969-2089}  \qquad
Qiqi Hou\inst{1}\orcidlink{0009-0009-3472-6401} \qquad
Hoang Le\inst{1}\orcidlink{0000-0001-9170-9581} \qquad
Amir Said \inst{1}\orcidlink{0000-0002-1802-3513} \qquad
Randall Rauwendaal\inst{2}\orcidlink{4444-5555-6666-7777} \qquad 
Alex Bourd\inst{2}\orcidlink{4444-5555-6666-7777} \qquad
Fatih Porikli\inst{1}\orcidlink{0000-0002-1520-4466}
}

\authorrunning{F.~Farhadzadeh et al.}

\institute{Qualcomm AI Research$^{\dagger}$, San Diego, CA 92121, USA \\
\email{\{ffarhadz,qhou,hoanle,asaid,fporikli\}@qti.qualcomm.com} 
\and
Qualcomm Technologies, Inc., San Diego, CA 92121, USA \\
\email{\{abourd,rrauwend\}@qti.qualcomm.com}}

\maketitle

\renewcommand{\thefootnote}{\fnsymbol{footnote}}
\footnotetext[1]{\hspace{-0.01in}Corresponding author} 
\footnotetext[4]{\hspace{-0.01in}Qualcomm AI Research is an initiative of Qualcomm Technologies, Inc.} 
\renewcommand*{\thefootnote}{\arabic{footnote}}
\setcounter{footnote}{0}

\begin{abstract}
Advances in rendering have led to tremendous growth in texture assets, including resolution, complexity, and novel textures components, but 
this growth in data volume has not been matched by advances in its compression. 
Meanwhile Neural Image Compression (NIC) has advanced significantly and shown promising results, but
the proposed methods cannot be directly adapted to neural texture compression.
First, texture compression requires on-demand and real-time decoding with random access during parallel rendering (e.g. block texture decompression on GPUs).  
Additionally, NIC does not support multi-resolution reconstruction (mip-levels), nor does it have the ability to efficiently jointly compress different sets of texture channels. 
In this work, we introduce a novel approach to texture set compression that integrates traditional GPU texture representation and NIC techniques, designed to enable random access and support many-channel texture sets. 
To achieve this goal, we propose an asymmetric auto-encoder framework that employs a convolutional encoder to capture detailed information in a bottleneck-latent space, and at decoder side we utilize a fully connected network, whose inputs are sampled latent features plus positional information, for a given texture coordinate and mip level. 
This latent data is defined to enable simplified access to multi-resolution data by simply changing the scanning strides.
Experimental results demonstrate that this approach provides much better results than conventional texture compression, and significant improvement over the latest method using neural networks.
\end{abstract}

\section{Introduction}
\label{sec:intro}
\begin{figure*}
\centering
     \begin{subfigure}[b]{0.72\textwidth}
         \centering
         \includegraphics[width=1\textwidth]{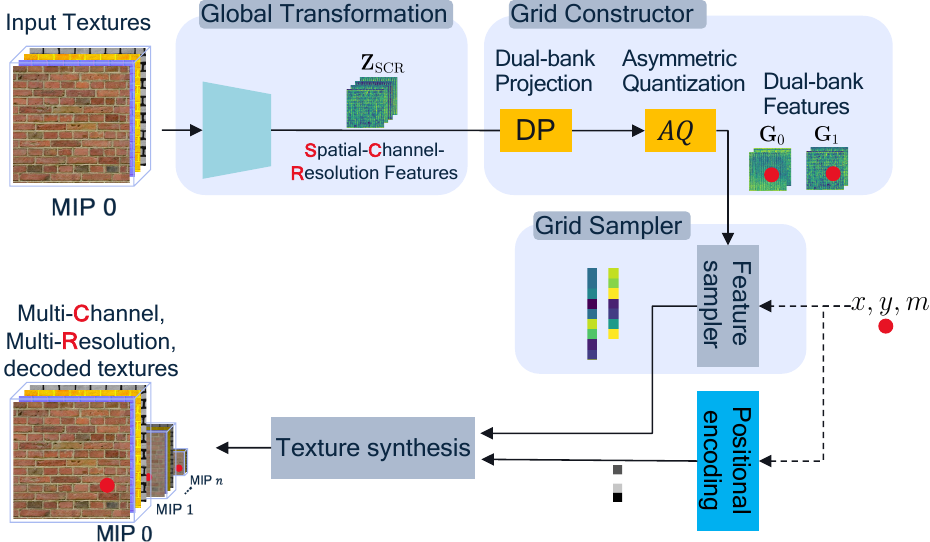}
         \caption{}
         \label{fig:main_arch_0}
     \end{subfigure}
     \hspace{-0.0cm}
     \begin{subfigure}[b]{0.25\textwidth}
         \centering
         \includegraphics[width=1\textwidth]{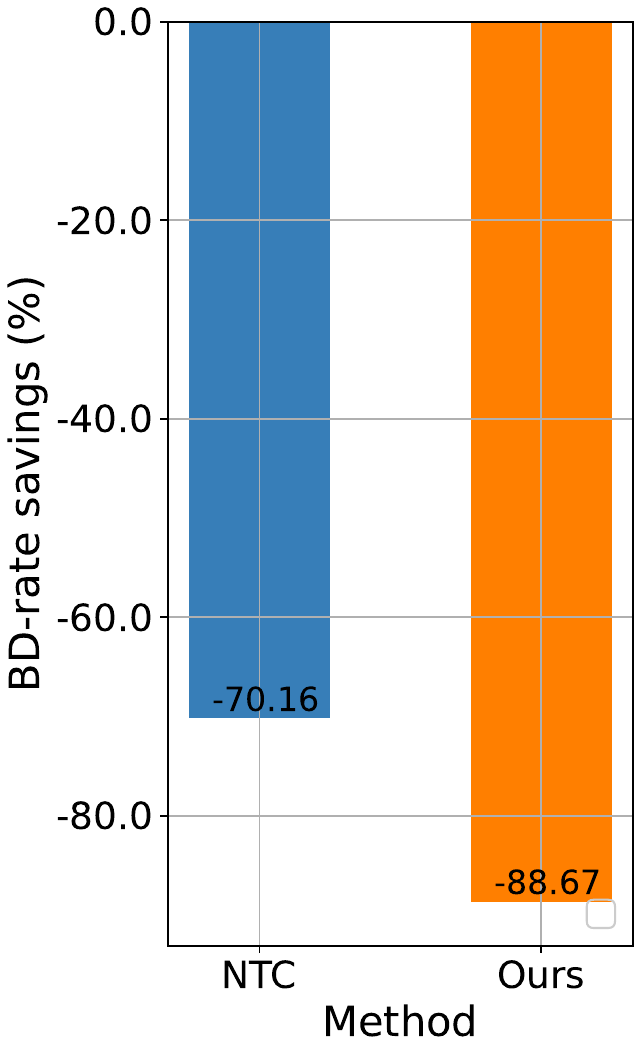}
         \caption{}
         \label{fig:bd_bar}
     \end{subfigure}
     \vspace{-0.2cm}
        \caption{(a) Overview of our neural texture compression (b) Bitrate comparison over ASTC in terms of PSNR.  Our method significantly outperforms the state-of-the-art texture compression method, NTC~\cite{Vaidyanathan2023}, with a 40.8\% in the BD-rate saving.}
        \label{fig:archs}
    \vspace{-0.3cm}
\end{figure*}

The rapid growth of graphics and gaming industries hinges on new technologies that create increasingly photorealistic scenes, a goal that heavily relies on texturing. Textures are collections of 2D arrays with information about how a 3D object’s surface appearance should change according to lighting and viewing position~\cite{shirley2009fund,hughes2013intro}. Texturing endows surfaces with realistic characteristics like roughness, smoothness, reflectivity, and intricate patterns, allowing faithful replication of various materials such as wood, metal, fabric, glass, or concrete. This significantly contributes to the overall realism of rendered scenes. However, achieving this level of realism necessitates vast amounts of data which can impact download times, rendering speed, and local storage. While the amount of data has been reduced with some widely used texture compression methods, these are comparatively limited and inefficient~\cite{Vaidyanathan2023}. The development of better compression methods has been very slow due to several technical difficulties.

In computer graphics, textures often appear magnified and can have large sizes with multiple components. These are organized in a multi-resolution mipmap pyramid, which plays a crucial role in texture resampling. Mipmaps can be visualized as an image pyramid for efficient texture filtering, where each mip level represents a filtered version of the texture, corresponding to a specific image pixel-to-texture pixel ratio. However, only a small fraction of this data is needed for a specific scene view, and therefore efficient texture compression methods are essential to decompress only the required components. This compression must also support asynchronous parallelized execution, allowing each rendering thread to independently access the necessary values for its tasks at the required position and resolution. 

This requirement eliminates several of the most efficient techniques used for compressing 2D images and videos, that exploit wide scale dependencies, statistical consistency, and serialized entropy coding, and that consequently cannot be independently decompressed. 

Another complication is that rendering is fully programmable, enabling the creation of an increasing variety of texture components. Thus, the design of methods that can optimally compress all new texture types cannot be done effectively manually, but can be automated via learning-based techniques.

Neural network-based compression methods have been proposed and demonstrated to be competitive with conventional methods for images and video. However, these codecs are not designed for, nor can they be easily modified to support random access and other rendering requirements.
Unlike digital image compression, 
which focuses on compressing an entire image at a fixed resolution and typically handles a fixed number of color channels (e.g., 3 channels for RGB or YUV), 
texture compression must exploit multiple levels of redundancy in the texture components. 
This paper addresses these complexities by proposing new learning-based methods and neural-network architectures specifically designed for texture compression and rendering.

\section{Motivation}
\label{sec:motivation}

Conventional texture compression achieves random access by employing block-wise vector quantization. These techniques often leverage both spatial and cross-channel correlations. However, their limitation lies in the fact that they can only compress textures with up to ``four channels''. In contrast, modern renderers commonly utilize a broader range of material properties, including color channels as well as other channels for normal maps, height maps, ambient occlusion, glossiness, roughness, and other information related to the Bidirectional Reflectance Distribution Function (BRDF).

\begin{figure}
\captionsetup[subfigure]{labelformat=empty}
\centering
    \begin{subfigure}[b]{0.15\textwidth}
         \centering         
         \includegraphics[width=1\textwidth]{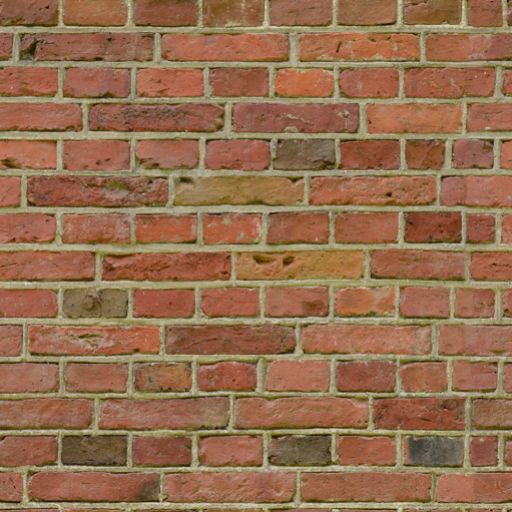}         
         \caption{\scriptsize diffuse}
         \label{}
    \end{subfigure}
    \hspace{0.5cm}
    \begin{subfigure}[b]{0.15\textwidth}
         \centering
         \includegraphics[width=1\textwidth]{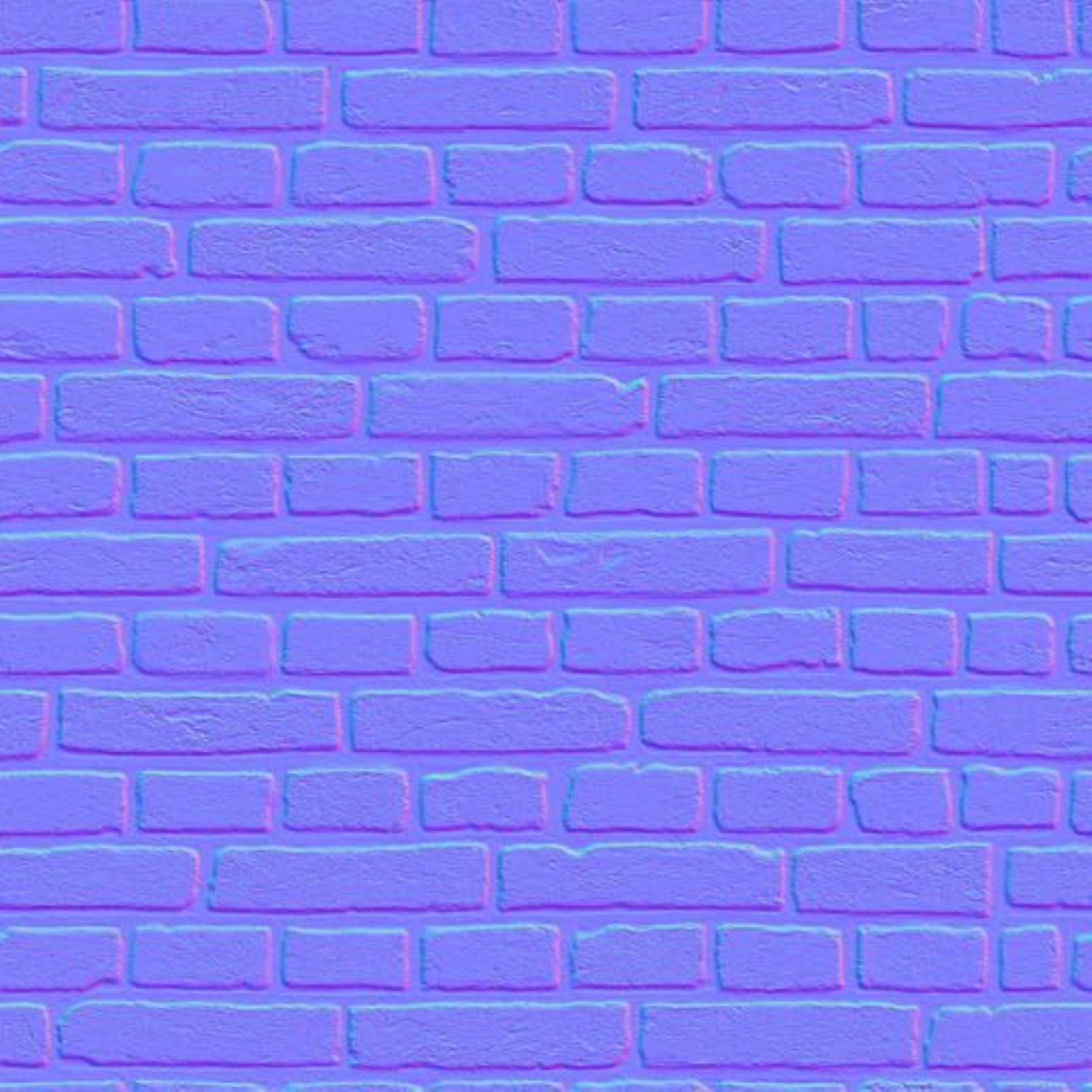}
         \caption{\scriptsize normal}
         \label{}
    \end{subfigure}
    \hspace{0.5cm}
    \begin{subfigure}[b]{0.15\textwidth}
         \centering
         \includegraphics[width=1\textwidth]{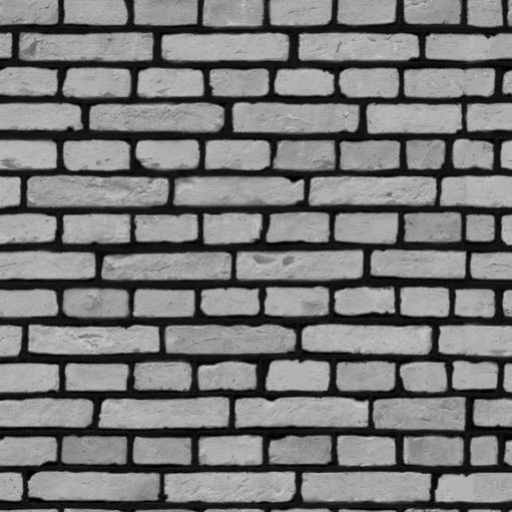}         
         \caption{\scriptsize displacement}
         \label{}
    \end{subfigure}
    \hspace{0.5cm}
    \begin{subfigure}[b]{0.15\textwidth}
         \centering
        \includegraphics[width=1\textwidth]{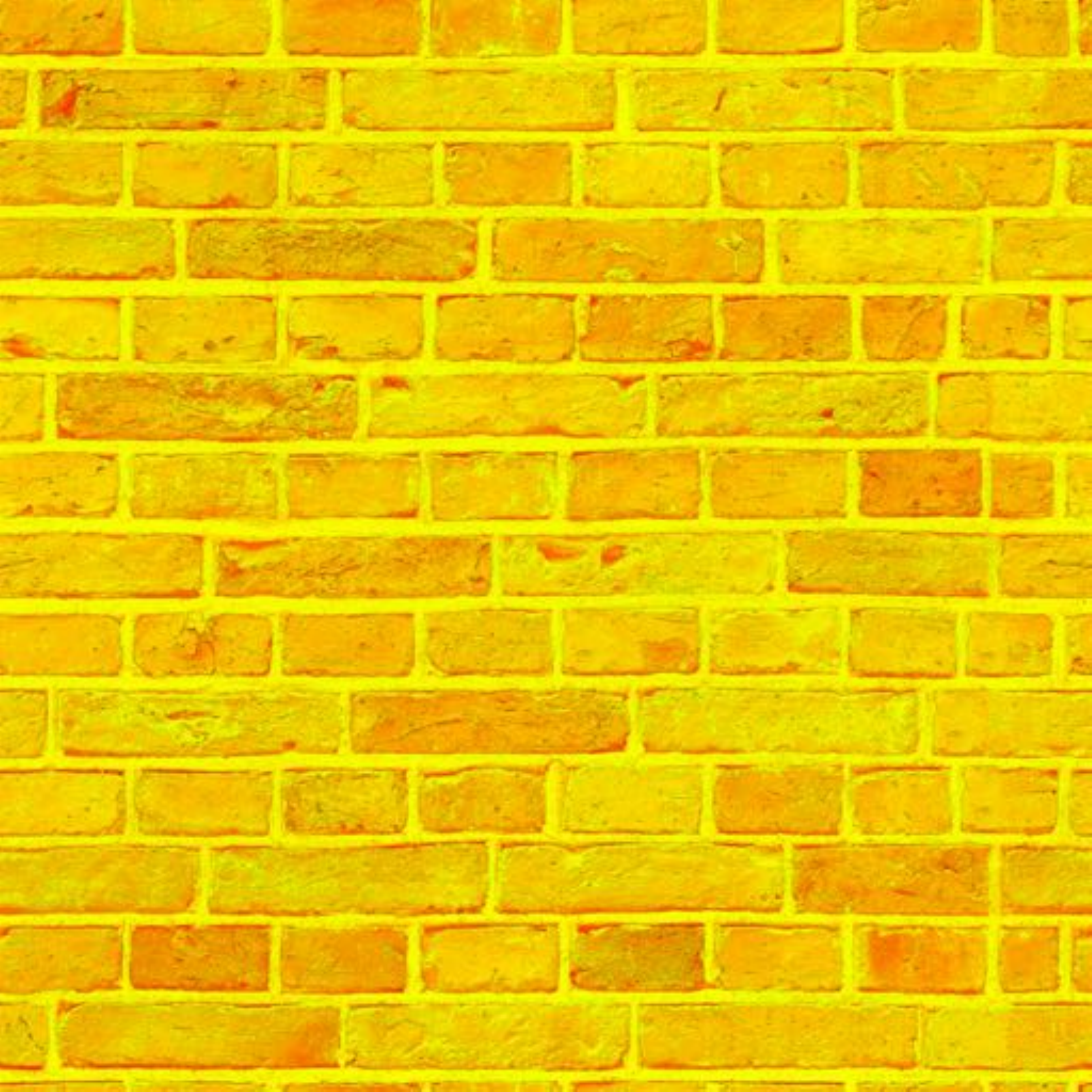}
         \caption{\scriptsize ao \& roughness}
         \label{}
    \end{subfigure}
    \vspace{-0.3cm}
    \caption{An example of texture set consisting of a diffuse map, normal map, displacement map and combined ambient occlusion (ao) and roughness  map for a brick wall pattern retrieved from  \url{https://polyhaven.com}.}
    \label{fig:texture_set_brick}
    \vspace{-0.3cm}
\end{figure}

These channels exhibit significant correlation within the texture set. This correlation arises from a combination of factors, including physical material properties, geometric features, and artistic layering during material authoring~\cite{Hill2020}. For instance, consider Figure~\ref{fig:texture_set_brick}, where a texture set related to a brick wall pattern includes color channels as well as other channels for normal map, combined ambient occlusion (ao) and roughness map, and displacement map. 

Furthermore, in computer graphics, for efficient rendering textures are stored at different resolutions, commonly referred to as mipmaps. 
As shown by~\cite{Bartlomiej2021} there are significant redundancies at different scales. 

In their pioneering work, Vaidyanathan et al.~\cite{Vaidyanathan2023}, introduced a novel neural texture compression scheme. Their approach exploits spatial redundancies across all channels of a texture set and across different mip levels. They achieved this by compressing the entire texture set, including all channels and mip levels simultaneously. Their autodecoder framework~\cite{Park2019-nu} utilizes a pair of feature pyramids, referred to as Girds, to represent the texture set throughout all mip levels. Each pyramid level is responsible for reconstructing texels at one or more mip levels, and features stored at a particular pyramid level can be shared across multiple mip levels. This innovative method significantly advances texture compression efficiency and quality in texture representation.

\begin{figure}[t]
\centering
    \begin{subfigure}[b]{0.2\textwidth}
         \centering         
         \includegraphics[width=1\textwidth]{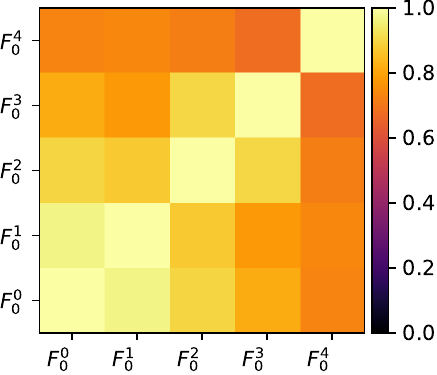}         
         \caption{$\mathrm{CKA}_0$}
         \label{}
    \end{subfigure}
    \hspace{1cm}
    \begin{subfigure}[b]{0.2\textwidth}
         \centering
         \includegraphics[width=1\textwidth]{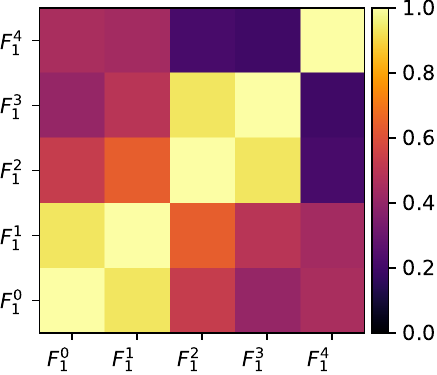}
         \caption{$\mathrm{CKA}_1$}
         \label{}
    \end{subfigure}
    \vspace{-0.2cm}
    \caption{Centered kernel alignment (CKA) of features stored at pyramid levels $k$ and $l$ of the feature pyramid $F_i$, i.e., $F^k_i$ and $F^l_i$, evaluated on “Ceramic roof 01” texture set,
retrieved from \url{https://polyhaven.com/}.}
    \label{fig:cka}
    \vspace{-0.2cm}
\end{figure}

In our analysis, we identified redundancy within the feature pyramids, which adversely affected compression performance. Particularly, as texture resolution increased, this redundancy became more pronounced. 
Figure~\ref{fig:cka} illustrates centered kernel alignment (CKA)~\cite{Kornblith2019-qn} across different levels of the feature pyramid $F^l_0$ and $F^l_1$. 
$\mathrm{CKA}_i(k,l)$ represents the similarity between features stored at pyramid levels $k$ and $l$ of a grid $F_i$, 
specifically $F^k_i$ and $F^l_i$ derived using the NTC method by~\cite{Vaidyanathan2023}. 

In our investigation (Figure~\ref{fig:cka}), we have observed substantial similarity across various levels of the feature pyramids, which suggests the presence of redundant data. To tackle this issue, we propose an innovative neural texture compression method based on an asymmetric autoencoder framework. The key contributions of our approach are as follows (see Figure~\ref{fig:main_arch_0}):
\begin{enumerate}
    \item Asymmetric autoencoder framework: Our ``Global transformer'' processes a texture set and generates representations that capture spatial-channel-resolution redundancy. The ``texture synthesizer'' then samples from these representations to reconstruct texels at specific positions and mip levels.
    \item Grid Constructor: The global transformer exclusively operates on the texture set at the highest resolution (mip level 0) and maps it into a pair of single-resolution representations. These pair of representations capture high and low frequency features of the texture set and serve as the representation across all mip levels.
    \item Grid sampler: To facilitate texel reconstruction from different mip levels, we sample the features stored in the representations using varying strides.
\end{enumerate}
By leveraging redundancies across pixels, different channels and mip levels, our method enhances texture compression while maintaining quality. 

\section{Related Work}
\label{sec:ralated}
\subsection{Conventional Texture Compression}
\label{sec:conv_tc}
Texture compression serves as a specialized technique for compressing texture maps within 3D computer graphics rendering systems. 
Unlike conventional image compression algorithms, which are designed for general images, texture compression algorithms specifically target random access scenarios~\cite{Beers96}. 
The primary challenge lies in allowing rapid random access to decompressed texture data, given the unpredictable order in which a renderer accesses texels. 
Notably, texture compression tolerates asymmetric encoding and decoding speeds, as the encoding process typically occurs only once during application authoring~\cite{Beers96}.

Most conventional texture compression algorithms involve lossy vector quantization of small fixed-size blocks of pixels into equally sized blocks of coding bits. These algorithms often include additional pre-processing and post-processing steps. 
For instance, Block Truncation Coding (BTC)~\cite{BTC1979} exemplifies a straightforward approach within this family of methods. 
Practical texture compression systems, such as S3 Texture Compression (S3TC)~\cite{S3TC}, PowerVR Texture Compression (PVRTC)~\cite{PVRTC2003}, Ericsson Texture Compression (ETC)~\cite{ETC2005}, and Adaptive Scalable Texture Compression (ASTC)~\cite{ASTC2012}, have been developed to address the growing texture storage demands in real-time applications. 
Interestingly, these systems still rely on block-based texture compression techniques initially introduced in the late 1980s for handling RGB data.
The primary limitation lies in the fact that existing systems can only compress textures with up to four color channels (e.g., RGB or RGBA). Additionally, they compress each mip level separately. Consequently, they cannot fully capture correlations across all channels of a texture set and its mip levels.
\subsection{Neural Compression}
\label{sec:nic}
Neural compression leverages the neural networks to compress data and has attracted great attention ~\cite{habibian2019video,rippel2019lvc,agustsson2020ssf,zhihao2022c2f,hu2021fvc,lu2019dvc,rippel2021elfvc,li2021deep,sheng2022temporal,li2022hybrid,li2023neural,Lucas2017,toderici2017full,balle2018variational,minnen2018joint,agustsson2019extreme,mentzer2020hific,he2022elic,agustsson2022multirealism,muckley2023improving,ghouse2023residual}. Among them, neural image compression (NIC)~\cite{Lucas2017,toderici2017full,balle2018variational,minnen2018joint,agustsson2019extreme,mentzer2020hific,he2022elic,agustsson2022multirealism,muckley2023improving,ghouse2023residual} performance is very competitive and some of them have surpassed conventional image compression. 
However, we can not apply them to the graphic texture compression directly due to the following limitations. 
Firstly, texture compression requires to random-access, which enabling to decompress only certain texels or texel blocks from a whole texture. 
Secondly, NIC methods often leverage entropy coding to compress the latent features, which is hardly possible for random-access due to the nature of entropy coding as a variable length coding method. 
Thirdly, neural image compression methods normally are designed for the RGB images, which contains three channels. In contrast, graphic textures often contain various channels for different materials, including albedo, normal, roughness and so on. 
For instance, in one single scenario containing a bottle of water, the materials of the bottle might contain more channels than the materials of the label attached to the bottle. 
It requires to develop a compression method working for materials with various channels. 
Lastly, neural image compression typically only needs to reconstruct the one single image at the input image resolution, while texture compression has to reconstruct texture with different resolutions (mip levels). 

\subsection{Implicit Neural Representation}
In earlier research, \cite{Dupont2021-tu,Strumpler2022-ha} explored RGB image compression by employing a Multi-Layer Perceptron (MLP). This MLP mapped pixel locations to RGB values and transmitted the MLP weights as a code for the Implicit Neural Representation (INR) of the image. 
However, overfitting such MLPs—referred to as implicit neural representations—proved challenging due to the high-frequency information inherent in natural images.
\cite{Dupont2021-tu,Strumpler2022-ha} demonstrated that this overfitting issue could be improved by incorporating sinusoidal encodings and activations. 
While~\cite{Dupont2021-tu} reported a Peak Signal-to-Noise Ratio (PSNR) below $30$dB, \cite{Strumpler2022-ha} achieved higher PSNR by leveraging meta-learning, specifically Model-Agnostic Meta-Learning~\cite{Finn2017-yo}, for initialization. 
However, this meta-learning approach is not directly applicable to texture compression, where a texture set may include information beyond color, and the number of channels can vary.
Despite enabling random access, INR-based compression fell short in achieving high PSNR. 
Subsequent research by~\cite{mueller2022instant,Takikawa2022-wi} addressed this limitation by introducing grid-based neural representations. 
These grid-based representations form the foundation for the first neural texture compression, as proposed by~\cite{Vaidyanathan2023}.

\subsection{Neural Texture Compression}
\label{sec:ntc}
In their groundbreaking work, Vaidyanathan et al.~\cite{Vaidyanathan2023} introduced the Neural Texture Compression (NTC) method, which fulfills random-access requirements. The NTC scheme leverages spatial redundancies across different mip levels and channels within a texture set. Their approach involves storing essential texel features for decompression in a pair of feature pyramids. These pyramids are optimized to minimize reconstruction loss while adhering to a specific bit rate determined by quantization levels. However, despite their effectiveness, there remains some redundancy between feature pyramid levels, impacting compression performance.

\section{Method}
\label{sec:method}
To handle the rendering requirements of texture compression, we utilized an asymmetric autoencoder framework to compress texture set $\mathbf{T}$. 
Texture set $\mathbf{T}$ is represented by a tensor of size $c \times h \times w$, where $c$ indicates the total number of channels of the textures in the texture set, $h$ and $w$ corresponds to the height and width of the textures, respectively. 
Here we assume each texture in a texture set has the same width and height.
Similar to~\cite{Vaidyanathan2023}, we do not make any assumptions about the channel count or the order of textures in the texture set.  
For example, the normals or diffuse, albedo could be mapped to any channels without having any impact on compression. 
This is because we train a compression model specifically for a given texture set. 

\begin{figure}[t]
    \centering
    \includegraphics[width=0.6\linewidth]{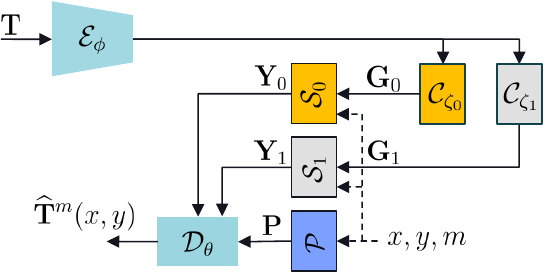}
    \caption{Overview of our method. Given a texture set $\mathbf{T}$, the encoder $\mathcal{E}_{\phi}$ produces a bottleneck latent representation $\mathbf{Z}_{\mathrm{SCR}}$.  The constructors $\mathcal{C}_i, i=0,1$ then utilize this latent representation to construct the grid-pair $\mathbf{G}_i$. At a specific position $x,y$ and mip level $m$, the grid samplers $\mathcal{S}_i$ extract features $\mathbf{Y}_i$ from $\mathbf{G}_i$. These extracted features, along with positional encoding $\mathbf{P}$ are subsequently fed into the decoder $\mathcal{D}_{\theta}$, which reconstructs the texel $\widehat{\mathbf{T}}^m(x,y)$ at the given position and mip level.}
    \label{fig:main_arch}
\end{figure}

Our neural texture compression consists of four main parts: (1) Global transformer $\mathcal{E}$, (2) Grid Constructors $\mathcal{C}_0, \mathcal{C}_1$, (3) Grid Samplers $\mathcal{S}_0, \mathcal{S}_1$ and (4) Texture synthesizer $\mathcal{D}$. 
In the following we elaborate each part. 
Figure~\ref{fig:main_arch} illustrate the framework of our neural texture compression. 

\subsection{Global transformation}
\label{sec:encoder}
The global transformer denoted as $\mathcal{E}_{\phi}(\mathbf{T})$ maps a texture set $\mathbf{T}$, which lies in the range $[0,1] ^ {c \times h \times w}$, to a bottleneck latent representation $\mathbf{Z}_{\mathrm{SCR}}$ in the space $[-0.5,0.5] ^ {c_z \times h_z \times w_z}$:
\[\mathcal{E}_{\phi}(\mathbf{T})=\mathbf{Z}_{\mathrm{SCR}}\]
The global transformer architecture we employ is closely aligned with the encoder in the neural image compression described in  \cite{he2022elic} excluding attention blocks. However, there is a crucial difference in our setup: the resolution of the bottleneck latent representation is down-scaled by a factor of $8$, specifically $h_z=h/8$ and $w_z=w/8$. Additionally, we enforce the constraint that $\mathbf{Z}_{\mathrm{SCR}}$ lies within the subspace $[-0.5,0.5] ^ {c_z \times h_z \times w_z}$ by applying the $\frac{1}{2}\tanh$ activation function at the last layer of the global transformer. 

Note: The input to the global transformer is always the texture set $\mathbf{T}$ with a resolution of $h \times w$. This is because texture sets corresponding to all mip levels (down to a resolution of $4 \times 4$) are derived by down-scaling $\mathbf{T}$. Specifically, for each level $m$ where $0 \leq m \leq M = \log_2 \max(h,w) - 2$, we have $\mathbf{T}^m \in [0,1]^{c \times h/2^m \times w/2^m}$. Importantly, all the information from these down-scaled texture sets must be preserved within the original texture set $\mathbf{T}$.

For further details on the global transformer architecture, please refer to Section\ref{sec:append_encoder}. 

\subsection{Grid Constructors}
\label{sec:grid_contructor}
Two grid constructors, denoted as $\mathcal{C}_{\zeta 0}$ and $\mathcal{C}_{\zeta 1}$, map the bottleneck latent representation $\mathbf{Z}_{\mathrm{SCR}}$ to grid pairs $\mathbf{G}_0$ and $\mathbf{G}_1$:
\[\mathcal{C}_{\zeta_i}(\mathbf{Z}_{\mathrm{SCR}}) = \mathbf{G}_i, \quad i=0,1\]
Each grid $\mathbf{G}_i$ is a tensor of size $c_{g_i} \times h_z \times w_z$, storing quantized features necessary for reconstructing texture sets at all mip levels.

In our setup, the grid constructors $\mathcal{C}_{\zeta_i}$ consist of two components: a linear projection $\mathcal{L}_{\zeta_i}$ and an asymmetric scalar quantizer $\mathcal{Q}_i$:
\[\mathcal{C}_i(\mathbf{Z}_{\mathrm{SCR}}) = (\mathcal{Q}_i \circ \mathcal{L}_i)(\mathbf{Z}_{\mathrm{SCR}}) = \mathbf{G}_i\]
Here, we constrain $\mathbf{G}_i$ to reside within the subspace $[-0.5,0.5]^{c_{g_i} \times h_z \times w_z}$.

To quantize the features of grid $\mathbf{G}_i$, we employ $\mathcal{Q}_i$ asymmetric scalar quantization, following the approach described by~\cite{Jacob_2018_CVPR}. The quantization range is:
\[\left[-\frac{2^{B_i}-1}{2^{B_i+1}}, \frac{1}{2}\right]\]
where $B_i$ denotes the number of bits allocated to store each element of $\mathbf{G}_i$.

\subsection{Grid sampler}
\label{sec:grid_sampler}
The grid samplers, denoted as $\mathcal{S}_i$ for $i=0,1$, sample the grids $\mathbf{G}_i$ based on a given texture coordinate $(x,y)$ within the range $-1 \leq x,y \leq +1$ and at a specific mip level $m$ where $0 \leq m \leq M$:
\[\mathcal{S}_i(\mathbf{G}_i | x,y,m) = \mathbf{Y}_i, \quad i=0,1\]
Given the coordinate $(x,y)$ at mip level $m$, the grid sampler $\mathcal{S}_0(\mathbf{G}_0 | x,y,m)$ identifies the four surrounding voxels and concatenates the $c_{g_0}$-dimensional features stored at each corner of the voxel to produce $\mathbf{Y}_0$. 

Similarly, the grid sampler $\mathcal{S}_1(\mathbf{G}_1 | x,y,m)$ locates the four surrounding voxels and linearly interpolates the $c_{g_1}$-dimensional features stored at each corner of the voxel based on the relative position of $(x,y)$ within the voxel. The resulting output is $\mathbf{Y}_1$.

By employing this approach, $\mathbf{G}_0$ captures more detailed features of the texture set, while $\mathbf{G}_1$ focuses on more abstract information.

In contrast to the method proposed by~\cite{Vaidyanathan2023}, which utilizes multiple grid-pairs of varying resolutions, our setup employs a single resolution grid-pair. To handle mip levels $m$ where $m > 3$ (which have lower resolution than the grid-pair $\mathbf{G}_i$ for $i=0,1$), we perform sampling with a stride of $s=2^{m - 3}$. After identifying the top-left corner of the surrounding voxel, the remaining corners are chosen with respect to this top-left corner using the specified stride.

\subsection{Text synthesis}
\label{sec:Text_synthesis}
The text synthesizer, denoted as $\mathcal{D}_{\theta}$, takes as input the concatenation of grid samples $\mathbf{Y}_0, \mathbf{Y}_1$, the normalized mip level $\Tilde{m}=\sfrac{m}{M}$ and the positional encoding $\mathcal{P}(x,y)$ \cite{mueller2022instant} and outputs  $\mathbf{T}^m(x,y)$ texel at coordinate $(x,y)$ of mip level $m$:
    \[\mathcal{D}_{\theta}(\mathbf{Y}_0, \mathbf{Y}_1, \Tilde{m}, \mathbf{P})=\widehat{\mathbf{T}}^m(x,y)\] 
The text synthesizer  $\mathcal{D}_{\theta}$, comprises fully connected layers with skip connections. Detailed information about the text synthesizer architecture is provided in Section~\ref{sec:append_decoder}. 

\subsection{Loss functions}
The loss function for a given texture set $\mathbf{T}$ at coordinate $(x,y)$ of mip level $m$ is defined as 
{\small
\begin{equation}
    \mathcal{L}_{\mathbf{T}}^m(x,y) = D\left( 
    \mathcal{D}_{\theta}(\mathcal{S}_0(\mathcal{C}_{\zeta_0}(\mathcal{E}_{\phi}(\mathbf{T}))|x,y,m), \mathcal{S}_1(\mathcal{C}_{\zeta_1}(\mathcal{E}_{\phi}(\mathbf{T}))|x,y,m), \Tilde{m}, \mathbf{P}) - \mathbf{T}^m(x,y) 
    \right)
\end{equation}
}
where $D(\cdot)$ indicates the distortion metric for the reconstructed texture set.

\section{Experiment}
\label{sec:exp}
\subsection{Experimental setups}
\subsubsection{Datasets}
We commence by assessing the performance of various texture compression techniques across diverse texture sets. 
Our selection comprises seven distinct materials, each associated with texture sets that exhibit varying channel counts. 
These texture sets boast a resolution of $2048 \times 2048$ texels, with channel counts spanning from $5$ to $12$.
Notably, these public texture sets partially align with those employed in a previous study by~\cite{Vaidyanathan2023}, as their complete dataset is not publicly available. 
These publicly accessible texture sets are sourced from ambientCG\footnote{\url{https://ambientcg.com/}} and PolyHaven\footnote{\url{https://polyhaven.com/}}.

\subsubsection{Implementation details}
Our model undergoes training in multiple stages. 
When dealing with a specific texture set $\mathbf{T}$, the initial step involves randomly cropping it to a size of $256 \times 256$. 
Subsequently, we pass this cropped texture set to a global transformer, followed by grid constructors that facilitate the construction of the corresponding grid pair.

For texel reconstruction, we randomly select a mip level in proportion to the area of that mip level. 
This selection is achieved by sampling from an exponential distribution with a rate parameter $\lambda=\log 4$~\cite{Vaidyanathan2023}. 
To address the issue of undersampling low-resolution mip levels, we incorporate a strategy where $10\%$ of the batches randomly select their mip level from a uniform distribution spanning the entire chain of mip levels. 
This approach helps ensure a more balanced representation across different resolution levels during training.

In each subsequent stage, we increase the crop-size $C_s$ by a factor of $2$ until the texture synthesizer is capable of reconstructing the complete chain of mip levels. 
Our training setup employs a batch size of $4$ and a learning rate (LR) of $10^{-4}$, which we decrease by a factor of $2$ as we increment the crop size at each stage. 
At the final stage, we train the model for $20,000$ steps with a LR of $10^{-5}$.
 
During the earlier stages, we follow the approach proposed by~\cite{balle2017} to replace quantization with additive uniform noise within the range $\left(-\frac{1}{2^{B_i+1}},\frac{1}{2^{B_i+1}}\right)$. 
However, in the final stage, we explicitly quantize the feature values using the straight-through estimator (STE) introduced by~\cite{Lucas2017}. 
We enforce a fixed quantization rate of $B_i=4, i=0,1$ for all feature values in the grid-pair $\mathbf{G}_i$, optimizing solely for distortion. 
Across all stages, we employ the Mean Squared Error (MSE) as the distortion metric for calculating the loss, and we use the Adam optimizer~\cite{kingma2017adam}. Table~\ref{tab:training_stages} shows the training stages for a $2048 \times 2048$ texture set and a mip level range of $m=0,\ldots,9$.
\begin{table}[t]
    \centering
        \caption{Training stages for a $2048 \times 2048$ texture set across a range of mip levels.}
    \begin{tabular}{c|c|c|c|c|c}
    \hline
        Stage & Crop-size & Learning-rate & Steps & Possible mip levels & Quantization\\ \hline
        0 & $256 \times 256$ & $10 \times 10^{-5}$ & $160,000$ & $m=0,\ldots, 8$ & uniform noise\\
        1 & $512 \times 512$ & $5 \times 10^{-5}$ & $80,000$ & $m=0,\ldots, 9$ & uniform noise\\
        2 & $512 \times 512$ & $1 \times 10^{-5}$ & $20,000$ & $m=0,\ldots, 9$ & quantized STE \\
        \hline
    \end{tabular}
    \label{tab:training_stages}
\end{table}

\subsubsection{Evaluation metrics}
We measure the rate in bits-per-pixel-per-channel (BPPC), considering that the total number of channels varies across different texture sets. 
The BPPC encompasses both the bits required to allocate the grid-pair and the parameters specific to the text synthesizer, which is trained uniquely for each texture set. 
Throughout our evaluation, we consistently set $B_i=4$ to store each grid-pair $\mathbf{G}_i$ (where $i=0,1$). The total number of bits needed to store a grid-pair is given by $\frac{c_{g_i} \cdot h \cdot w}{32}$. 
Here, $c_{g_i}$ represents the number of channels of the grid $\mathbf{G}_i$, and $h\times w$ denotes the resolution of the texture set at the zero mip level. 
Additionally, the total number of bits required to store the texture synthesizer corresponds to its total number of parameters, multiplied by the parameter precision. 
To control the bit rate, we manipulate the number of channels in the grid-pair ($c_{g_i}$) and adjust the hidden layers in both the global transformer and texture synthesizer. 

Following the same approach as in the work by~\cite{Vaidyanathan2023}, for assessing reconstruction fidelity,
we rely on two commonly used metric, 
\begin{enumerate}
    \item Peak Signal-to-Noise Ratio (PSNR): This metric evaluates the reconstruction quality between the original texture set $\{\mathbf{T}^m\}_{m=0}^M$ and the reconstructed one $\{\widehat{\mathbf{T}}^m\}_{m=0}^M$ through all mip levels. 
    It is defined as:
    \[\small\mathrm{PSNR}\left(\{\mathbf{T}^m\}_{m=0}^M, \{\widehat{\mathbf{T}}^m\}_{m=0}^M\right) = -10 \log_{10} \left[ \frac{\sum_m (\mathbf{T}^m, \widehat{\mathbf{T}}^m)^2}{c \sum_m h_m w_m}\right]\]
    \item Structural SIMilarity (SSIM)~\cite{Wang2004-of}: This metric also evaluates the similarity between the original and reconstructed texture sets across all mip levels. SSIM is a single-channel measure, providing a quality index between 0 and 1, where higher is better.
    It is defined as: 
    \[\small\mathrm{SSIM}\left(\{\mathbf{T}^m\}_{m=0}^M, \{\widehat{\mathbf{T}}^m\}_{m=0}^M\right) = \left[ \frac{\sum_m\sum_i h_m w_m \mathrm{SSIM}(\mathbf{T}_i^m, \widehat{\mathbf{T}}_i^m)}{c \sum_m h_m w_m}\right]
    \]
    where $\mathbf{T}_i^m$ denotes the $i$th channel of the texture set $\mathbf{T}^m$ at the mip level $m$. 
\end{enumerate}

To summarize the rate-distortion curve in a single number, we also report the Bj{\o}ntegaard-Delta rate (BD-rate)~\cite{bjontegaard2001bdrate}. 
This represents the average bitrate saving for a fixed quality when compared to a reference compression method.

\subsubsection{Compared methods}
In our comparative evaluation, we assessed our compression method alongside NTC (as proposed by~\cite{Vaidyanathan2023}). 
Since the NTC source code was not publicly available at the time of writing this paper, we re-implemented their approach and conducted evaluations on the same dataset as ours. 
Notably, our re-implementation closely aligns with the performance reported in~\cite{Vaidyanathan2023}, with the minor discrepancy arising from our use of a subset of publicly available texture sets.

Additionally, we compared our method's performance against ASTC\footnote{Available at \url{https://github.com/ARM-software/astc-encoder}.}. 
Following the same settings as outlined in \cite{Vaidyanathan2023}, we utilized the `\verb|-exhaustive|' flag to achieve the best quality. Specifically, we employed the two most aggressive variants of ASTC, which compressed $12 \times 12$, $10 \times 10$, $8 \times 8$ and $6 \times 6$ tiles.

\subsection{Results}
\subsubsection{Rate-Distortion performance}
\begin{figure}[t]
    \centering
     \begin{subfigure}[b]{0.47\textwidth}
         \centering
         \includegraphics[width=1\textwidth]{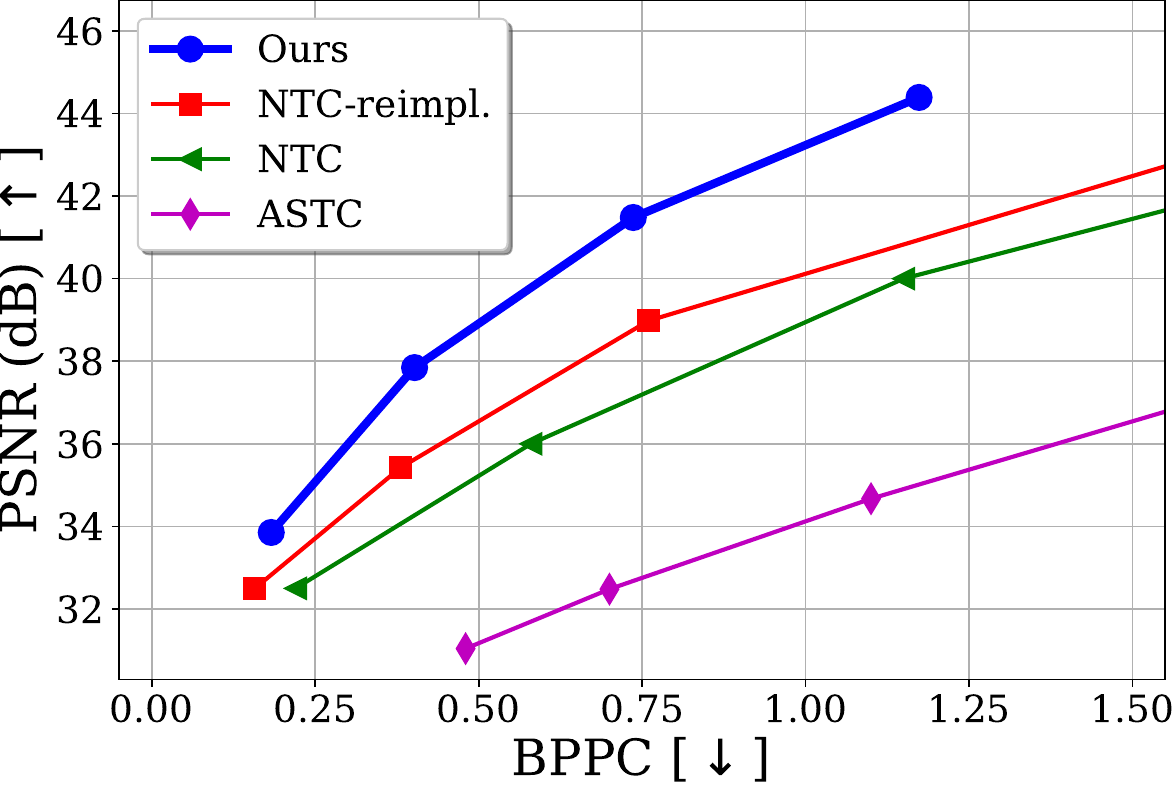}
         \label{fig:rd_curve_psnr}
     \end{subfigure}
     \hspace{-0cm}
     \begin{subfigure}[b]{0.485\textwidth}
         \centering
         \includegraphics[width=1\textwidth]{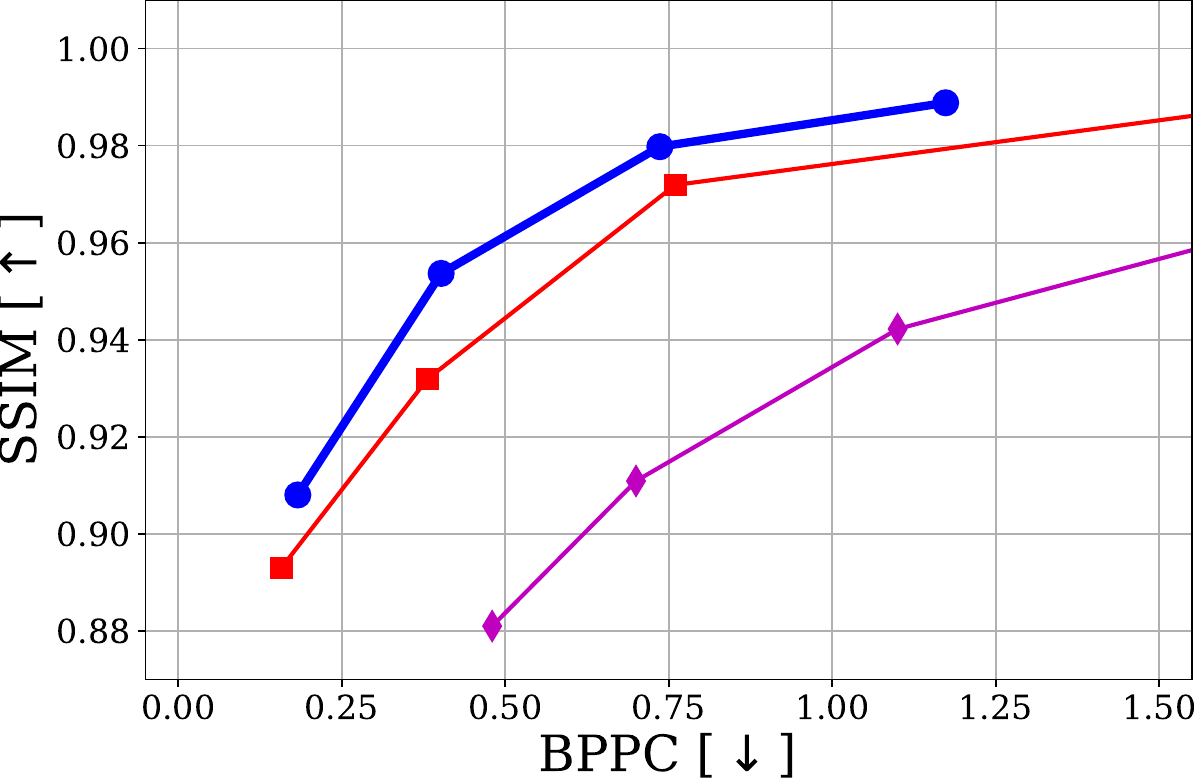}
         \label{fig:rd_curve_ssim}
     \end{subfigure}
    \caption{Comparison with the state-of-the-art method NTC~\cite{Vaidyanathan2023}.}
    \label{fig:rd_curve}
\end{figure}

Figure~\ref{fig:rd_curve} illustrates the rate-distortion curve in terms of PSNR for our compression method, our re-implementation of NTC (as described in~\cite{Vaidyanathan2023}), and ASTC. 
These methods were evaluated on the same dataset, including the NTC results reported in \cite{Vaidyanathan2023}. 

As depicted in Figure~\ref{fig:rd_curve}, our compression method consistently outperforms both NTC and ASTC across all bit ranges. Notably, our approach demonstrates a significant BD-rate improvement of $40.8\%$ compared to NTC (our re-implementation).

Although our compression is not optimized for SSIM~\cite{Wang2004-of}, we have included in Figure~\ref{fig:rd_curve}. Similarly, our method consistently outperforms the other methods. 

\subsubsection{Qualitative results}
Figure~\ref{} shows original and reconstructed texture set of our method, referred to as Convolutional Neural Texture Compression (CNTC), versus NTC~\cite{Vaidyanathan2023}.  
The compression profiles corresponds to the lowest BPPC cases shown in Figure~\ref{fig:rd_curve}. 

\begin{table}[t]
    \centering
        \caption{PSNR scores of our re-implementation of \cite{Vaidyanathan2023} vresus ours (CNTC) for the diffuse (first row) and normal (second row) maps in ``Ceramic roof 01'' texture set.}    \vspace{-0.3cm}
    \resizebox{\textwidth}{!}{
    \begin{tabular}{c||c|c||c|c||c|c||c}
    \hline 
    & NTC 0.2 & CNTC 16 & NTC 0.5 & CNTC 32 & NTC 1.0 & CNTC 64 & Reference \\ \hline\hline
           \includegraphics[width=0.2\textwidth]{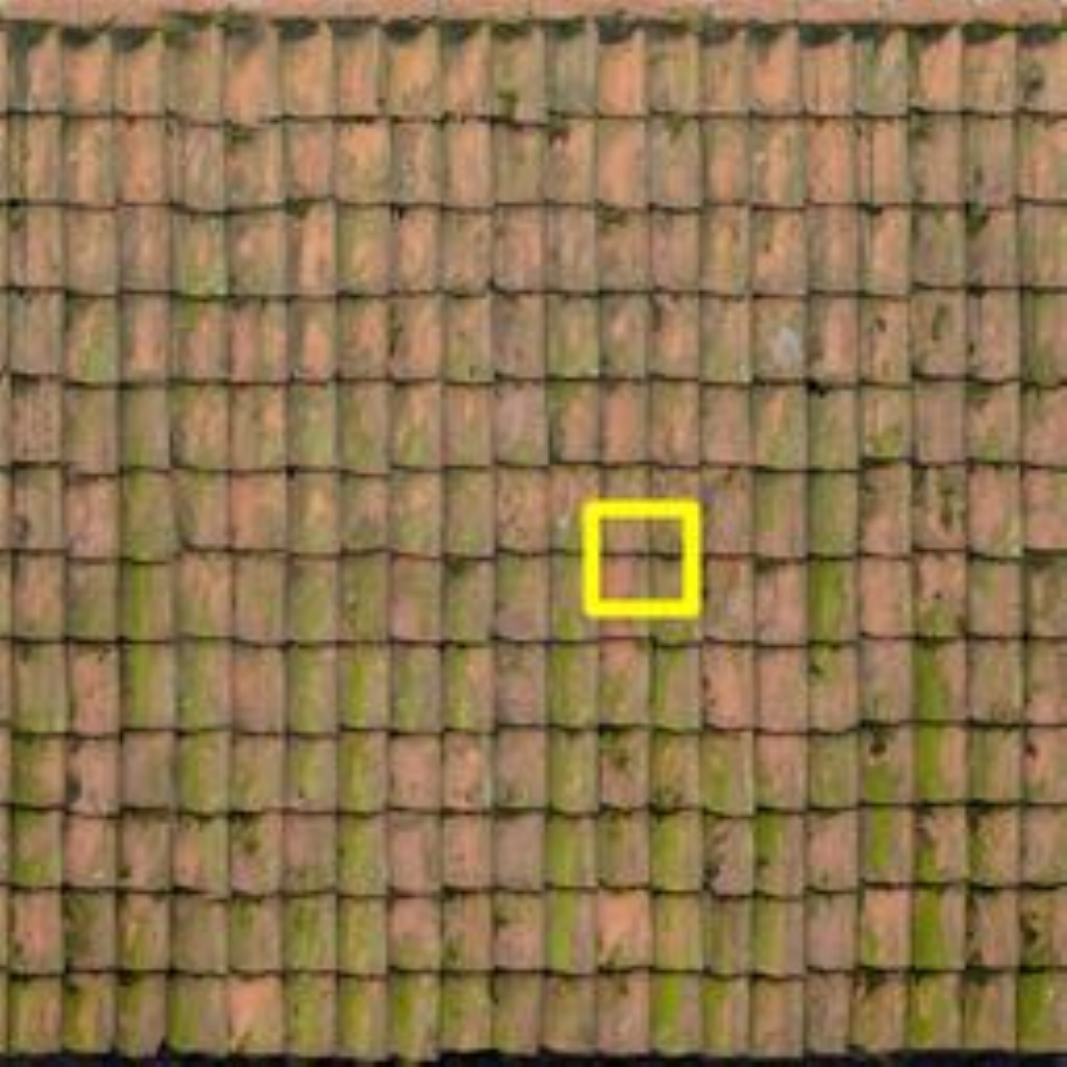} 
       & \includegraphics[width=0.2\textwidth]{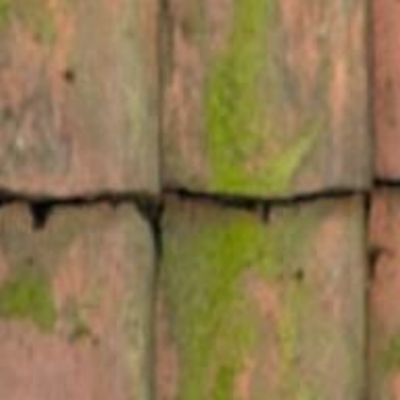}
       & \includegraphics[width=0.2\textwidth]{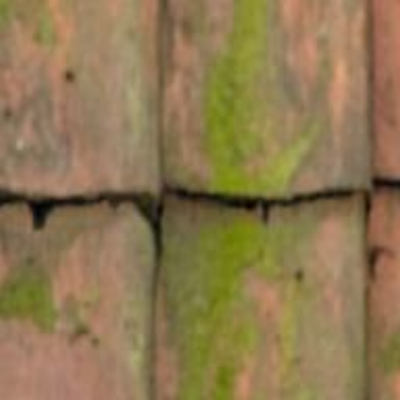} 
       & \includegraphics[width=0.2\textwidth]{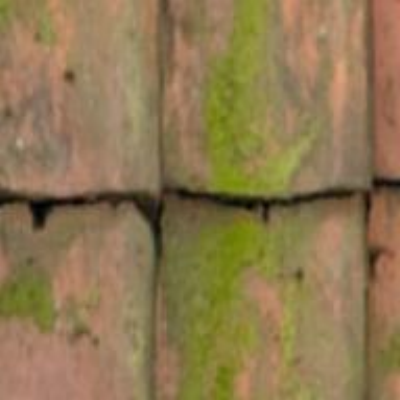} 
       & \includegraphics[width=0.2\textwidth]{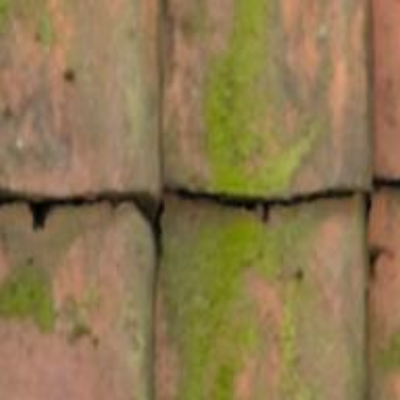}
       & \includegraphics[width=0.2\textwidth]{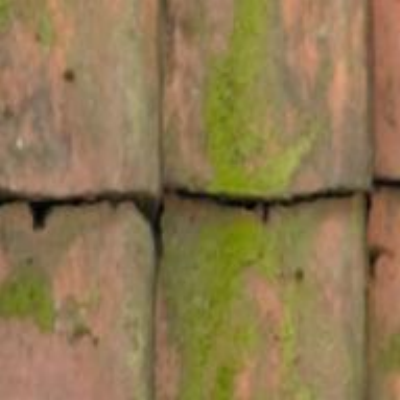}
       & \includegraphics[width=0.2\textwidth]{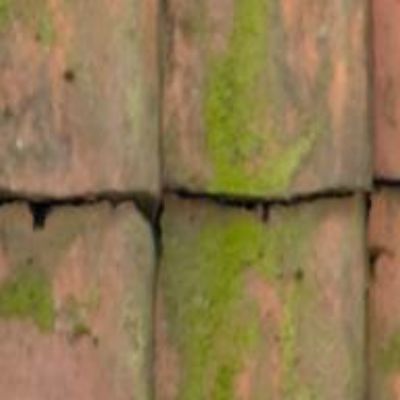}
       & \includegraphics[width=0.2\textwidth]{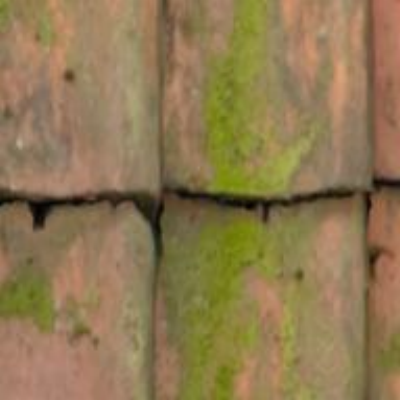}
       \\
        {\small PSNR [$\uparrow$] (BPPC [$\downarrow$])} & $35.81$ ($0.17$) & $36.82$ ($0.18$) & $38.78$ ($0.38$)& $40.53$ ($0.39$) & $41.29$ ($0.76$)&  $44.04$ ($0.73$) & Ground Truth 
     \\ \hline \hline 
           \includegraphics[width=0.2\textwidth]{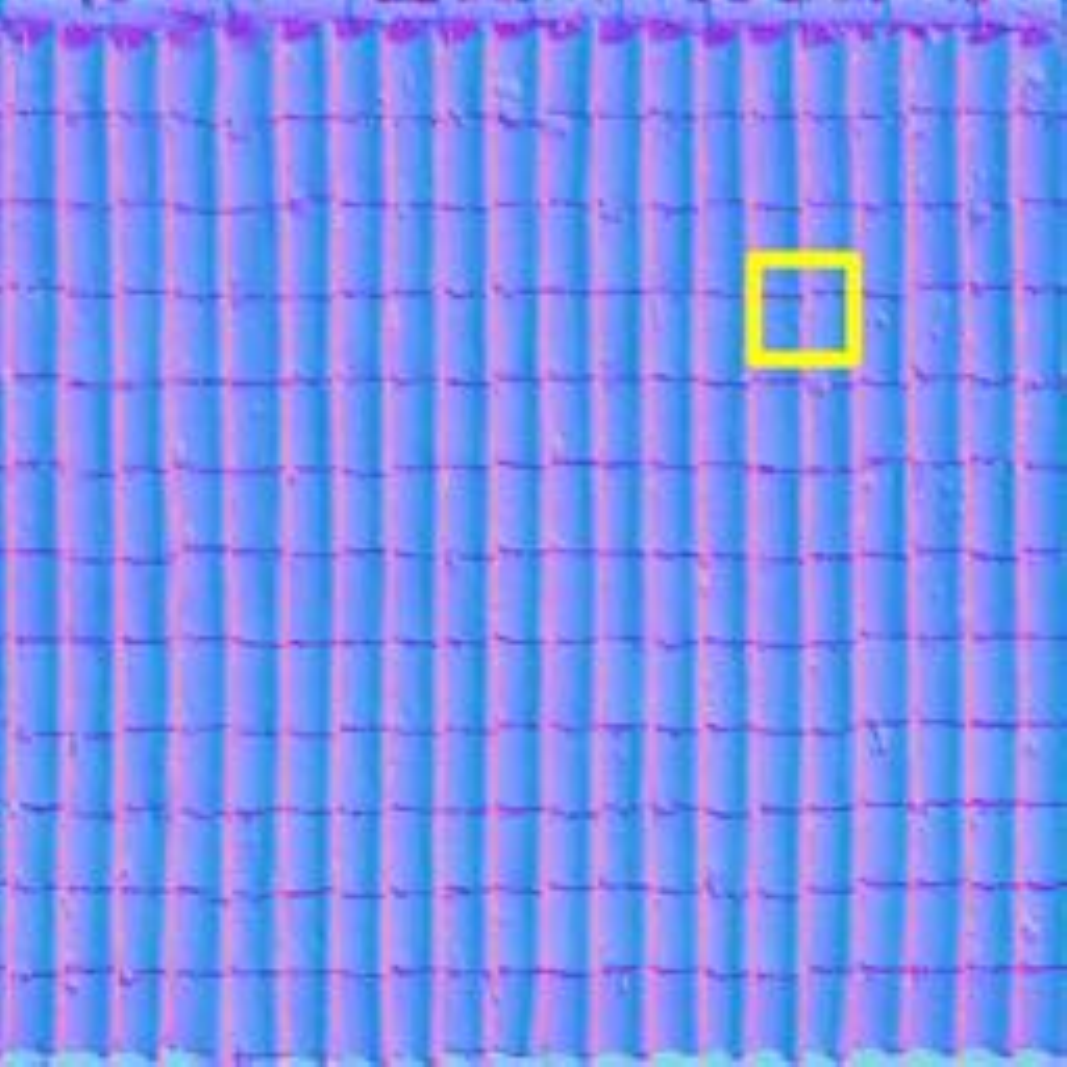} 
       & \includegraphics[width=0.2\textwidth]{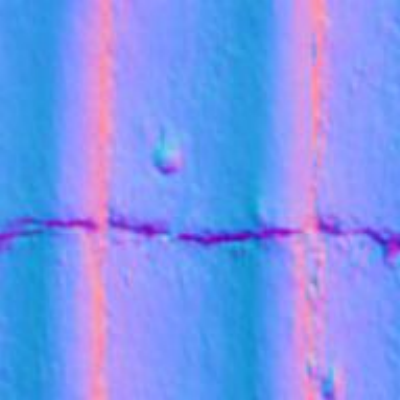}
       & \includegraphics[width=0.2\textwidth]{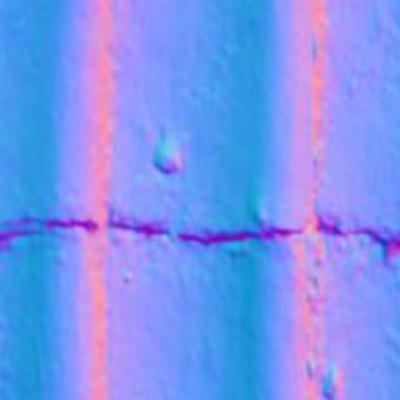} 
       & \includegraphics[width=0.2\textwidth]{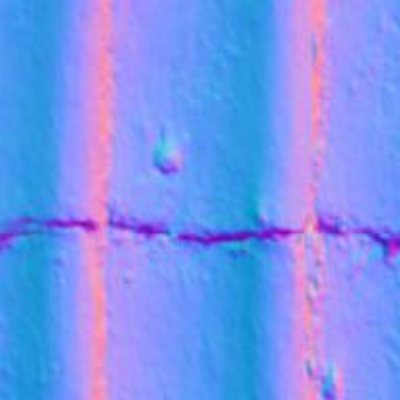} 
       & \includegraphics[width=0.2\textwidth]{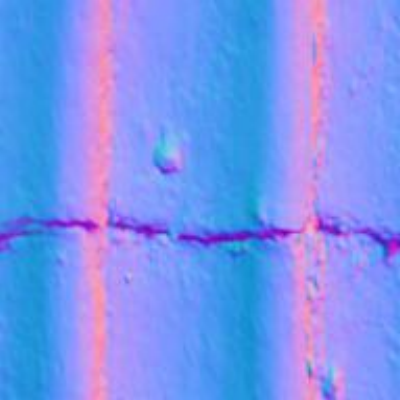}
       & \includegraphics[width=0.2\textwidth]{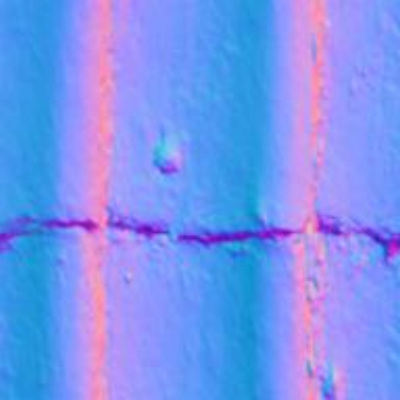}
       & \includegraphics[width=0.2\textwidth]{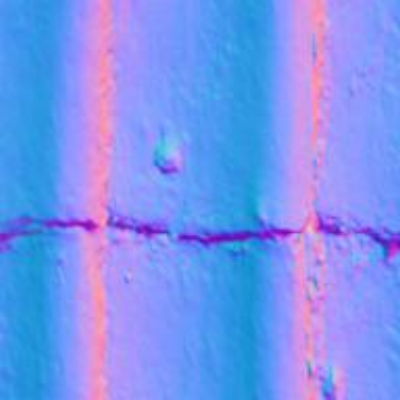}
       & \includegraphics[width=0.2\textwidth]{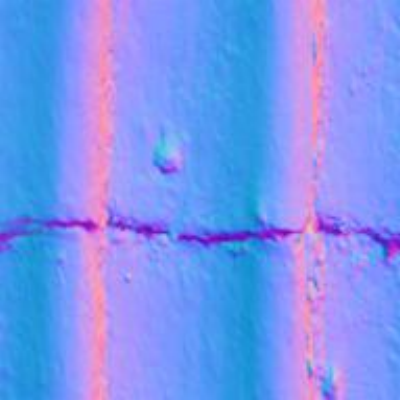}
       \\
        {\small PSNR [$\uparrow$] (BPPC [$\downarrow$])} & $33.64$ ($0.17$) & $35.15$ ($0.18$) & $36.242$ ($0.38$) & $39.85$ $(0.39)$ & $39.32$ ($0.76$) & $42.66$ ($0.73$) & Ground Truth \\ \hline
    \end{tabular}
    }
    \label{tab:my_label}
\end{table}

\subsection{Analysis}
\subsubsection{Effect of interpolation in grid-samplers}
\begin{figure}[t!]
\centering
    \begin{subfigure}[t]{0.20\textwidth}
         \centering         
         \includegraphics[width=1\textwidth]{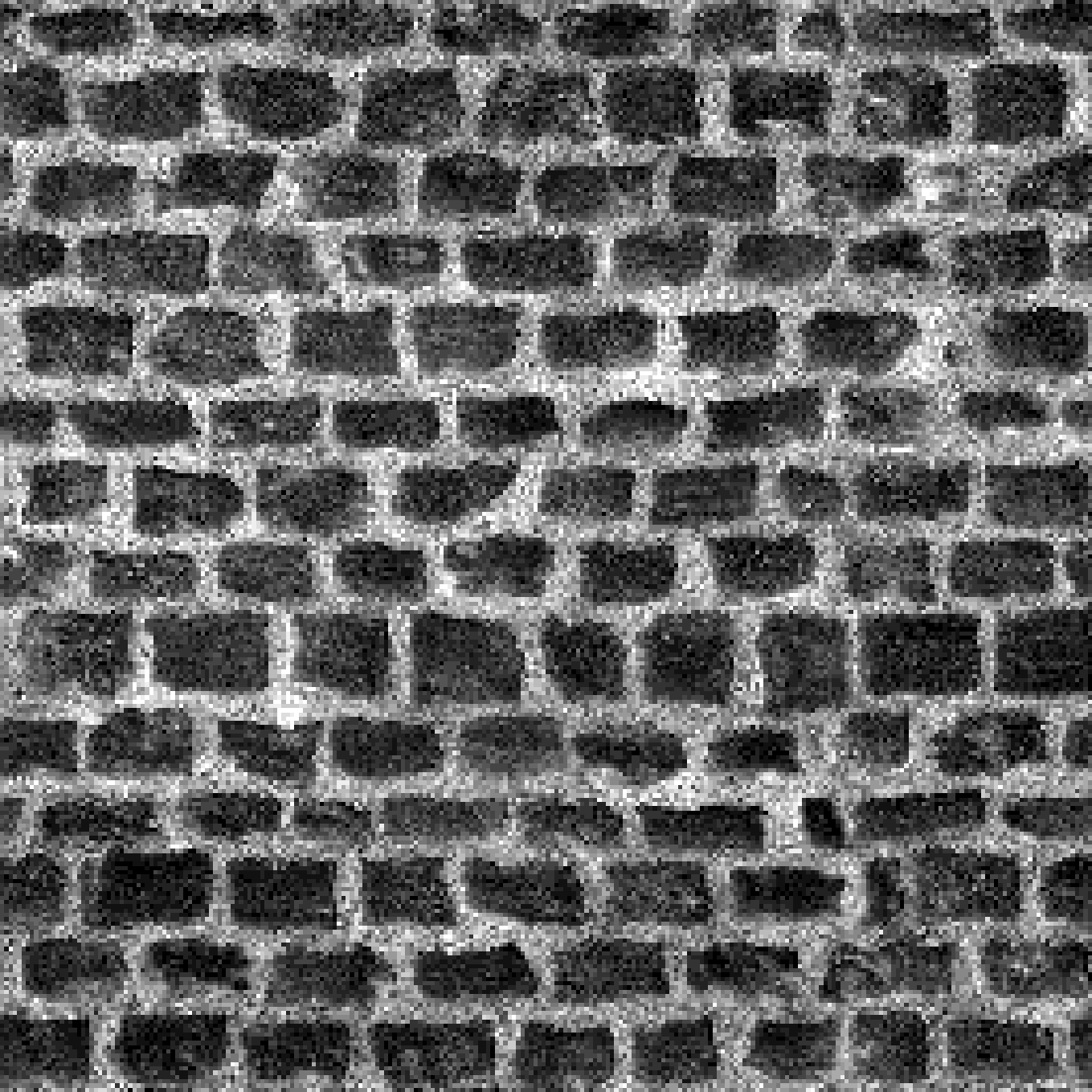}         
         \caption{$\mathbf{G}_0$}
    \end{subfigure}
    \hspace{0.5cm}
    \begin{subfigure}[t]{0.20\textwidth}
         \centering
         \includegraphics[width=1\textwidth]{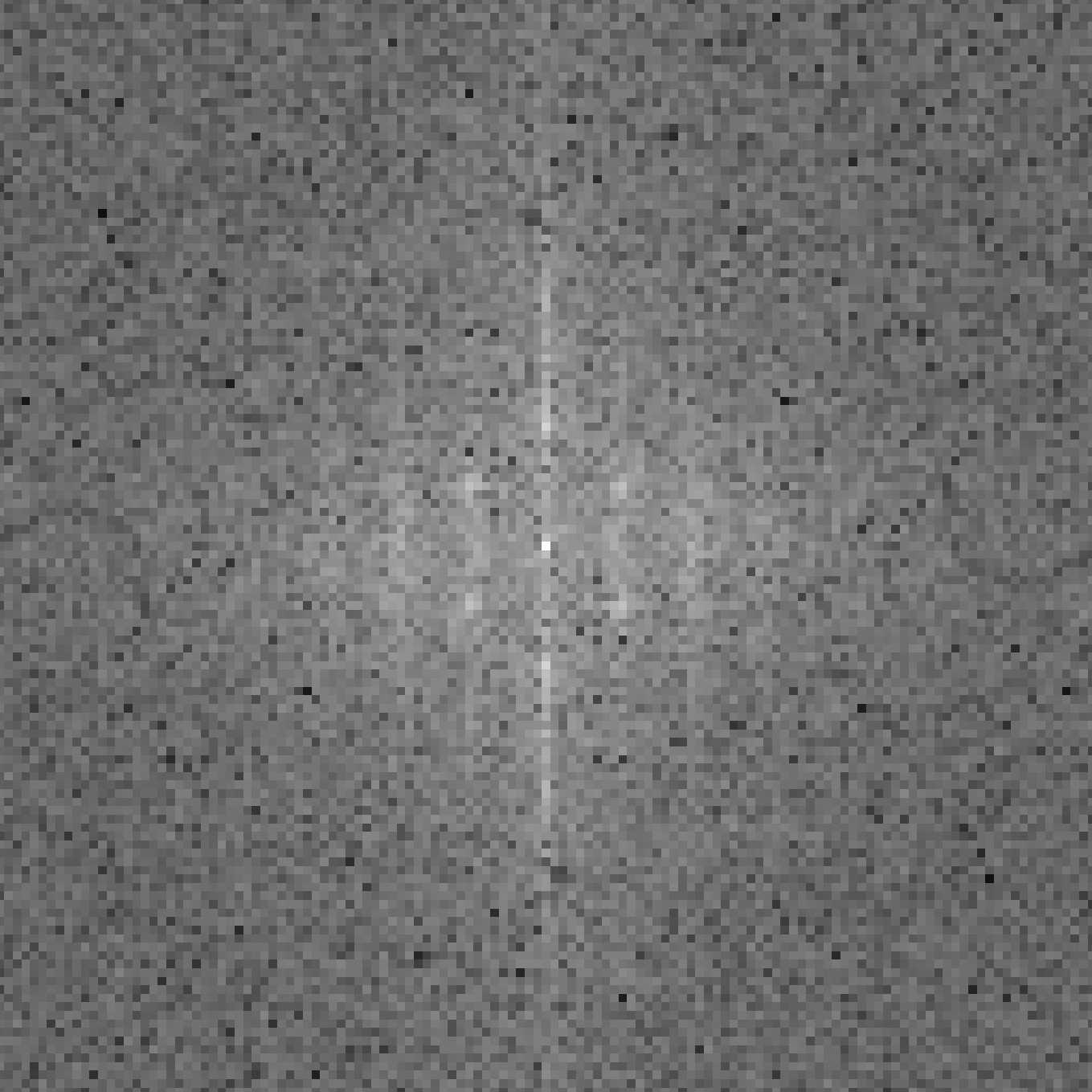}
         \caption{$\mathcal{F}(\mathbf{G}_0)$}
         \label{fig:grid_0_fft}
    \end{subfigure}
    \hspace{0.5cm}
    \begin{subfigure}[t]{0.20\textwidth}
         \centering
         \includegraphics[width=1\textwidth]{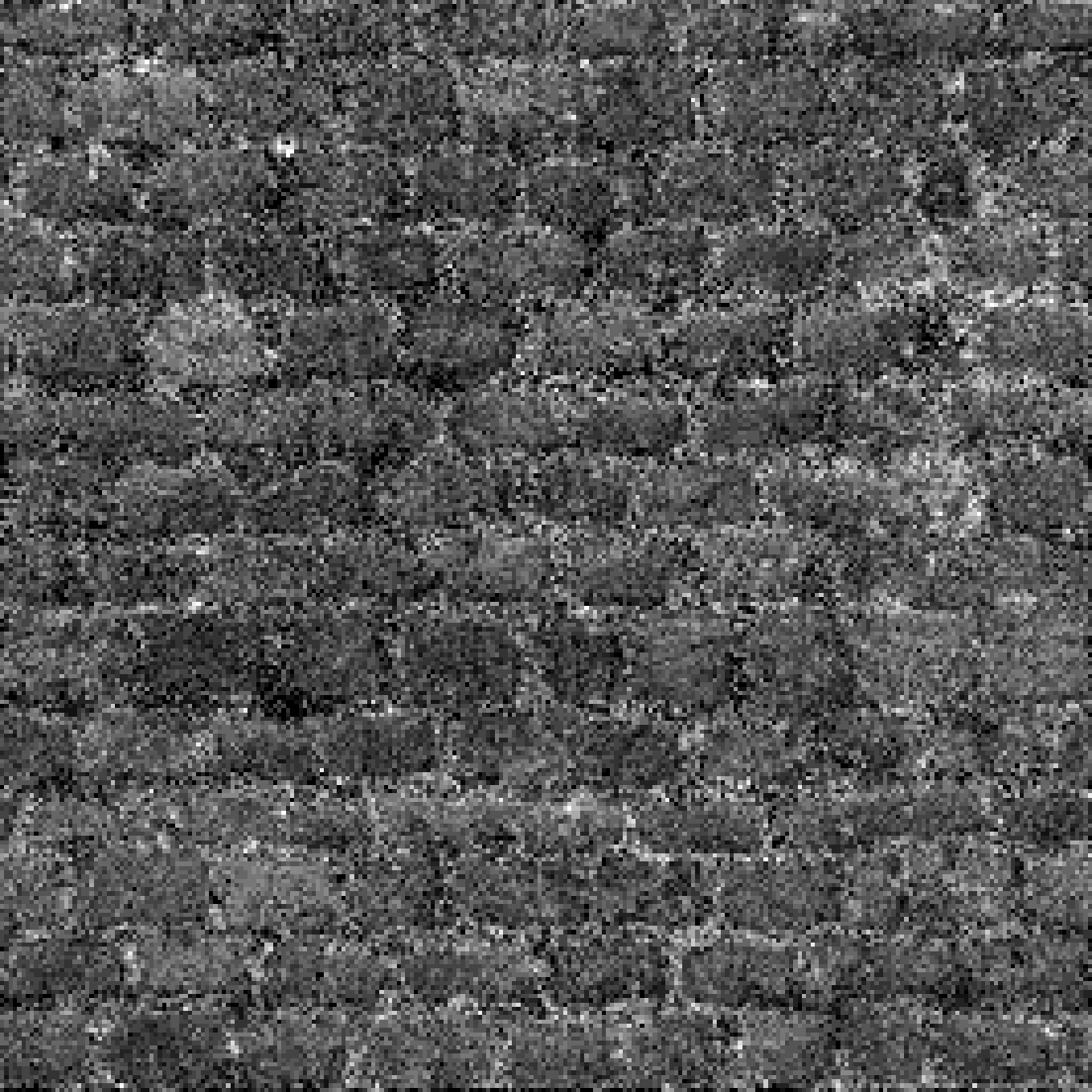}         
         \caption{$\mathbf{G}_1$}
    \end{subfigure}
    \hspace{0.5cm}
    \begin{subfigure}[t]{0.20\textwidth}
         \centering
        \includegraphics[width=1\textwidth]{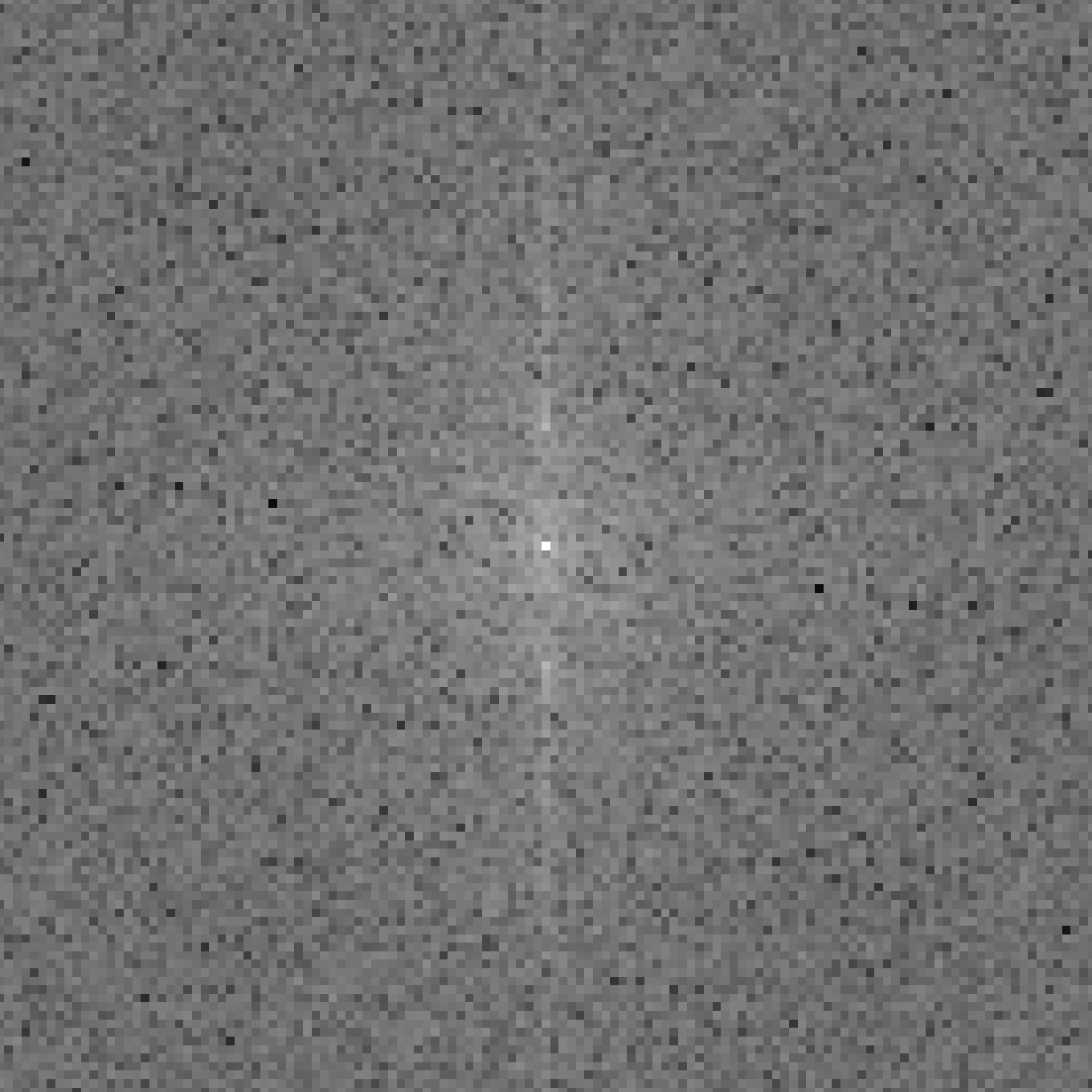}
         \caption{$\mathcal{F}(\mathbf{G}_1)$}
         \label{fig:grid_1_fft}
    \end{subfigure}
    \caption{
The Fourier transform ($\mathcal{F}$) of the 4th channel in Grid-pair $\mathbf{G}_i, i=0,1$ for the Paving Stones 131 texture set reveals interesting characteristics. Specifically, $\mathbf{G}_0$ captures higher-frequency features, while $\mathbf{G}_1$ focuses on low-frequency features.}
    \label{fig:grid_fft}
\end{figure}

Figure~\ref{fig:grid_fft} displays the 4th channel of the grid-pair $(\mathbf{G}_0, \mathbf{G}_1)$ alongside their corresponding absolute Fourier transforms.

In Figure~\ref{fig:grid_1_fft}, the Fourier transform of $\mathbf{G}_1$ exhibits a concentration around the center (low-frequency components). 
This design choice aims to capture more abstract features.
Conversely, Figure~\ref{fig:grid_0_fft} showcases that $\mathbf{G}_0$ is adept at capturing higher frequency components and providing more detailed information.

\subsubsection{Effect of Global transformer}
\begin{figure}[t!]
\centering
     \begin{subfigure}[b]{0.4\textwidth}
         \centering
         \includegraphics[width=1\textwidth]{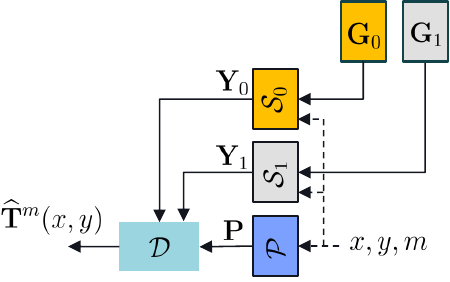}
         \caption{No encoder}
         \label{fig:wo_encoder}
     \end{subfigure}
     \hspace{0.5cm}
     \begin{subfigure}[b]{0.5\textwidth}
         \centering
         \includegraphics[width=1\textwidth]{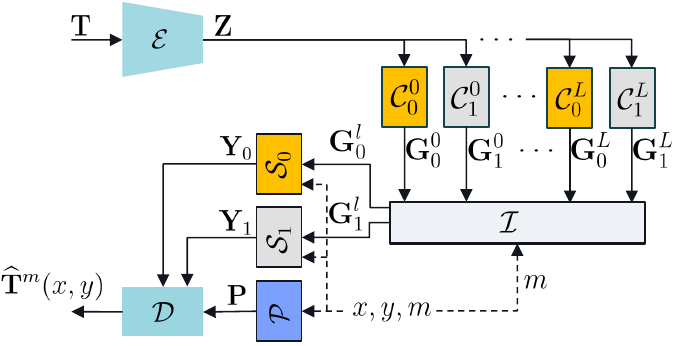}
         \caption{Multi-resolution grid-pairs}
         \label{fig:wo_stride}
     \end{subfigure}
        \caption{Analysis of the effect of (a) Encoder and (b) grid sampling with stride.}
        \label{fig:grid_samplers}
\end{figure}

As illustrated in Figure~\ref{fig:wo_encoder} in order to demonstrate the impact of the Global transformer and Grid constructors, we removed them from the compression framework and directly trained grid-pairs as model parameters using Stochastic Gradient Descent (SGD), following the approach in~\cite{Vaidyanathan2023}.
However, due to the grid-pair being $8$ times smaller (both horizontally and vertically) than the texture set $\mathbf{T}$, the grid-pair learned via SGD struggles to capture high-frequency information. 
As depicted in Figure~\ref{fig:wo_encoder_rd}, the PSNR of the model without a global transformer is significantly lower than that of the model with one. 
Increasing the bit-rate does not substantially improve the situation.

\subsubsection{Effect of sampling with stride}
To investigate the impact of grid sampling with stride, we made a crucial modification. 
Instead of using the original grid-pair $(\mathbf{G}_0,\mathbf{G}_1)$, we introduced ``multi-resolution grid-pairs'' denoted as $\{\mathbf{G}^l_0, \mathbf{G}^l_1\}_{l=0}^{M-3}$. 
Each grid-pair $(\mathbf{G}^l_0,\mathbf{G}^l_1)$ has a resolution $2^l$ times lower than that of $(\mathbf{G}_0,\mathbf{G}_1)$.
Refer to Figure~\ref{fig:wo_stride} for an illustration of the model without stride. 
Additionally, we introduce an indicator function $\mathcal{I}$ that selects the appropriate grid-pair $(\mathbf{G}^l_0,\mathbf{G}^l_1)$ based on the mip-level $m$. Specifically, $\mathcal{I}(m)$ equals 0 if $m\leq 3$, and otherwise, it is $m-3$.
In our study, the grid constructor $\mathcal{C}^l_i(\mathbf{Z})$ applies average pooling $\mathcal{A}^l$ with a kernel size and stride of $2^l$, following the linear projection $\mathcal{L}^l_i$:
\[\mathcal{C}^l_i(\mathbf{Z}) = (\mathcal{Q}_i \circ \mathcal{A}^l \circ \mathcal{L}^l_i)(\mathbf{Z}) = \mathbf{G}_i\]
This approach enables us to investigate the influence of stride on grid sampling and its implications for our model. As illustrated in Figure~\ref{fig:wo_encoder_rd}, despite the model with multi-resolution grid pairs having a higher BPPC (due to storing all resolutions of the grid-pairs), its PSNR is lower than that of our original model. 

\subsubsection{Effect of synthesizer depth}
\begin{figure}[t!]
\centering
     \begin{subfigure}[b]{0.31\textwidth}
         \centering
         \includegraphics[width=1\textwidth]{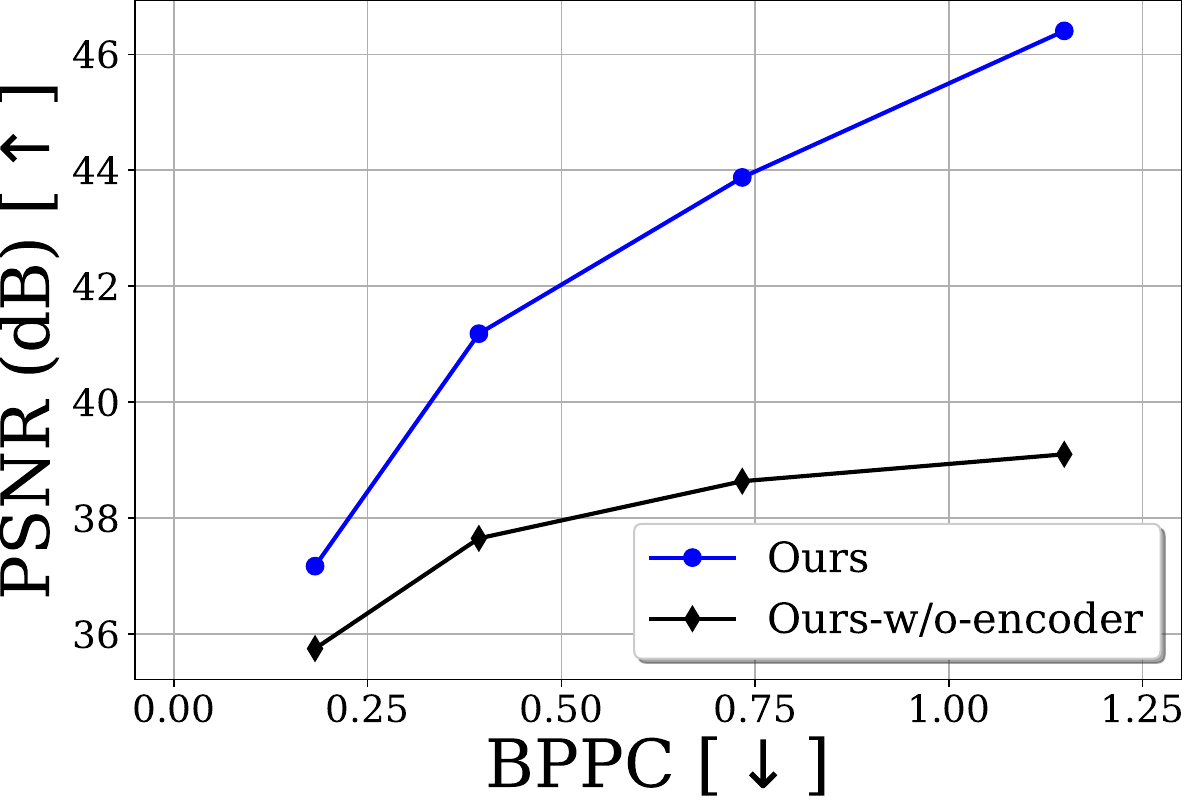}
         \caption{Encoder}
         \label{fig:wo_encoder_rd}
     \end{subfigure}
     \hspace{0.1cm}
     \begin{subfigure}[b]{0.31\textwidth}
         \centering
         \includegraphics[width=1\textwidth]{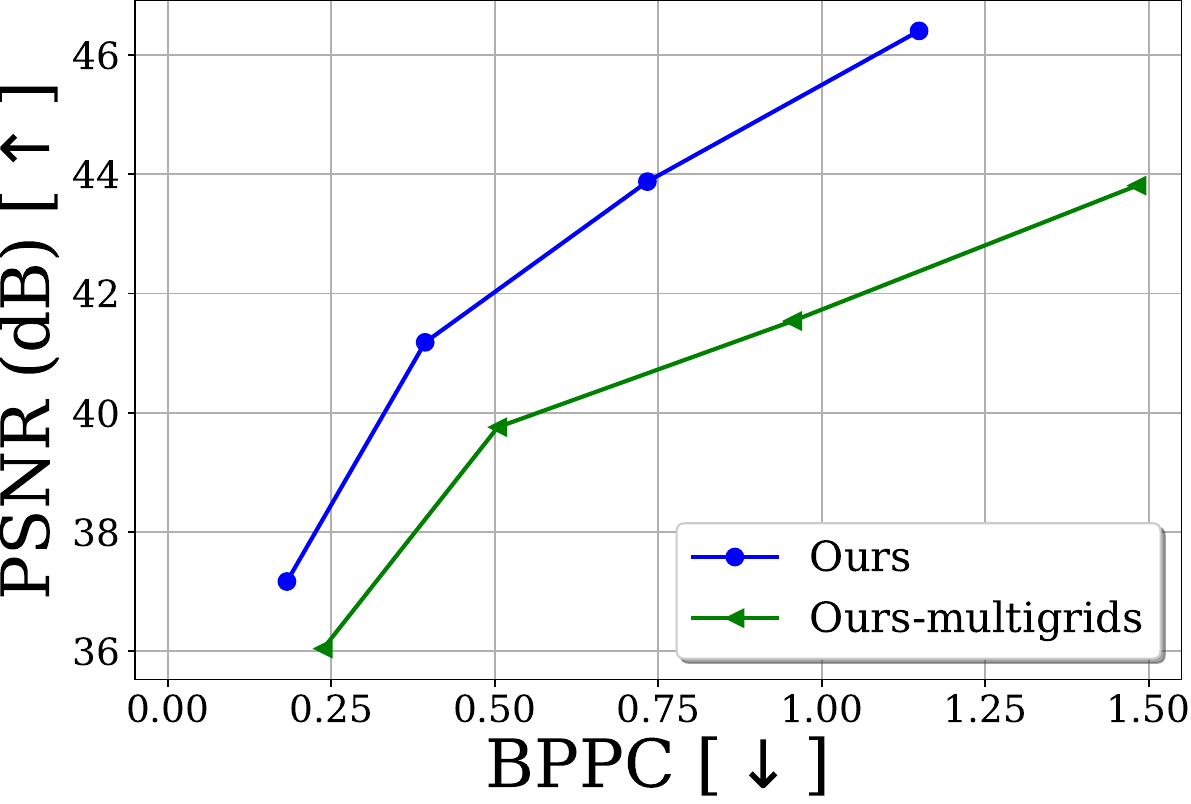}
         \caption{Sampling}
         \label{fig:wo_stride_rd}
     \end{subfigure}
     \hspace{0.1cm}
    \begin{subfigure}[b]{0.31\textwidth}
         \centering         
         \includegraphics[width=1\textwidth]{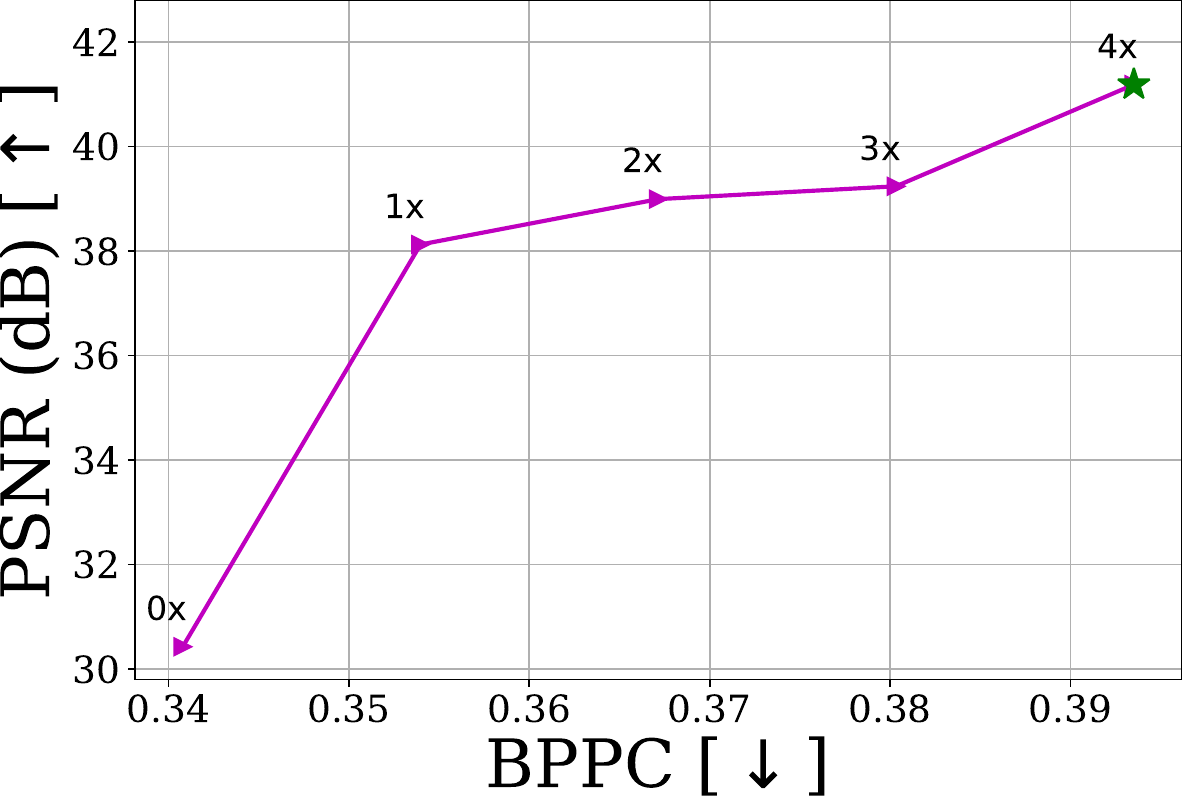}         
         \caption{Depth}
         \label{fig:ablation_dec_depth}
    \end{subfigure}
    \vspace{-0.2cm}
    \caption{Effectiveness of the encoder (a), sampling (b), and depth (c). The curves are evaluated on $2048\times 2048$ resolution ``Ceramic roof 01'' texture set.}
    \vspace{-0.3cm}
\end{figure}

In order to investigate the impact of the depth of the texture synthesizer, we conducted an ablation study. We systematically reduced the number of residual blocks from the original design (which had four blocks) down to zero block. This resulted in a synthesizer that includes only two MLP layers.
Figure~\ref{fig:ablation_dec_depth} illustrates the outcomes of this study. When using only a single residual layer, the BBPC decreased by $8\%$. However, there was a trade-off: the PSNR performance dropped by $11\%$. 
Interestingly, when we completely removed the residual block, the performance suffered significantly. This highlights the critical importance of the residual block in our design.

\subsection{Discussion and Future work}
\label{sec:discussion}
This evaluation has underscored the efficacy of our approach, which leverages multiple levels of redundancy in texture compression. This approach yields a significant advantage over competing methods. However, we acknowledge certain limitations that we aim to address in future work.
Firstly, our current implementation of texture synthesis utilizes a pair of single resolution grids. This necessitates an interpolation step to access in-between mip levels. This rudimentary method may lead to aliasing, as also noted in~\cite{Vaidyanathan2023}. Moving forward, we plan to investigate a data-driven interpolation method. This approach would adaptively address interpolation while preserving overall consistency across different mip levels.
Secondly, the success of our method is closely tied to the robustness of our global transformation, which exploits multi-dimensional spatial-channel-resolution redundancy. In future work, we aim to enhance this by exploring more representative and powerful architectures. These would better capture the multi-level redundancy inherent in the texture, potentially through the use of attention mechanisms, as suggested by~\cite{Zhu2021-so}.

\section{Conclusion}
\label{sec:conc}
This paper introduces an innovative and effective method to tackle the challenge of automating the development of texture compression in photorealistic rendering. The goal is to store textures efficiently, enabling random access and rendering at various resolutions, using learning methods to support all types of texture components.
Our key insight lies in analyzing the redundancy inherent in the problem. Specifically, we identify multiple levels of redundancy: among different channels of a texture, across various resolutions of the same texture, and within individual pixels within each channel. Leveraging these observations, we propose novel techniques for texture compression. Our method begins with a global transformation step that extracts a set of hidden global features~\ref{sec:encoder}. These features are then transformed into dual-bank features through a grid construction process. This process includes a dual-bank projection and an asymmetric quantization step, which adaptively separates the global features into two groups of different frequencies based on the global texture information~\ref{sec:grid_contructor}. Next, our grid-sampler samples these features corresponding to each encoding input position~\ref{sec:grid_sampler}. Finally, we employ a texture synthesis to reconstruct these features back into the pixel domain~\ref{sec:Text_synthesis}. Our results demonstrate that our method achieves state-of-the-art performance, significantly surpassing all conventional texture compression methods. Furthermore, it often outperforms competitive neural compression methods~\cite{Vaidyanathan2023}.

%
%
\bibliographystyle{splncs04}
\bibliography{main.bib}

\appendix
\section{Network architecture}
\label{sec:net}
\subsection{Global transformation}
\label{sec:append_encoder}

\begin{figure}[h!]
    \centering
    \begin{subfigure}[b]{0.7\textwidth}
         \centering
         \includegraphics[width=\linewidth]{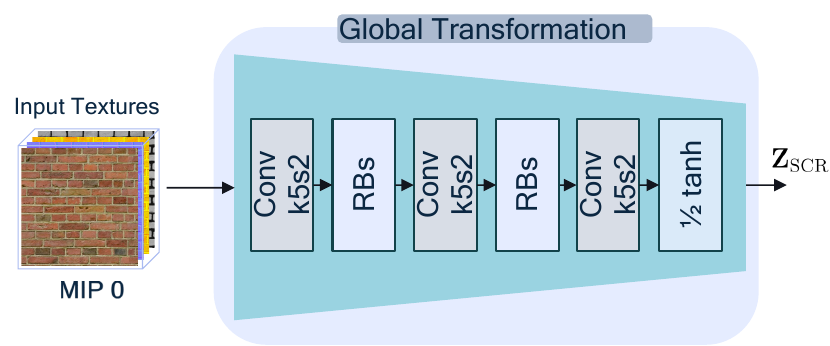}
         \caption{}
         \label{fig:encoder_block}
    \end{subfigure}
    \hspace{0.1cm}
    \begin{subfigure}[b]{0.11\textwidth}
         \centering
         \includegraphics[width=\linewidth]{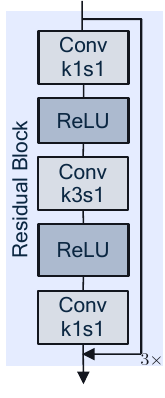}
         \caption{}
         \label{fig:residual_downsample_block}
    \end{subfigure}

    \caption{Overview of Global Transformation architecture. Conv kxsy indicates a convolution with the kernel size x and the stride y. }
    \label{fig:encoder}
\end{figure}

Figure~\ref{fig:encoder} shows the details of global transformation architecture. 
As mentioned in Section~4.1 the global transformer in our setup has the same architecture as~\cite{he2022elic} except that we removed the attention layers and reduced the number of downsampling blocks (Conv k5s2, representing a convolution with the kernel size 5 and the stride 2) from $4$ blocks to $3$ blocks. 
We also added $\frac{1}{2}\tanh$ to bound the output of encoder between $[-0.5,0.5]$. 

\subsection{Texture synthesizer}
\label{sec:append_decoder}
\begin{figure}[h!]
    \begin{subfigure}[b]{0.7\textwidth}
         \centering
         \includegraphics[width=\linewidth]{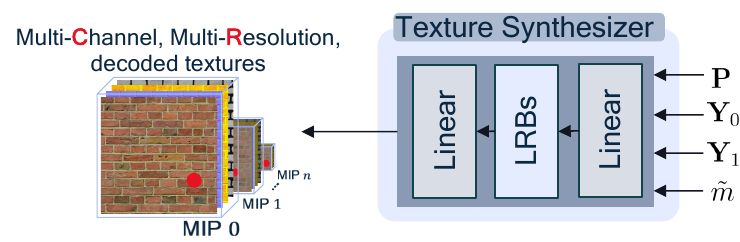}
         \caption{}
         \label{fig:decoder_block}
    \end{subfigure}
    \hspace{1cm}
    \begin{subfigure}[b]{0.13\textwidth}
         \centering
         \includegraphics[width=\linewidth]{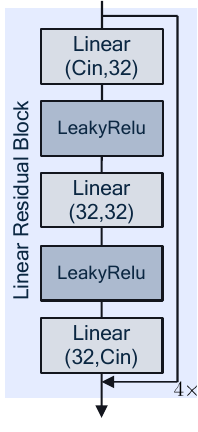}
         \caption{}
         \label{fig:linear_residual_block}
    \end{subfigure}

    \caption{Overview of decoder architecture. We use convention $(c_{in},c_{out})$ for linear layers. }
    \label{fig:decoder}
\end{figure}

Figure~\ref{fig:decoder} shows the architecture of texture synthesize in our compression setup. 

\section{More results}
\label{sec:more_results}

\subsection{Channel wise performance}
Table~\ref{tab:all_textures_diffuse} and Table~\ref{tab:all_textures_normal} present the performance results in terms of PSNR (dB) and BPPC. These results are linked to the diffuse and the normal maps, respectively, of all textures utilized in our evaluation. We compare our method, referred to as CNTC, with NTC~\cite{Vaidyanathan2023}. As demonstrated, in most cases, our method outperforms NTC. 
\begin{table}[t]
    \centering
        \caption{PSNR scores of our re-implementation of \cite{Vaidyanathan2023} versus ours (CNTC) for the diffuse, numbers are  indicating: PSNR [$\uparrow$] (BPPC [$\downarrow$]), where PSNR is calculated for MIP $0$ of the diffuse map. The texture sets at the first three rows are retrieved from PolyHaven (\url{https://polyhaven.com/}) and the rest from ambientCG (\url{https://ambientcg.com/}).}    \vspace{-0.3cm}
    \resizebox{\textwidth}{!}{
    \begin{tabular}{c||c|c||c|c||c|c||c}
    \hline 
    Texture set & NTC 0.2 & CNTC 16 & NTC 0.5 & CNTC 32 & NTC 1.0 & CNTC 64 & Reference \\ \hline\hline
       \includegraphics[width=0.2\textwidth]{Figs/texture_org/pdf/ceramic_roof_01_diff_mip0.pdf} 
       & \includegraphics[width=0.2\textwidth]{Figs/ntc_patch_figs/ntc_0,2_dataset_ceramic_roof_01_res_2k/rec_patch_diff.pdf}
       & \includegraphics[width=0.2\textwidth]{Figs/patch_figs/ceramic_roof_01_cntc_16/rec_patch_diff.pdf} 
       & \includegraphics[width=0.2\textwidth]{Figs/ntc_patch_figs/ntc_0,5_dataset_ceramic_roof_01_res_2k/rec_patch_diff.pdf} 
       & \includegraphics[width=0.2\textwidth]{Figs/patch_figs/ceramic_roof_01_cntc_32/rec_patch_diff.pdf}
       & \includegraphics[width=0.2\textwidth]{Figs/ntc_patch_figs/ntc_1,0_dataset_ceramic_roof_01_res_2k/rec_patch_diff.pdf}
       & \includegraphics[width=0.2\textwidth]{Figs/patch_figs/ceramic_roof_01_cntc_64/rec_patch_diff.pdf}
       & \includegraphics[width=0.2\textwidth]{Figs/patch_figs/ceramic_roof_01_cntc_16/org_patch_diff.pdf}
       \\
        ceramic roof 01 & $35.81$ ($0.17$) & $\mathbf{36.82}$ ($0.18$) & $38.78$ ($0.38$)& $\mathbf{40.53}$ ($0.39$) & $41.29$ ($0.76$)&  $\mathbf{44.04}$ ($0.73$) & Ground Truth \\ \hline
       \includegraphics[width=0.2\textwidth]{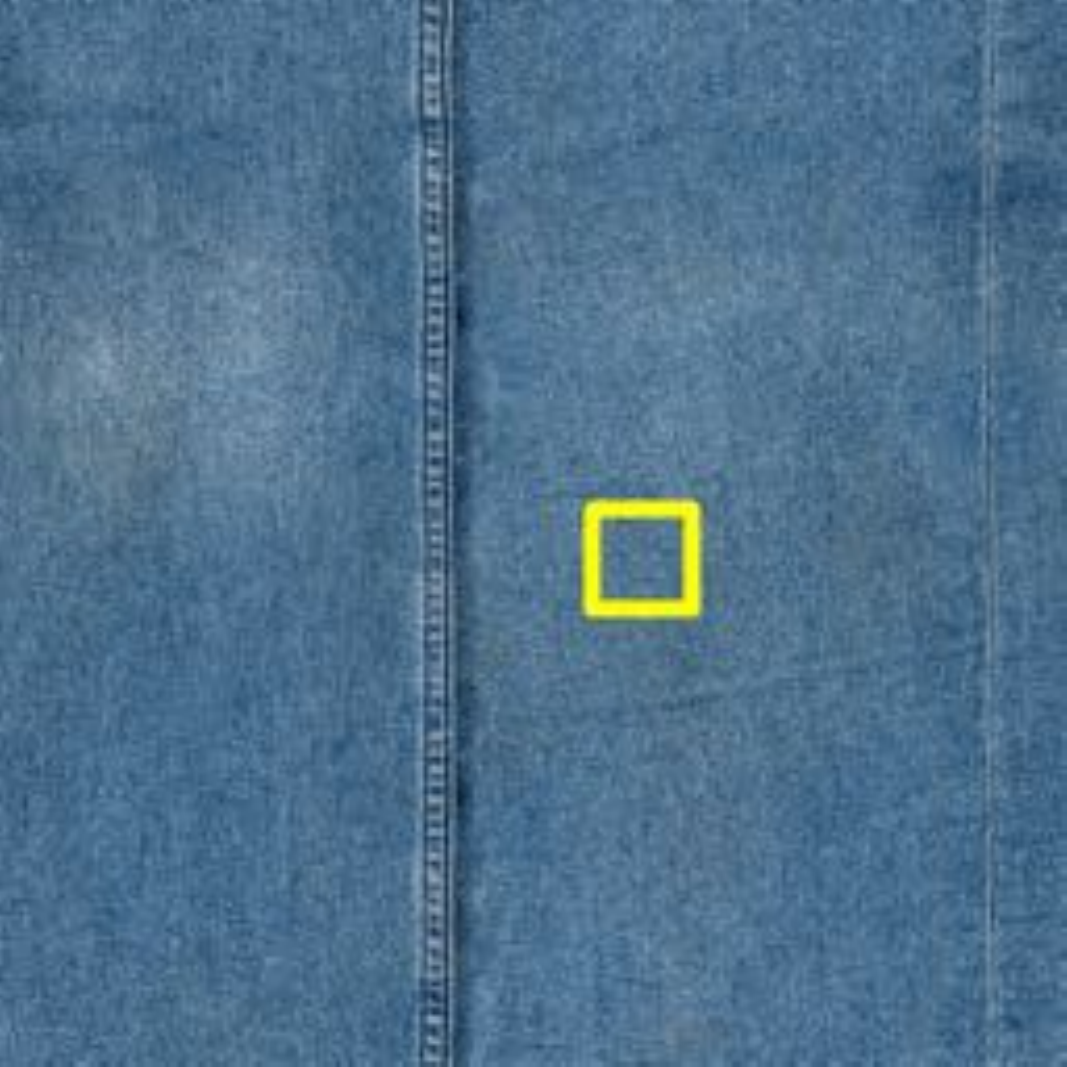} 
       & \includegraphics[width=0.2\textwidth]{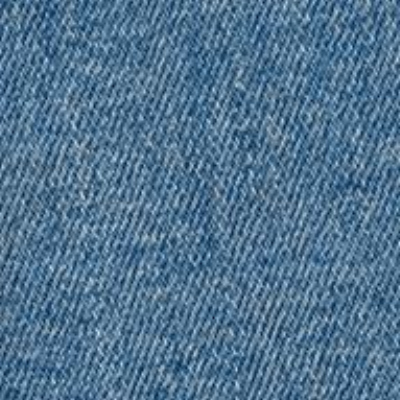}
       & \includegraphics[width=0.2\textwidth]{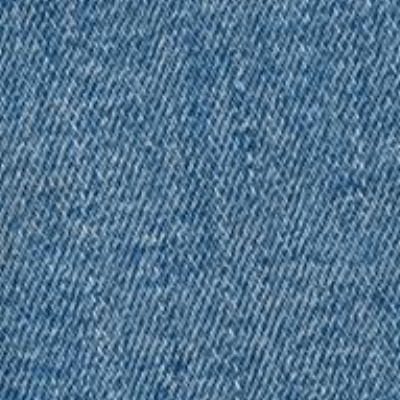} 
       & \includegraphics[width=0.2\textwidth]{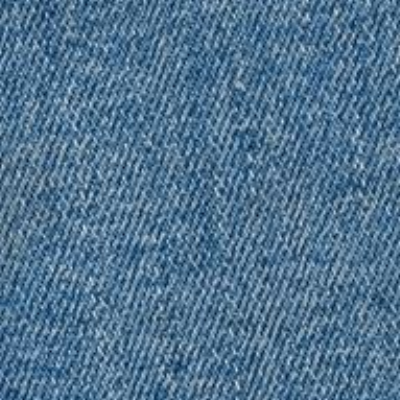} 
       & \includegraphics[width=0.2\textwidth]{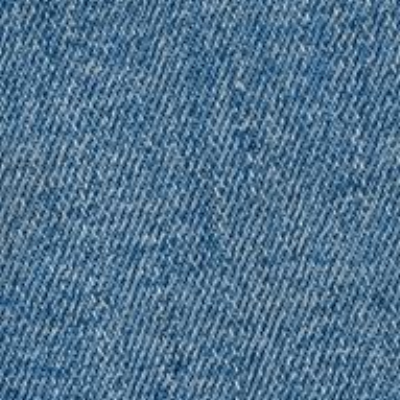}
       & \includegraphics[width=0.2\textwidth]{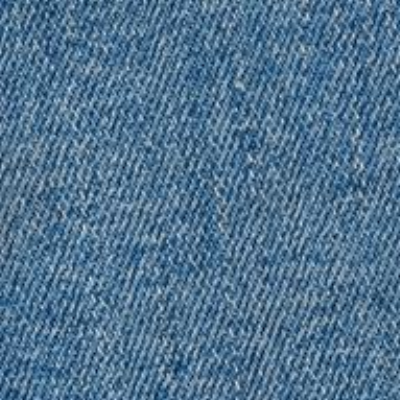}
       & \includegraphics[width=0.2\textwidth]{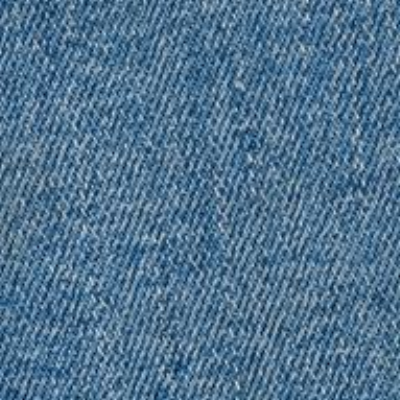}
       & \includegraphics[width=0.2\textwidth]{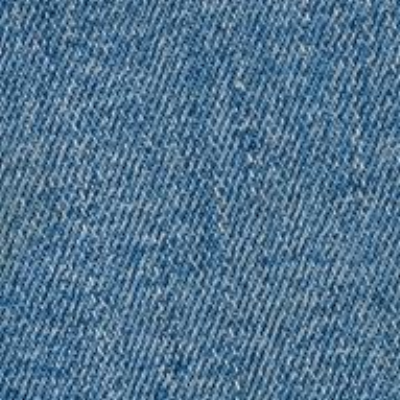}
       \\
        denim fabric & $26.49$ ($0.17$) & $\mathbf{27.23}$ ($0.18$) & $34.38$ ($0.38$)& $\mathbf{35.30}$ ($0.39$) & $35.14$ ($0.76$)&  $\mathbf{38.94}$ ($0.73$) & Ground Truth \\ \hline
       \includegraphics[width=0.2\textwidth]{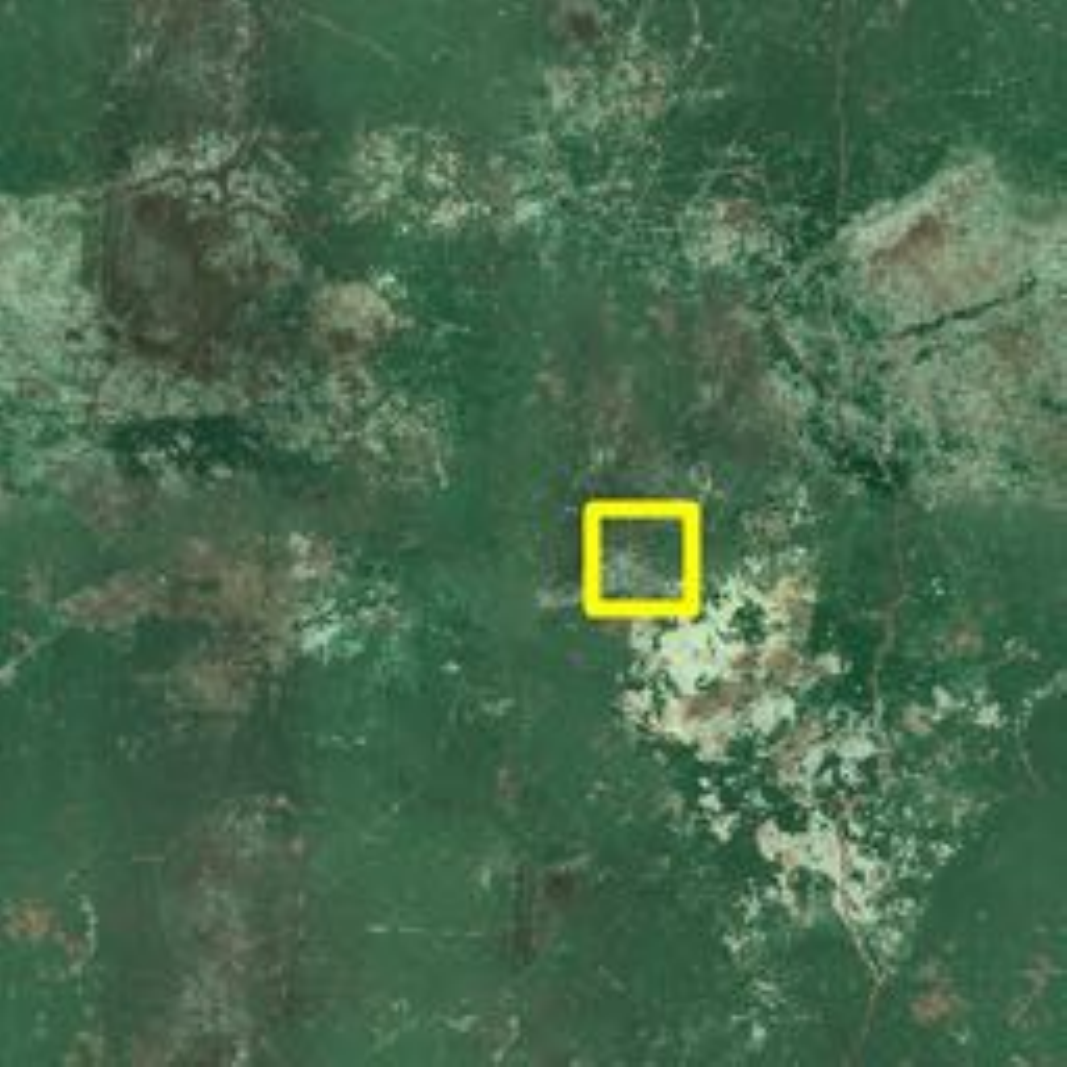} 
       & \includegraphics[width=0.2\textwidth]{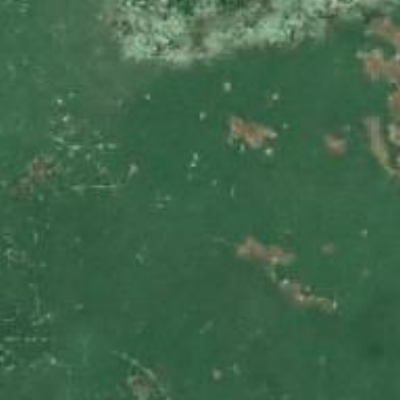}
       & \includegraphics[width=0.2\textwidth]{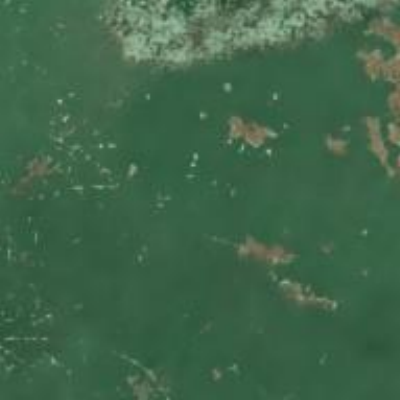} 
       & \includegraphics[width=0.2\textwidth]{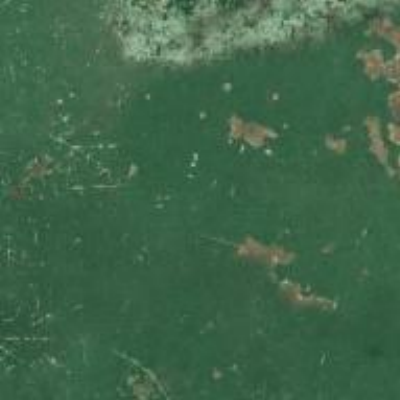} 
       & \includegraphics[width=0.2\textwidth]{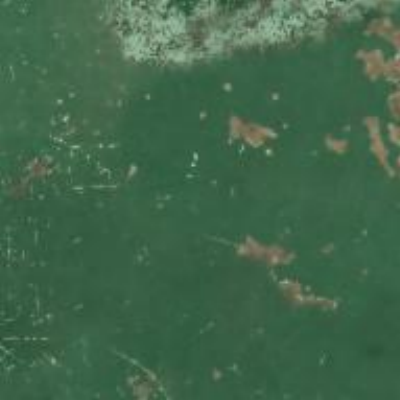}
       & \includegraphics[width=0.2\textwidth]{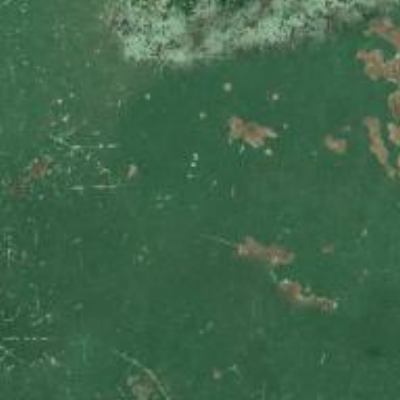}
       & \includegraphics[width=0.2\textwidth]{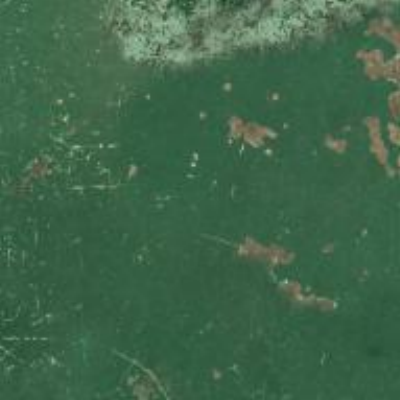}
       & \includegraphics[width=0.2\textwidth]{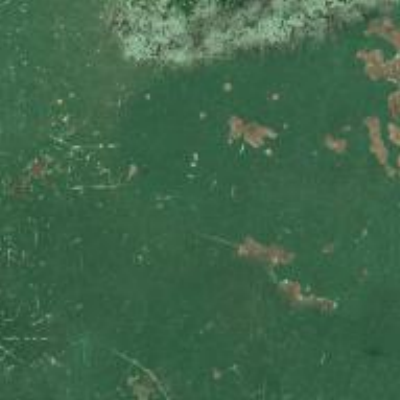}
       \\
        painted concrete & $28.62$ ($0.17$) & $\mathbf{29.69}$ ($0.18$) & $31.55$ ($0.38$)& $\mathbf{32.54}$ ($0.39$) & $36.48$ ($0.76$)&  $\mathbf{37.34}$ ($0.73$) & Ground Truth \\ \hline \hline
       \includegraphics[width=0.2\textwidth]{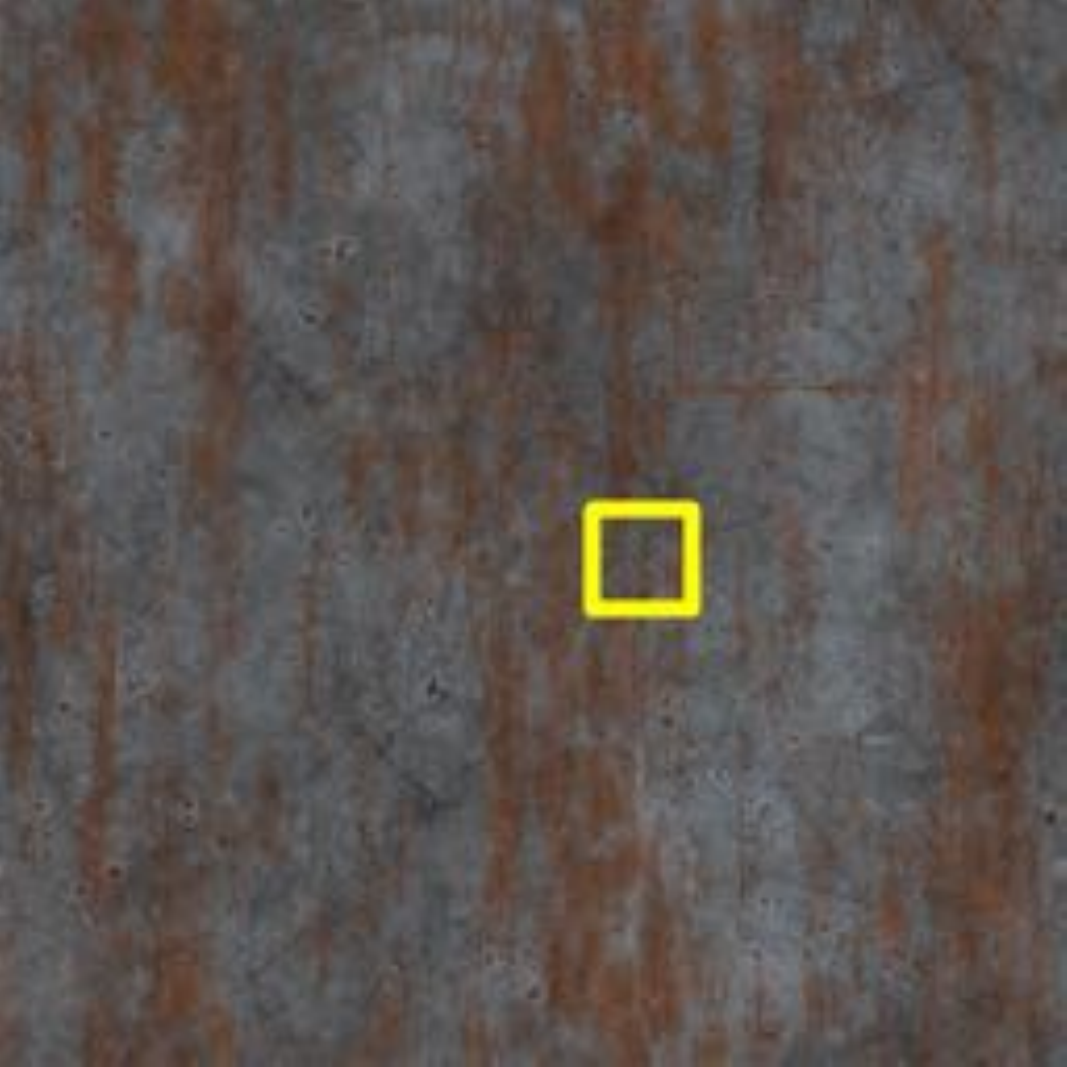} 
       & \includegraphics[width=0.2\textwidth]{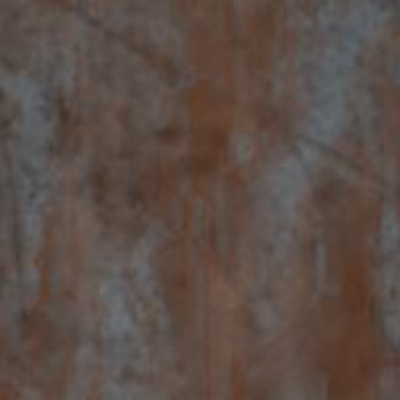}
       & \includegraphics[width=0.2\textwidth]{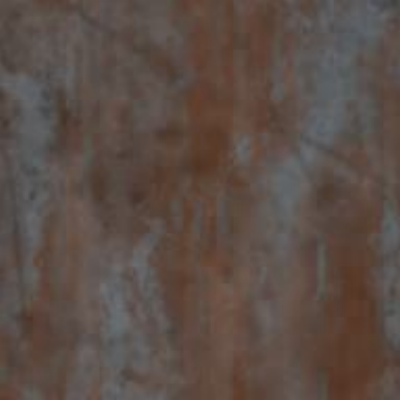} 
       & \includegraphics[width=0.2\textwidth]{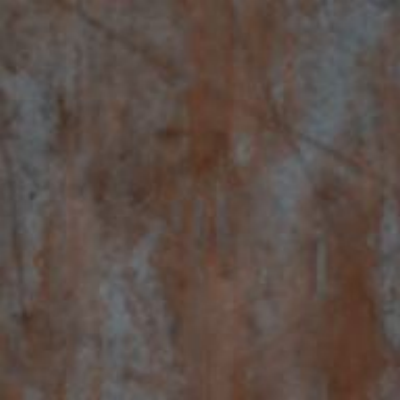} 
       & \includegraphics[width=0.2\textwidth]{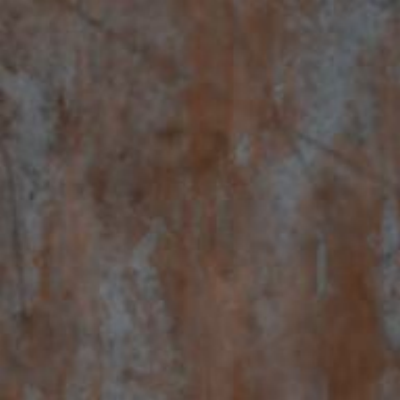}
       & \includegraphics[width=0.2\textwidth]{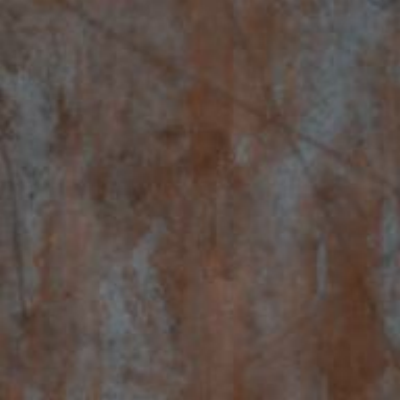}
       & \includegraphics[width=0.2\textwidth]{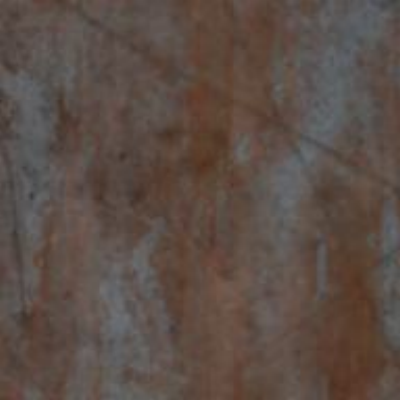}
       & \includegraphics[width=0.2\textwidth]{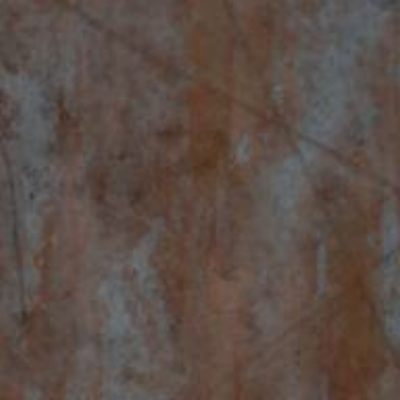}
       \\
        metal plates 013 & $37.77$ ($0.17$) & $\mathbf{37.98}$ ($0.18$) & $39.71$ ($0.38$)& $\mathbf{40.85}$ ($0.39$) & $42.63$ ($0.76$)&  $\mathbf{43.23}$ ($0.73$) & Ground Truth \\ \hline
       \includegraphics[width=0.2\textwidth]{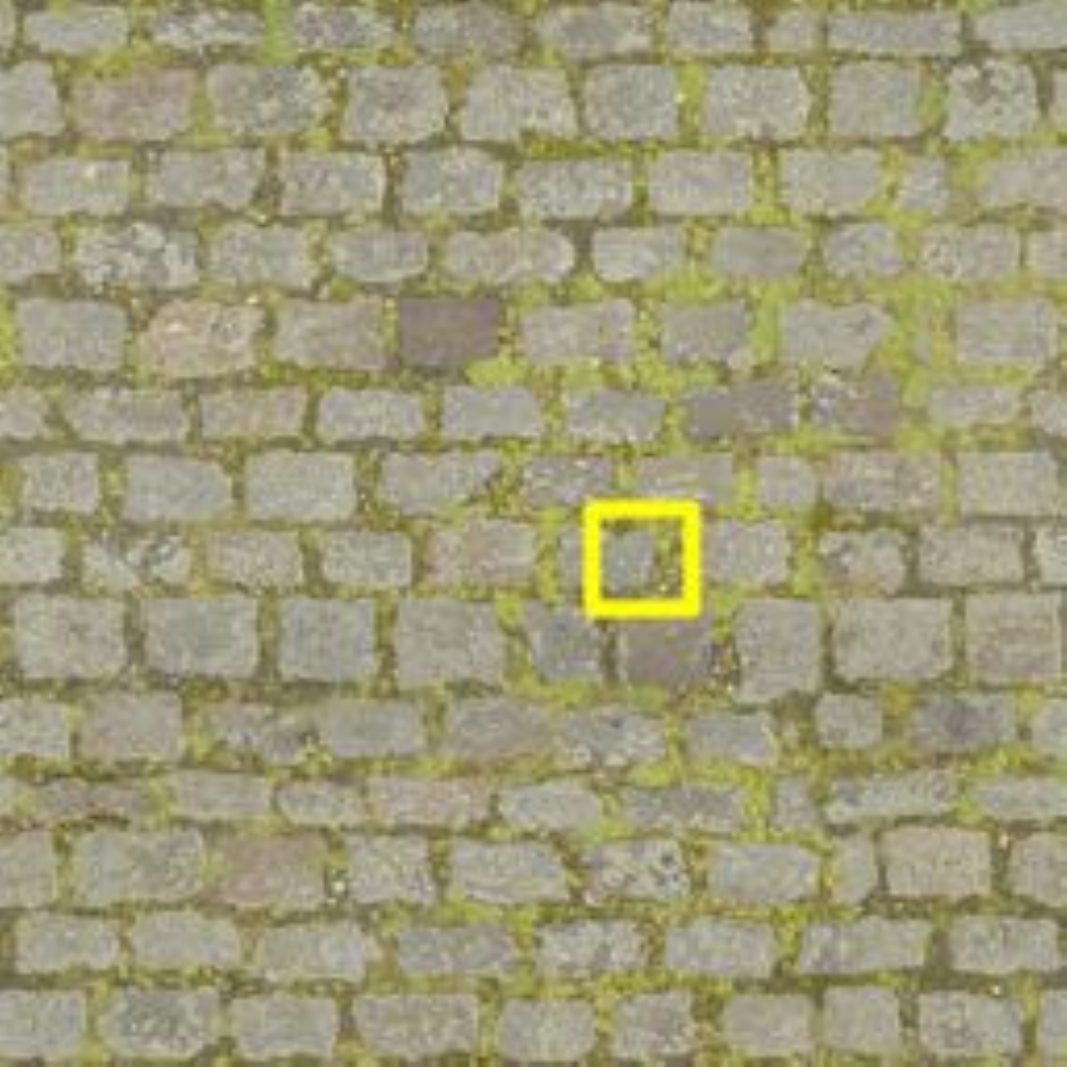} 
       & \includegraphics[width=0.2\textwidth]{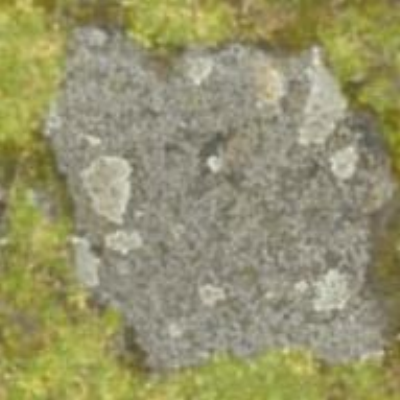}
       & \includegraphics[width=0.2\textwidth]{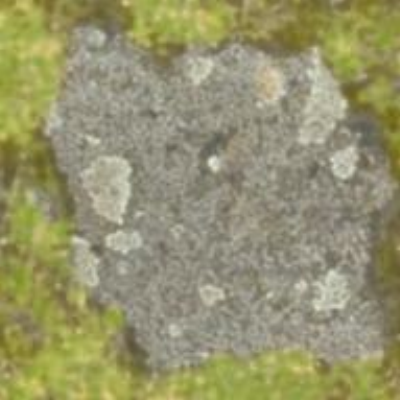} 
       & \includegraphics[width=0.2\textwidth]{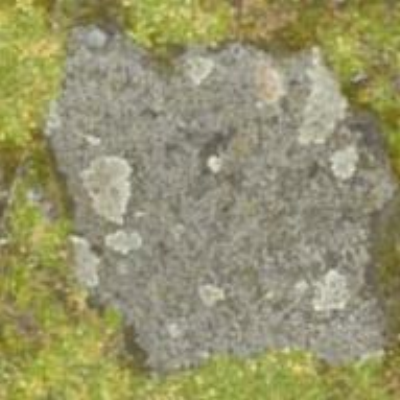} 
       & \includegraphics[width=0.2\textwidth]{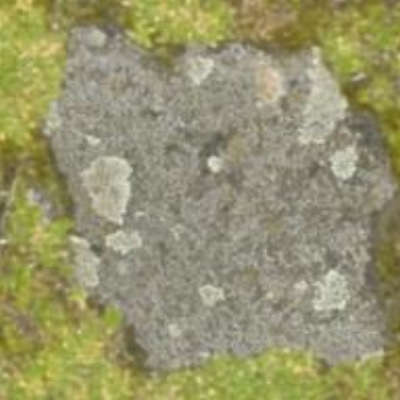}
       & \includegraphics[width=0.2\textwidth]{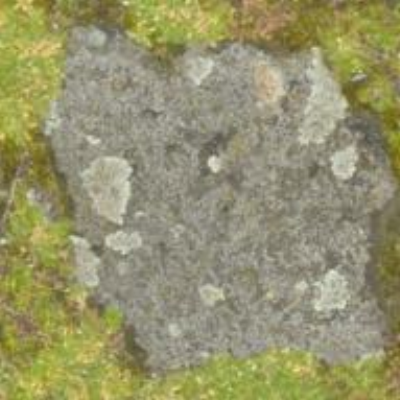}
       & \includegraphics[width=0.2\textwidth]{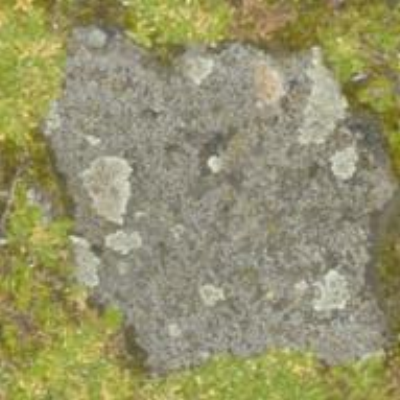}
       & \includegraphics[width=0.2\textwidth]{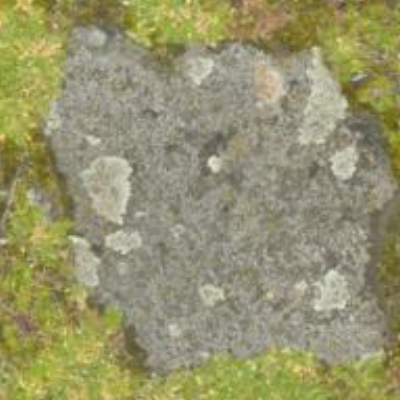}
       \\
        paving stones 131 & $28.40$ ($0.17$) & $\mathbf{28.54}$ ($0.18$) & $30.63$ ($0.38$)& $\mathbf{31.28}$ ($0.39$) & $\mathbf{35.28}$ ($0.76$)&  $34.85$ ($0.73$) & Ground Truth \\ \hline
       \includegraphics[width=0.2\textwidth]{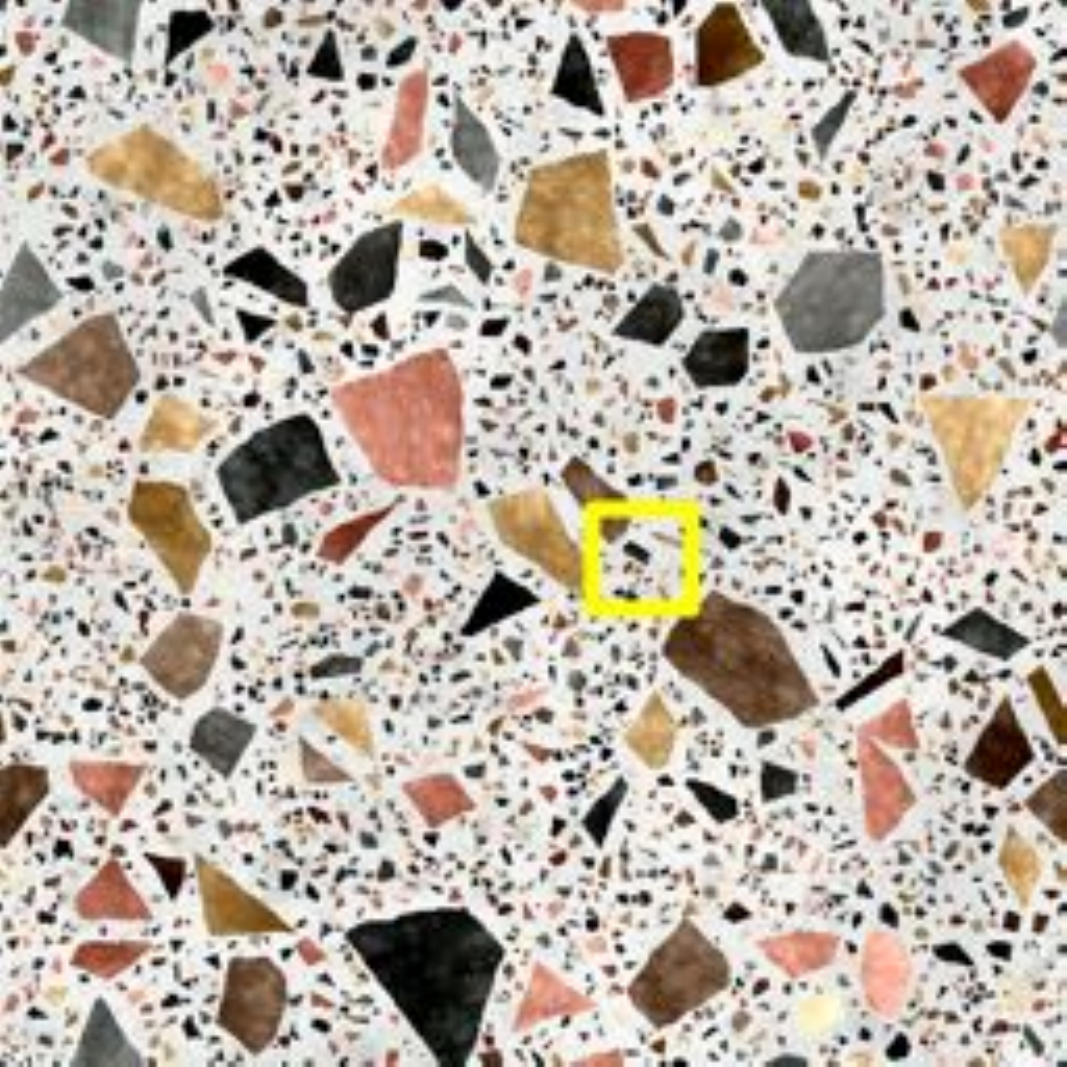} 
       & \includegraphics[width=0.2\textwidth]{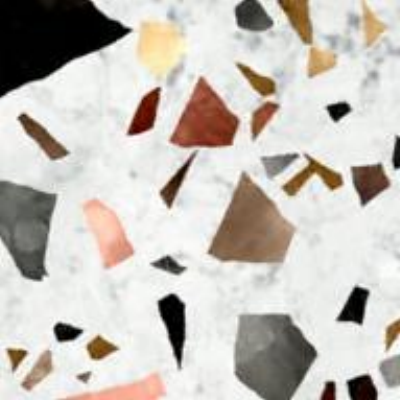}
       & \includegraphics[width=0.2\textwidth]{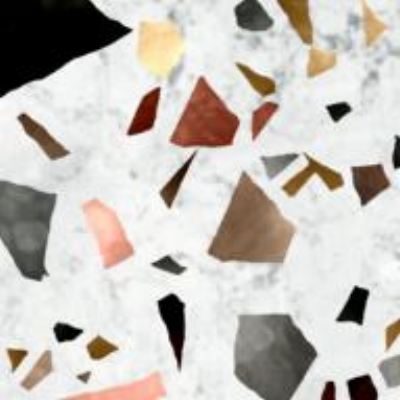} 
       & \includegraphics[width=0.2\textwidth]{Figs/ntc_patch_figs/ntc_0,2_dataset_terrazzo_res_2k/rec_patch_Color.pdf} 
       & \includegraphics[width=0.2\textwidth]{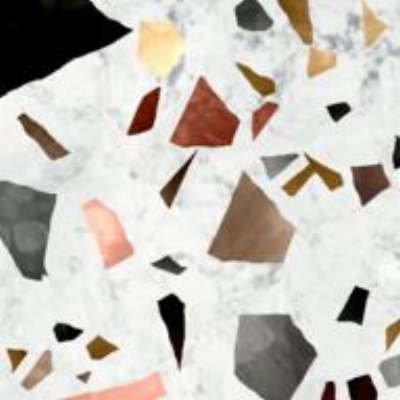}
       & \includegraphics[width=0.2\textwidth]{Figs/ntc_patch_figs/ntc_0,2_dataset_terrazzo_res_2k/rec_patch_Color.pdf}
       & \includegraphics[width=0.2\textwidth]{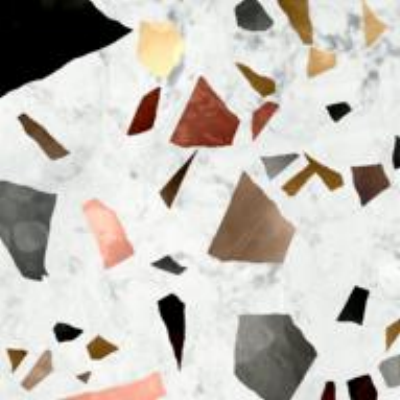}
       & \includegraphics[width=0.2\textwidth]{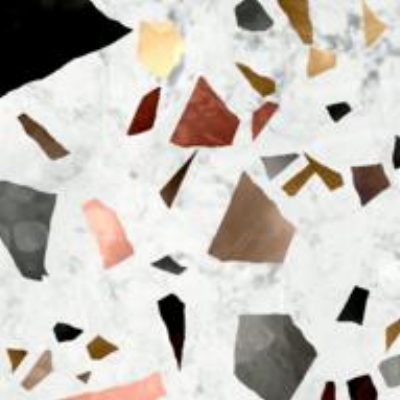}
       \\
        terrazzo 018 & $33.89$ ($0.17$) & $\mathbf{38.43}$ ($0.20$) & $38.07$ ($0.38$)& $\mathbf{42.54}$ ($0.44$) & $41.77$ ($0.76$)&  $\mathbf{46.52}$ ($0.82$) & Ground Truth \\ \hline
       \includegraphics[width=0.2\textwidth]{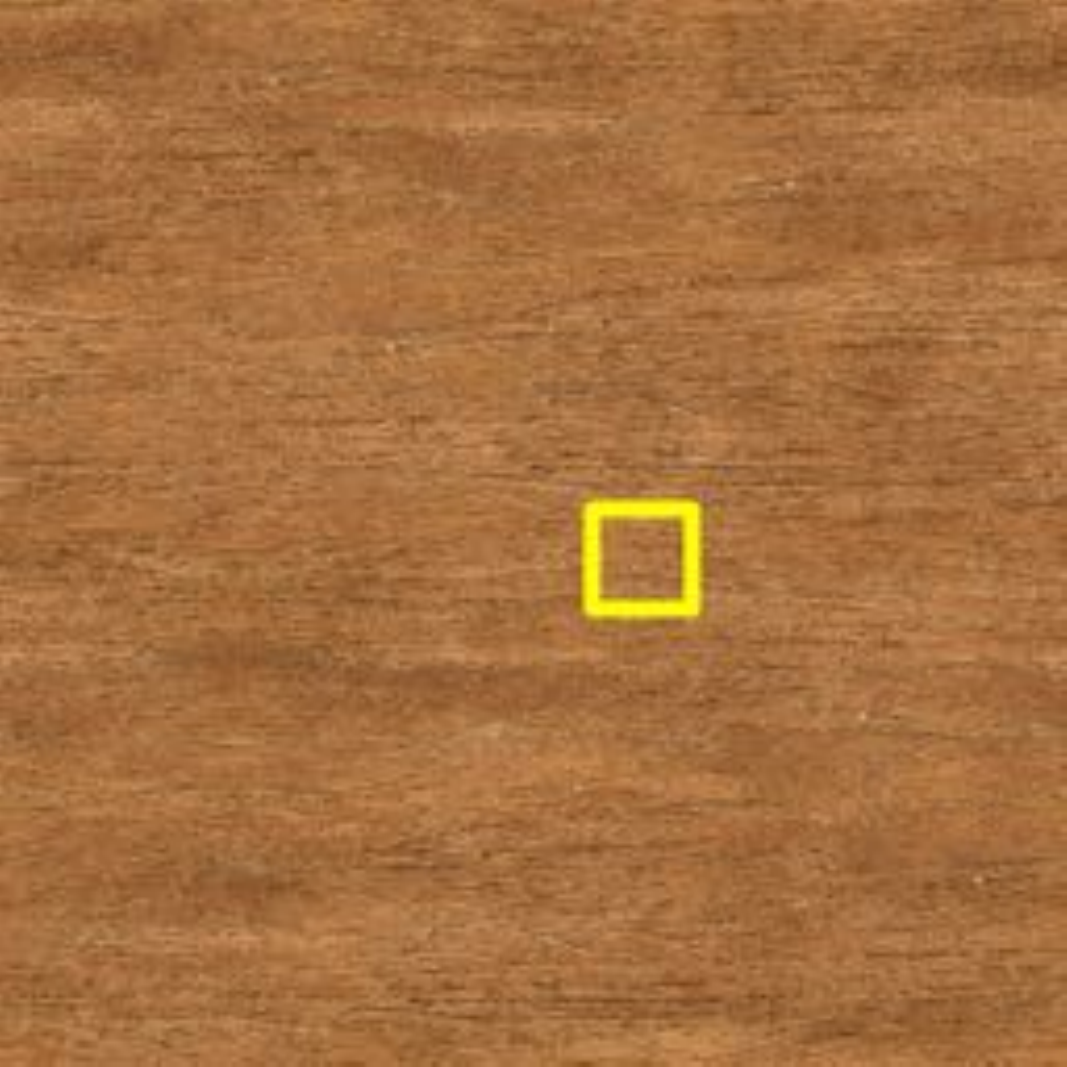} 
       & \includegraphics[width=0.2\textwidth]{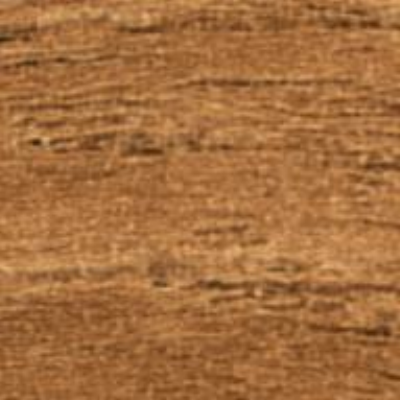}
       & \includegraphics[width=0.2\textwidth]{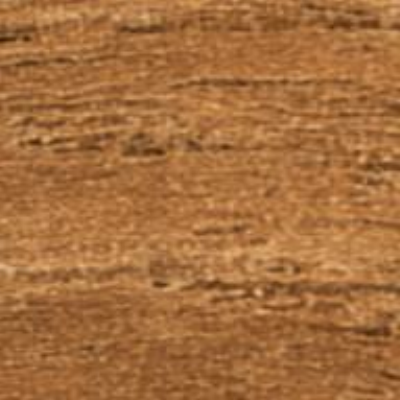} 
       & \includegraphics[width=0.2\textwidth]{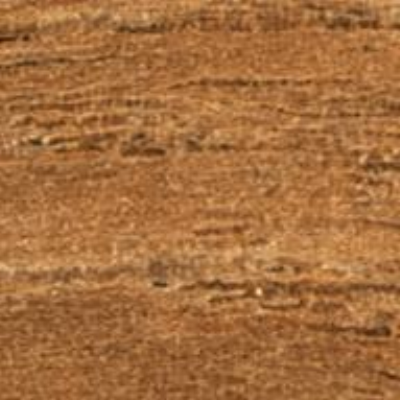} 
       & \includegraphics[width=0.2\textwidth]{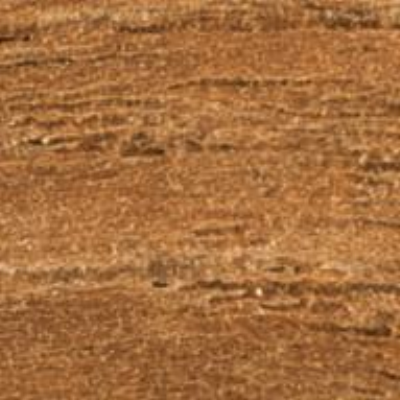}
       & \includegraphics[width=0.2\textwidth]{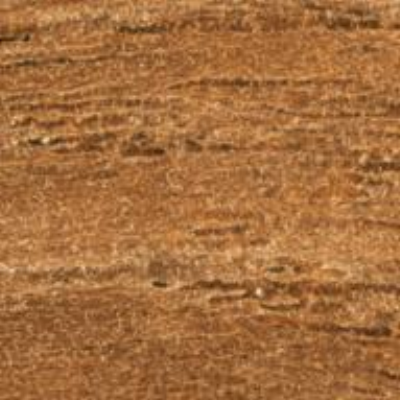}
       & \includegraphics[width=0.2\textwidth]{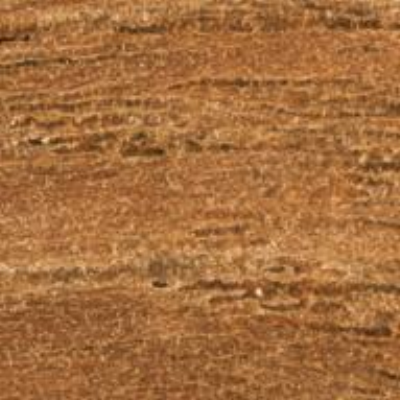}
       & \includegraphics[width=0.2\textwidth]{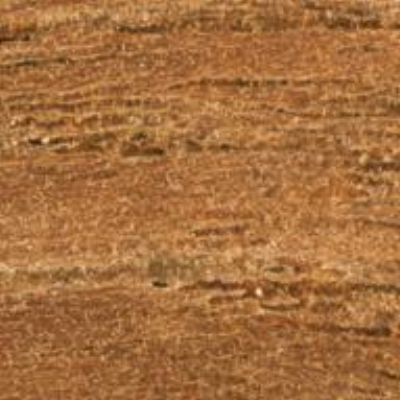}
       \\
        wood 063 & $27.28$ ($0.17$) & $\mathbf{27.77}$ ($0.18$) & $30.37$ ($0.38$)& $\mathbf{31.98}$ ($0.39$) & $35.08$ ($0.76$)&  $\mathbf{37.23}$ ($0.73$) & Ground Truth \\ \hline
     \end{tabular}
     }
     \label{tab:all_textures_diffuse}
\end{table}

\begin{table}[t!]
    \centering
        \caption{PSNR scores of our re-implementation of \cite{Vaidyanathan2023} vresus ours (CNTC) for the normal map, numbers are presenting: PSNR [$\uparrow$] (BPPC [$\downarrow$]), where PSNR is calculated for MIP $0$ of the normal map. The texture sets at the first three rows are retrieved from PolyHaven \url{https://polyhaven.com/}) and the rest from ambientCG (\url{https://ambientcg.com/}).}    \vspace{-0.3cm}
    \resizebox{\textwidth}{!}{
    \begin{tabular}{c||c|c||c|c||c|c||c}
    \hline 
    Texture set & NTC 0.2 & CNTC 16 & NTC 0.5 & CNTC 32 & NTC 1.0 & CNTC 64 & Reference \\ \hline\hline
       \includegraphics[width=0.2\textwidth]{Figs/texture_org/pdf/ceramic_roof_01_gl_mip0.pdf} 
       & \includegraphics[width=0.2\textwidth]{Figs/ntc_patch_figs/ntc_0,2_dataset_ceramic_roof_01_res_2k/rec_patch_gl.pdf}
       & \includegraphics[width=0.2\textwidth]{Figs/patch_figs/ceramic_roof_01_cntc_16/rec_patch_gl.pdf} 
       & \includegraphics[width=0.2\textwidth]{Figs/ntc_patch_figs/ntc_0,5_dataset_ceramic_roof_01_res_2k/rec_patch_gl.pdf} 
       & \includegraphics[width=0.2\textwidth]{Figs/patch_figs/ceramic_roof_01_cntc_32/rec_patch_gl.pdf}
       & \includegraphics[width=0.2\textwidth]{Figs/ntc_patch_figs/ntc_1,0_dataset_ceramic_roof_01_res_2k/rec_patch_gl.pdf}
       & \includegraphics[width=0.2\textwidth]{Figs/patch_figs/ceramic_roof_01_cntc_64/rec_patch_gl.pdf}
       & \includegraphics[width=0.2\textwidth]{Figs/patch_figs/ceramic_roof_01_cntc_16/org_patch_gl.pdf}
       \\
        ceramic roof 01 & $33.64$ ($0.17$) & $\mathbf{35.15}$ ($0.18$) & $36.242$ ($0.38$) & $\mathbf{39.85}$ $(0.39)$ & $39.32$ ($0.76$) & $\mathbf{42.66}$ ($0.73$) & Ground Truth \\ \hline
       \includegraphics[width=0.2\textwidth]{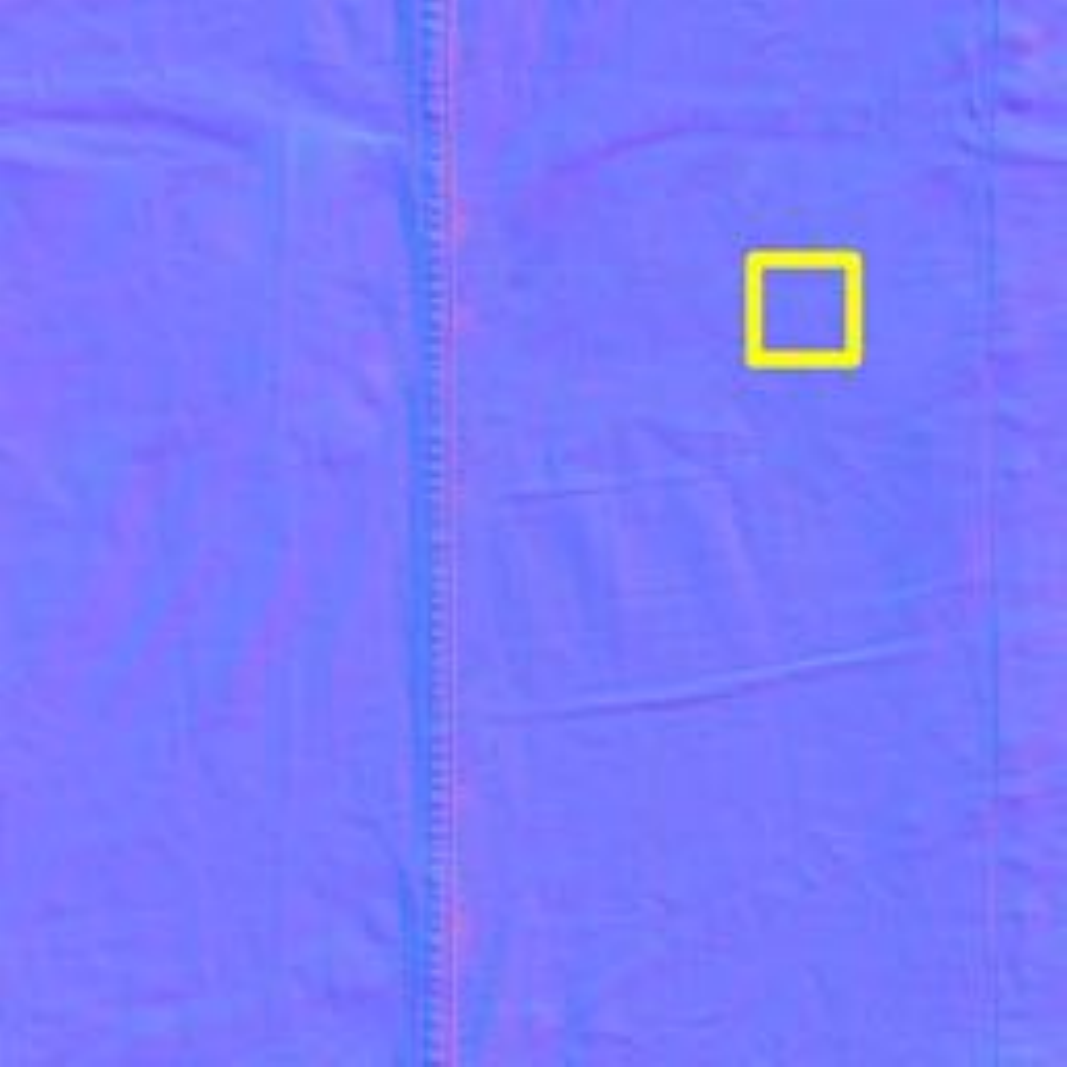} 
       & \includegraphics[width=0.2\textwidth]{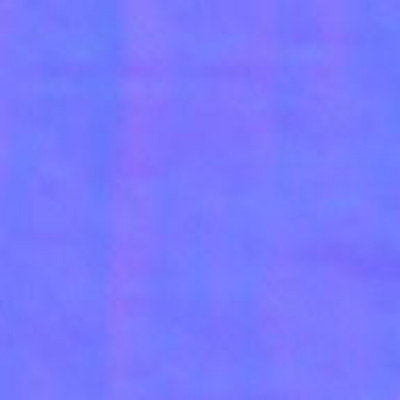}
       & \includegraphics[width=0.2\textwidth]{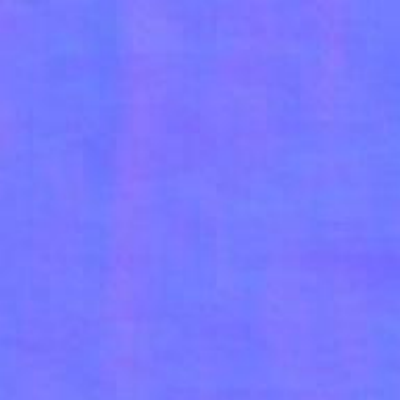} 
       & \includegraphics[width=0.2\textwidth]{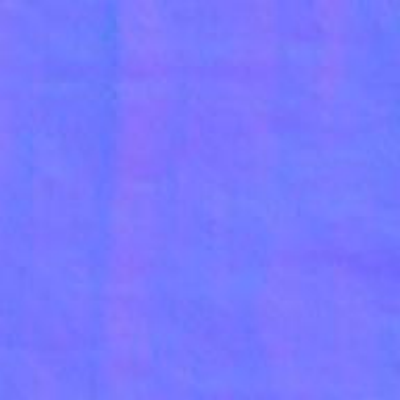} 
       & \includegraphics[width=0.2\textwidth]{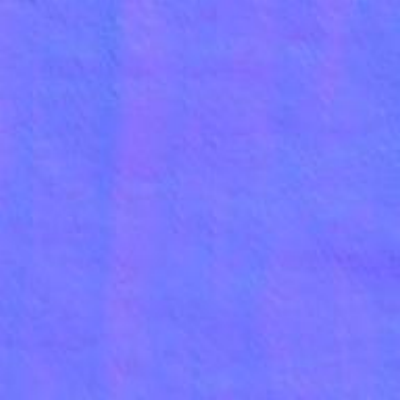}
       & \includegraphics[width=0.2\textwidth]{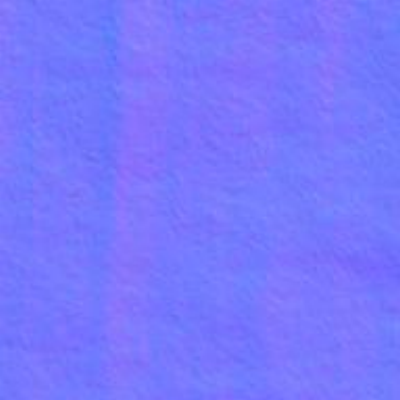}
       & \includegraphics[width=0.2\textwidth]{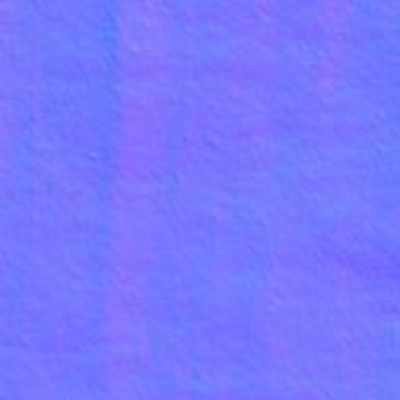}
       & \includegraphics[width=0.2\textwidth]{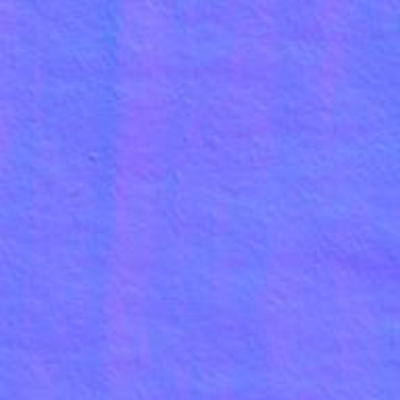}
       \\
        denim fabric & $\mathbf{34.51}$ ($0.17$) & $34.23$ ($0.18$) & $35.20$ ($0.38$)& $\mathbf{36.53}$ ($0.39$) & $39.49$ ($0.76$)&  $\mathbf{39.51}$ ($0.73$) & Ground Truth \\ \hline
       \includegraphics[width=0.2\textwidth]{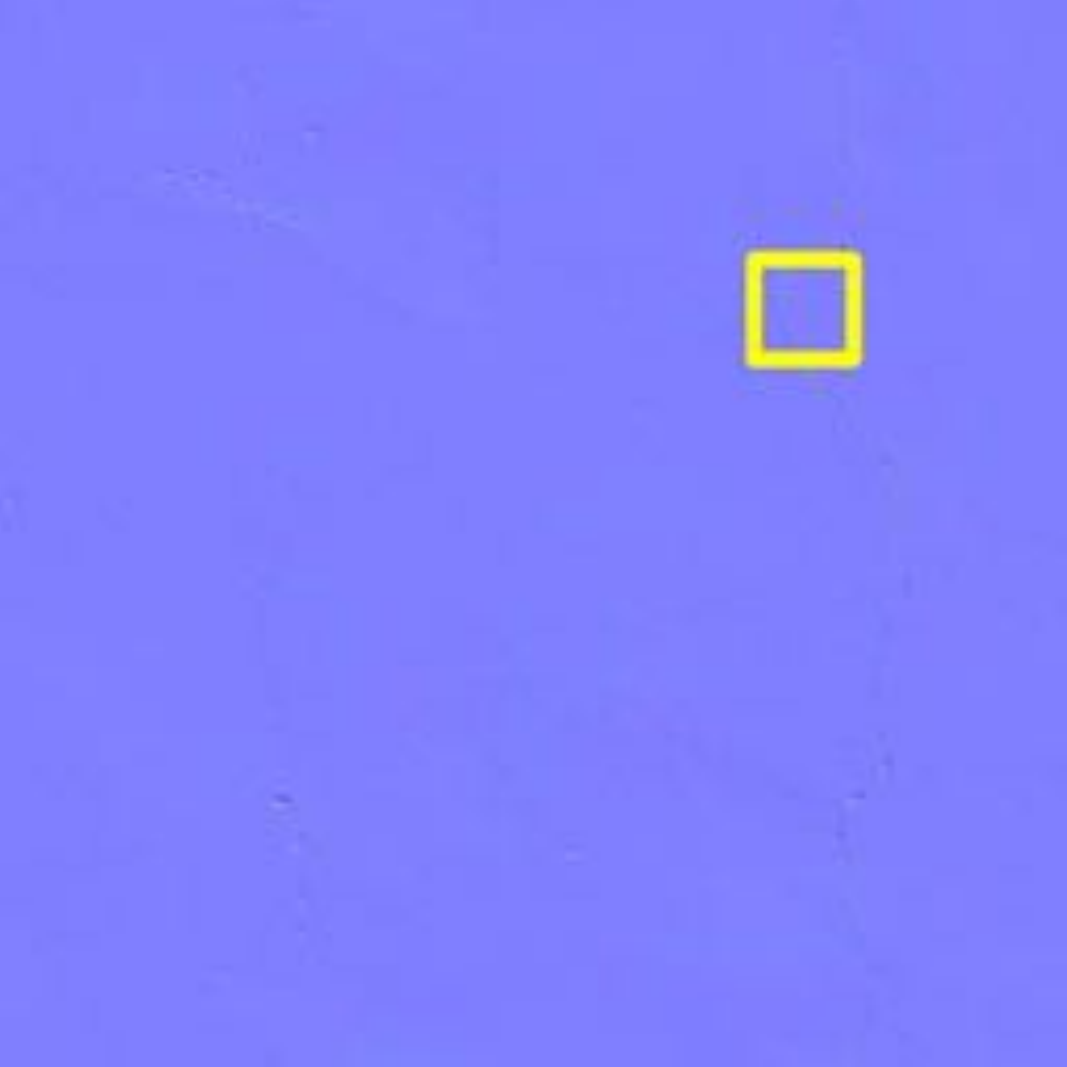} 
       & \includegraphics[width=0.2\textwidth]{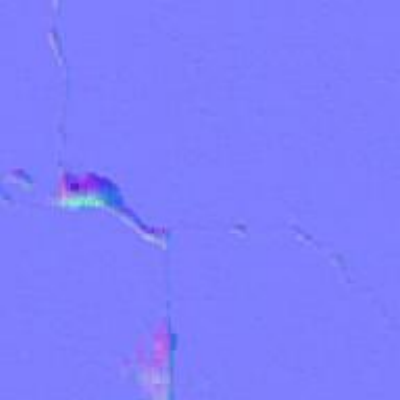}
       & \includegraphics[width=0.2\textwidth]{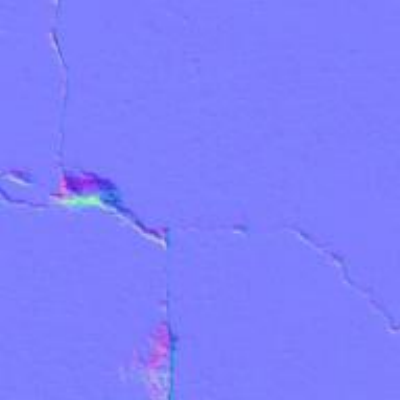} 
       & \includegraphics[width=0.2\textwidth]{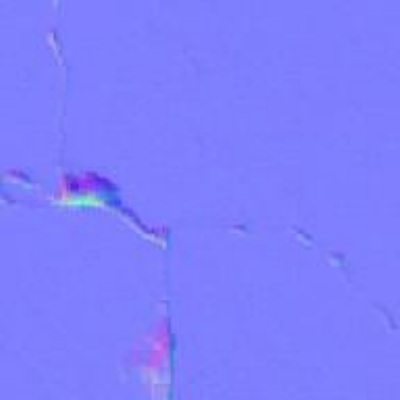} 
       & \includegraphics[width=0.2\textwidth]{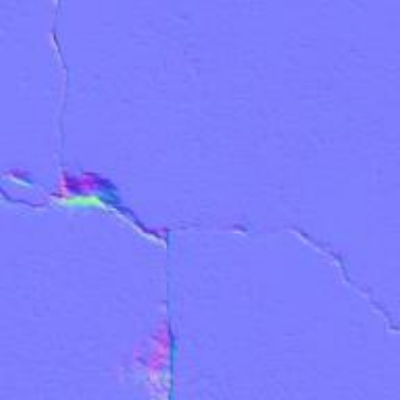}
       & \includegraphics[width=0.2\textwidth]{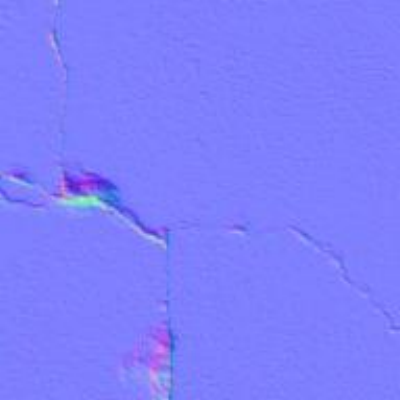}
       & \includegraphics[width=0.2\textwidth]{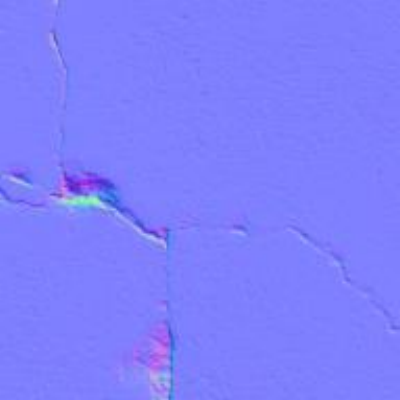}
       & \includegraphics[width=0.2\textwidth]{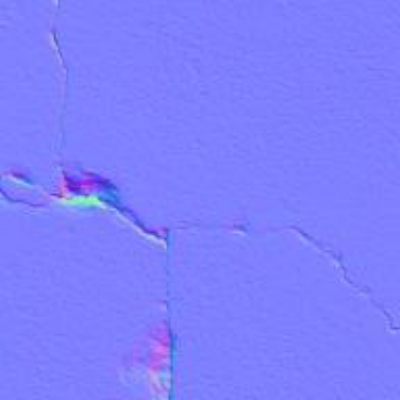}
       \\
        painted concrete & $34.58$ ($0.17$) & $\mathbf{35.43}$ ($0.18$) & $34.81$ ($0.38$)& $\mathbf{36.56}$ ($0.39$) & $\mathbf{38.77}$ ($0.76$)&  $37.64$ ($0.73$) & Ground Truth \\ \hline \hline
       \includegraphics[width=0.2\textwidth]{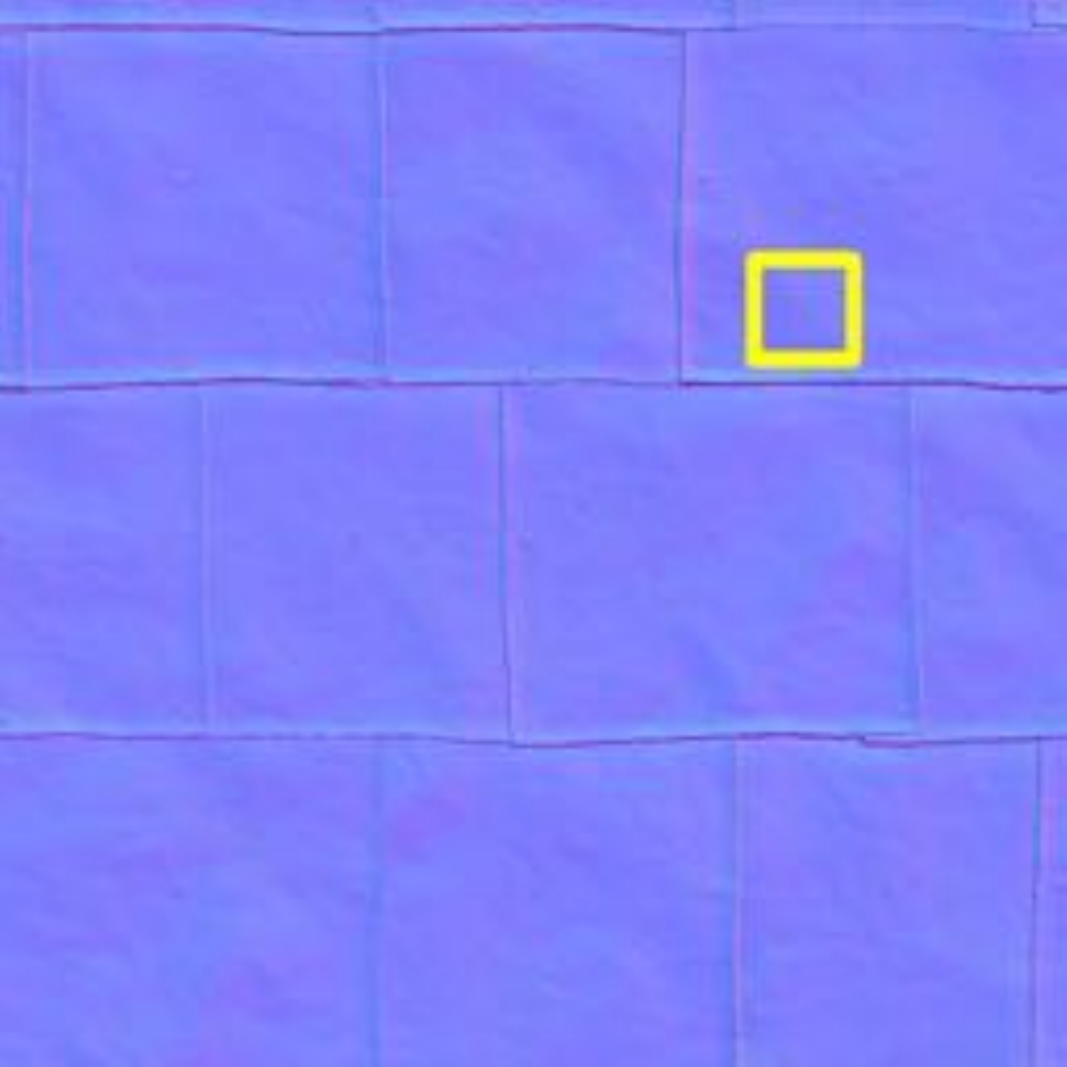} 
       & \includegraphics[width=0.2\textwidth]{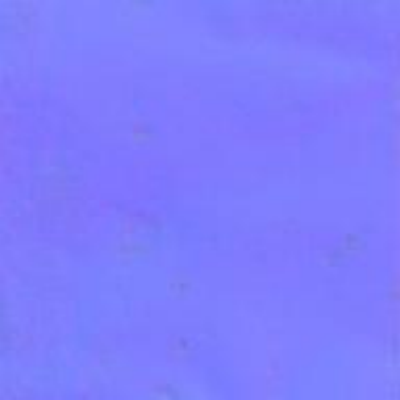}
       & \includegraphics[width=0.2\textwidth]{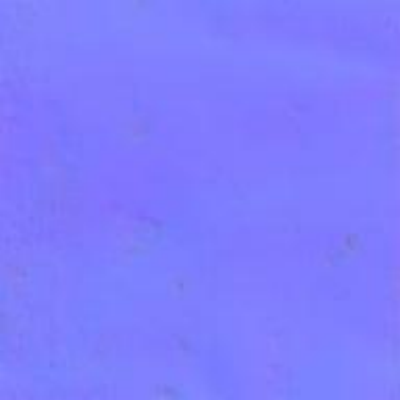} 
       & \includegraphics[width=0.2\textwidth]{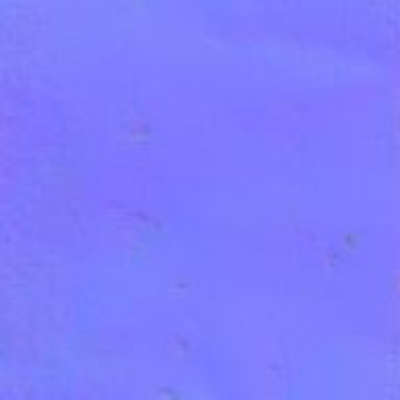} 
       & \includegraphics[width=0.2\textwidth]{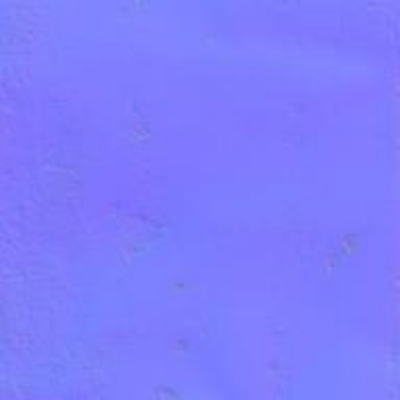}
       & \includegraphics[width=0.2\textwidth]{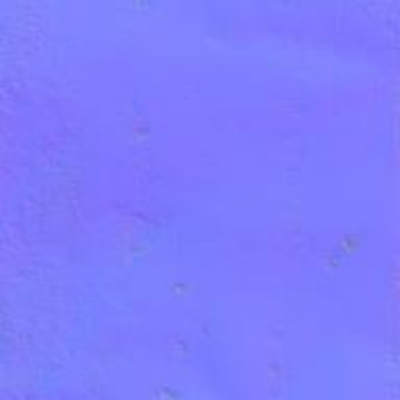}
       & \includegraphics[width=0.2\textwidth]{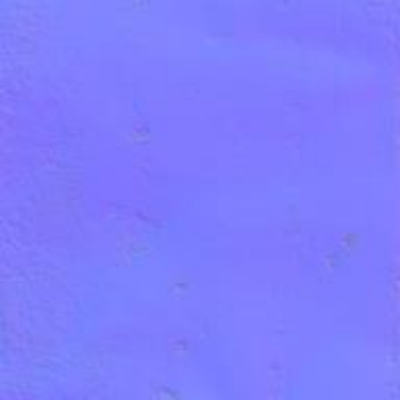}
       & \includegraphics[width=0.2\textwidth]{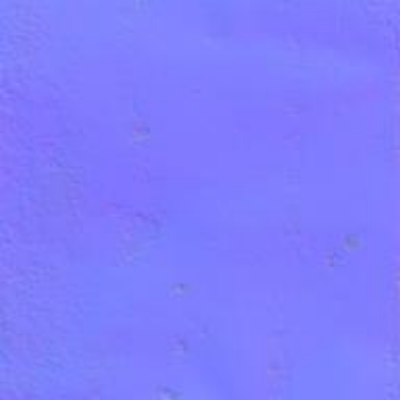}
       \\
        metal plates 013 & $36.68$ ($0.17$) & $\mathbf{37.88}$ ($0.18$) & $38.78$ ($0.38$)& $\mathbf{41.07}$ ($0.39$) & $41.29$ ($0.76$)&  $\mathbf{42.19}$ ($0.73$) & Ground Truth \\ \hline
       \includegraphics[width=0.2\textwidth]{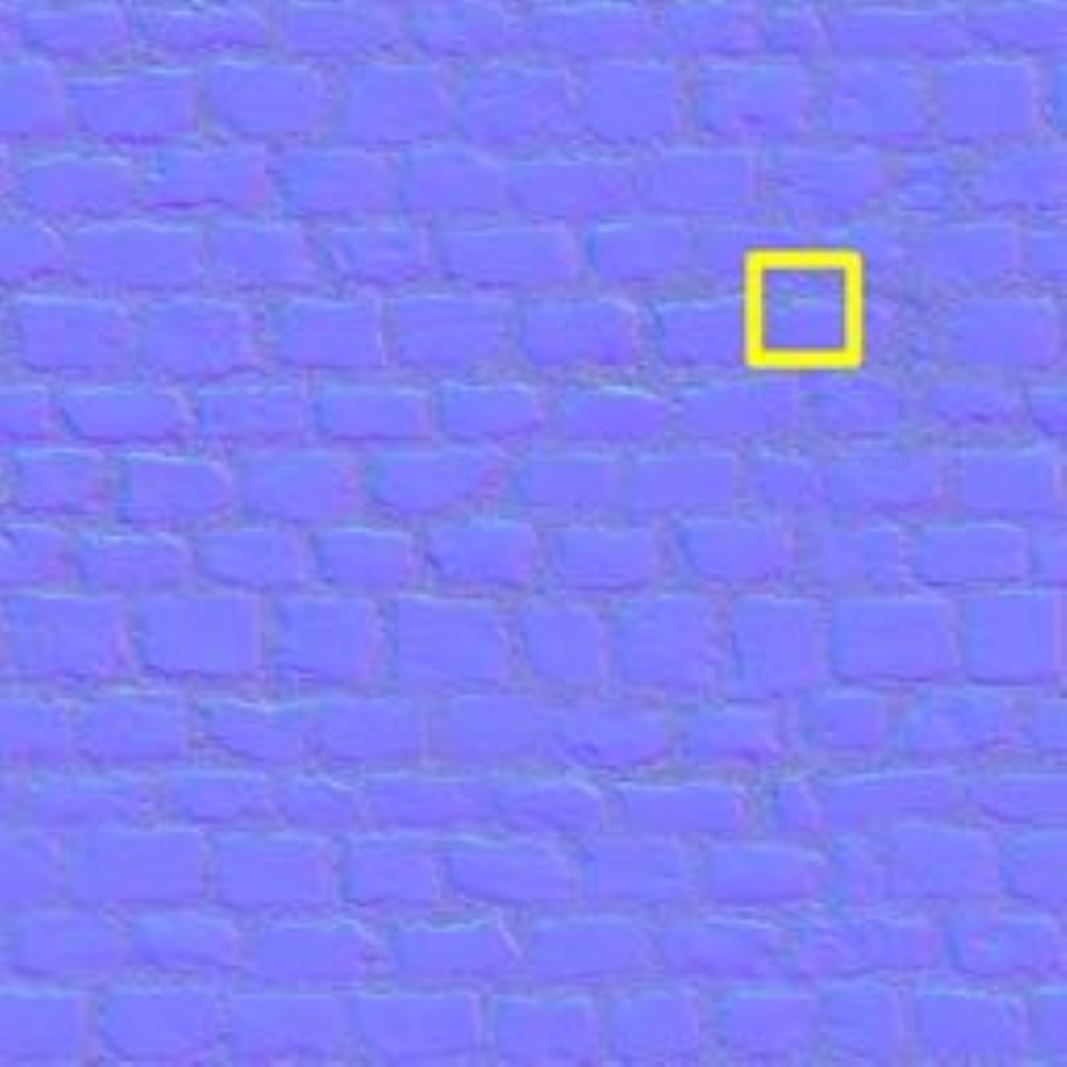} 
       & \includegraphics[width=0.2\textwidth]{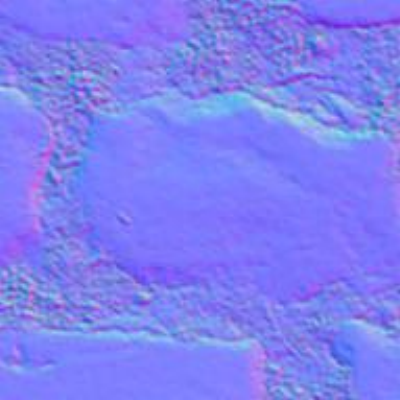}
       & \includegraphics[width=0.2\textwidth]{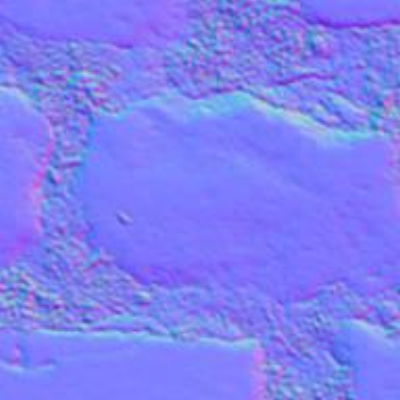} 
       & \includegraphics[width=0.2\textwidth]{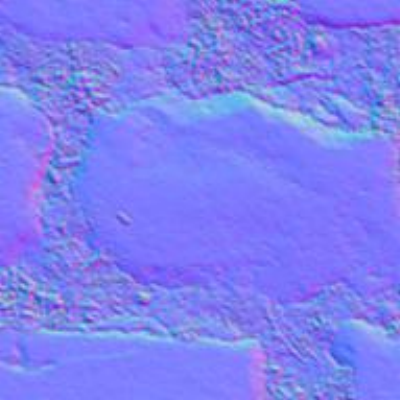} 
       & \includegraphics[width=0.2\textwidth]{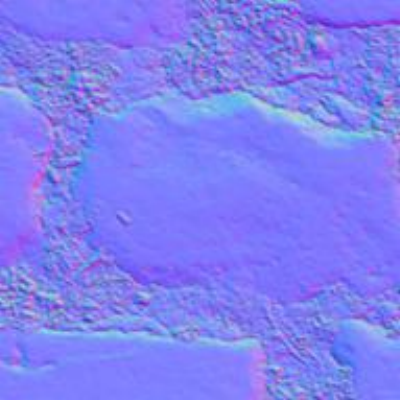}
       & \includegraphics[width=0.2\textwidth]{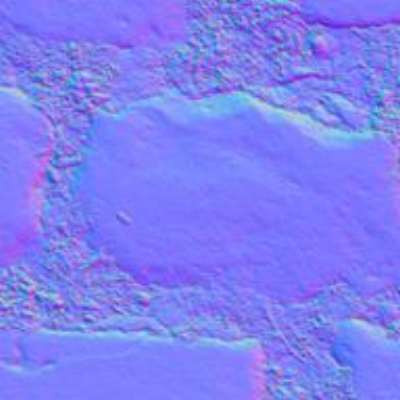}
       & \includegraphics[width=0.2\textwidth]{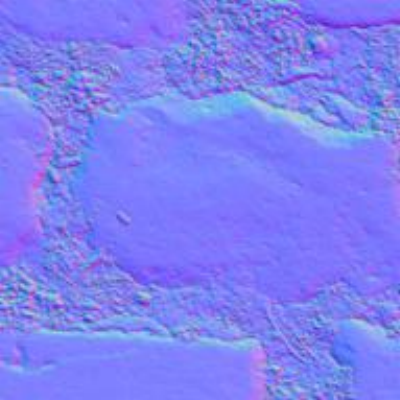}
       & \includegraphics[width=0.2\textwidth]{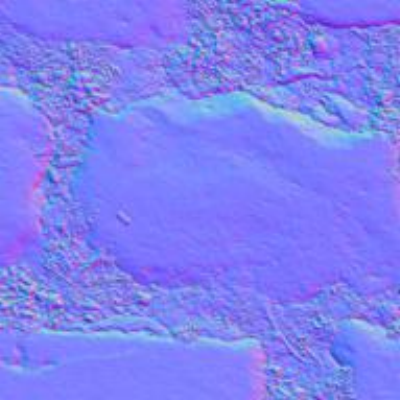}
       \\
        paving stones 131 & $28.52$ ($0.17$) & $\mathbf{28.98}$ ($0.18$) & $30.63$ ($0.38$)& $\mathbf{32.10}$ ($0.39$) & $\mathbf{34.86}$ ($0.76$)&  $34.81$ ($0.73$) & Ground Truth \\ \hline
       \includegraphics[width=0.2\textwidth]{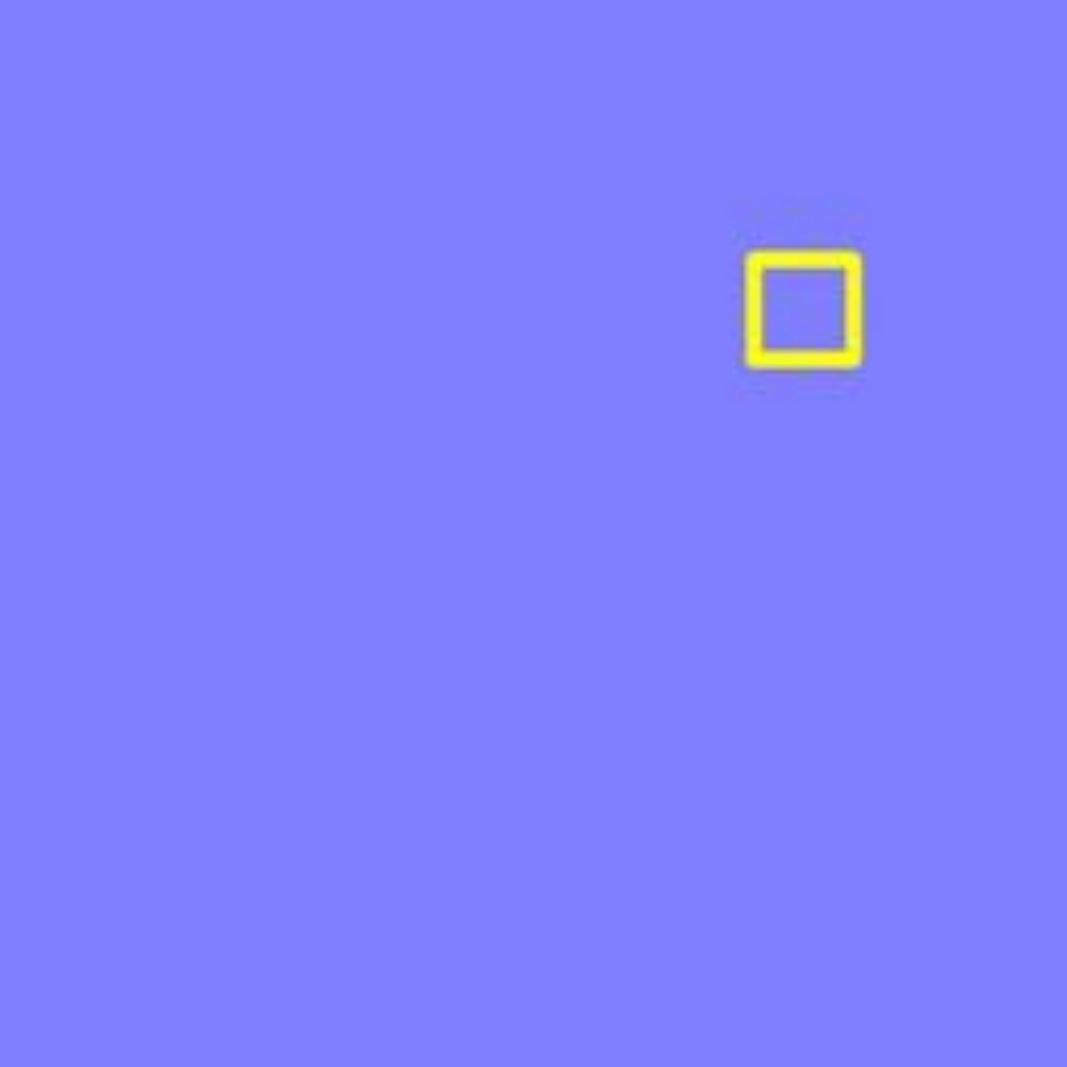} 
       & \includegraphics[width=0.2\textwidth]{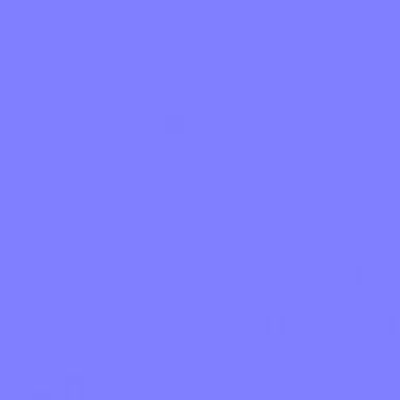}
       & \includegraphics[width=0.2\textwidth]{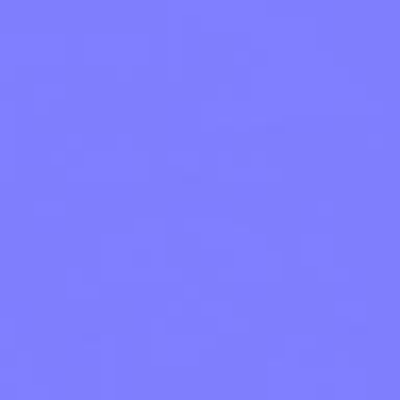} 
       & \includegraphics[width=0.2\textwidth]{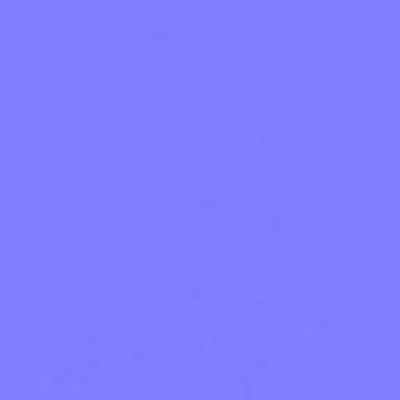} 
       & \includegraphics[width=0.2\textwidth]{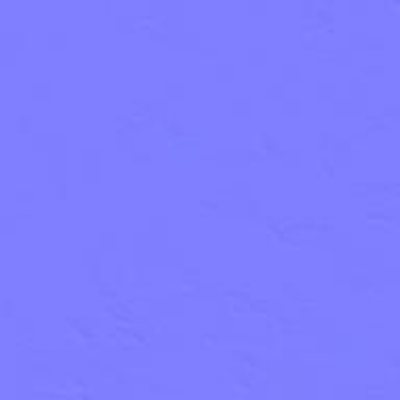}
       & \includegraphics[width=0.2\textwidth]{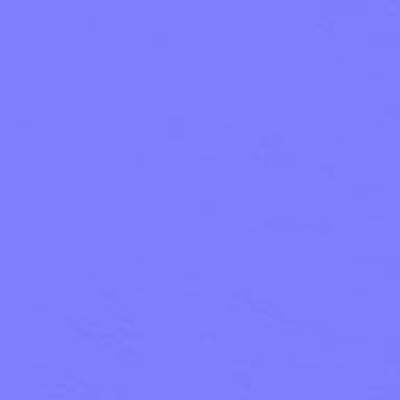}
       & \includegraphics[width=0.2\textwidth]{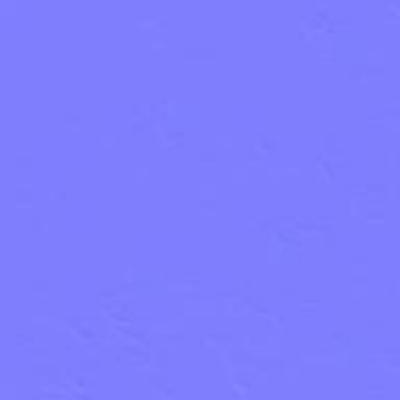}
       & \includegraphics[width=0.2\textwidth]{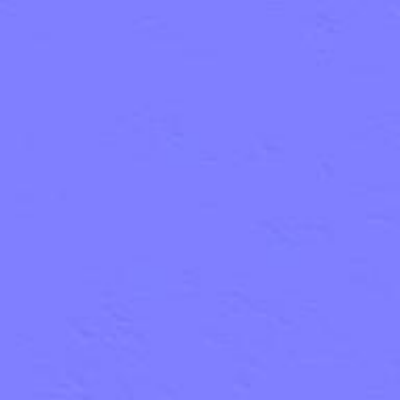}
       \\
        terrazzo 018 & $\mathbf{48.57}$ ($0.18$) & $48.29$ ($0.20$) & $49.50$ ($0.38$)& $\mathbf{52.44}$ ($0.44$) & $52.00$ ($0.76$)&  $\mathbf{52.42}$ ($0.82$) & Ground Truth \\ \hline
       \includegraphics[width=0.2\textwidth]{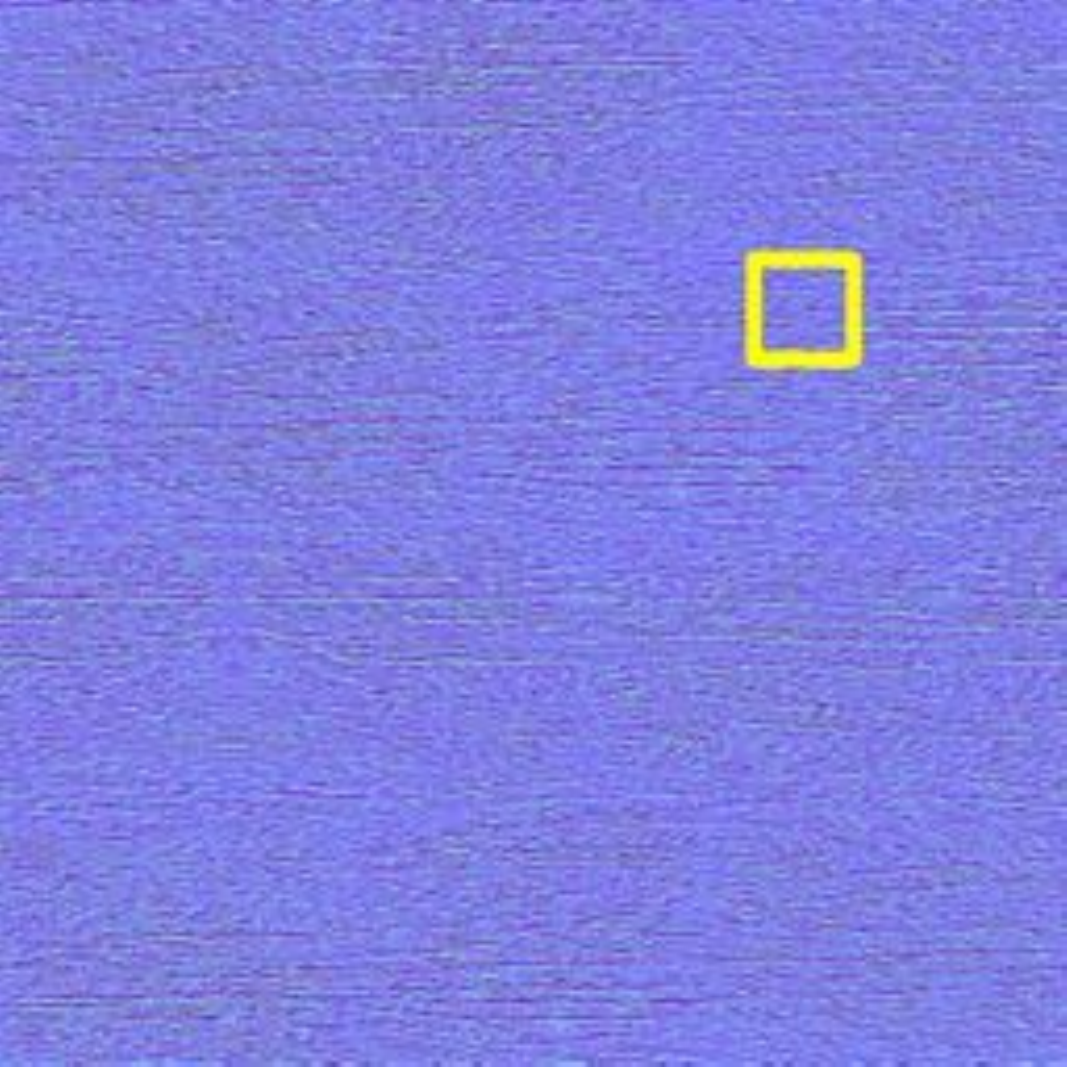} 
       & \includegraphics[width=0.2\textwidth]{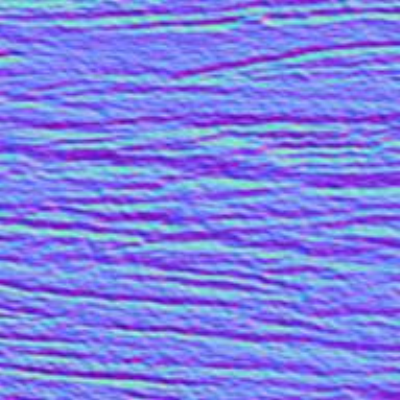}
       & \includegraphics[width=0.2\textwidth]{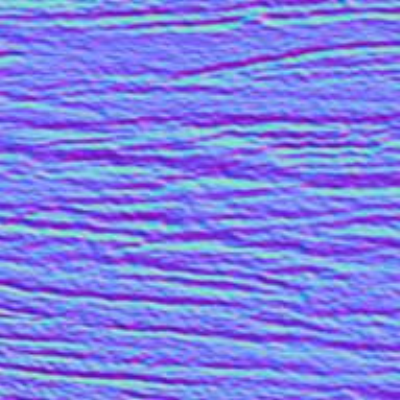} 
       & \includegraphics[width=0.2\textwidth]{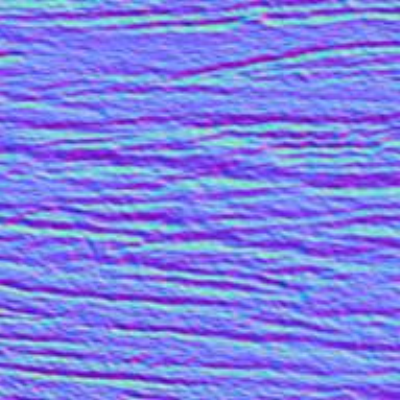} 
       & \includegraphics[width=0.2\textwidth]{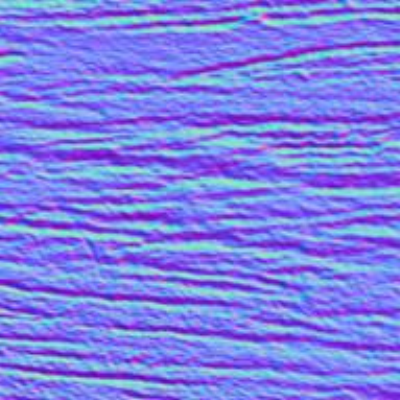}
       & \includegraphics[width=0.2\textwidth]{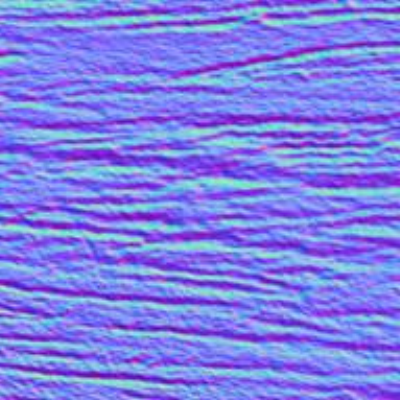}
       & \includegraphics[width=0.2\textwidth]{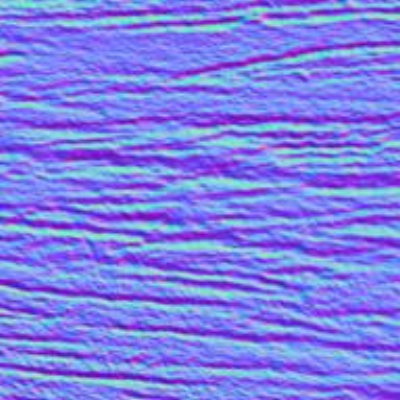}
       & \includegraphics[width=0.2\textwidth]{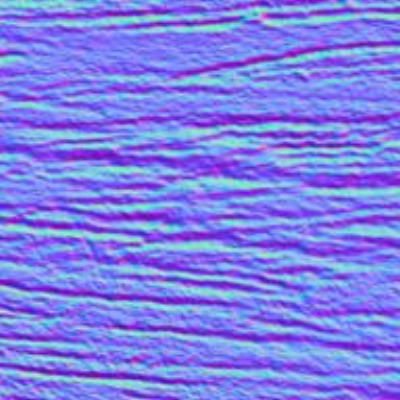}
       \\
        wood 063 & $26.40$ ($0.17$) & $\mathbf{27.83}$ ($0.18$) & $29.63$ ($0.38$)& $\mathbf{31.87}$ ($0.39$) & $33.07$ ($0.76$)&  $\mathbf{35.88}$ ($0.73$) & Ground Truth \\ 
        \hline

     \end{tabular}
     \label{tab:all_textures_normal}
     }
\end{table}

\subsection{Performance through all mip levels}
The performance of a compression technique can vary based on the frequency spectrum of the texture, resulting in different outcomes across mip levels~\cite{Vaidyanathan2023}. 
In Table~\ref{tab:ceramic_roof_psnr} and Table~\ref{tab:paving_stones_psnr}, we demonstrate the PSNR (dB) for the complete range of mip levels ($m=0,\ldots,9$) associated with texture sets ‘Ceramic roof 01’\footnote{retrieved from \url{https://polyhaven.com/}} and ‘Paving stones 131’\footnote{retrieved from \url{https://ambientcg.com/}}, respectively. The tables present the results of all the channels in the texture sets separately, with the second column showing the reconstructed mip level. These results are evaluated based on our method with the lowest BPPC=0.18 (as shown in Figure~\ref{fig:rd_curve_corrected}), referred to as CNTC 16.

\begin{table}[t]
    \centering
    \caption{``Ceramic roof 01'' reconstructed mips using our method with lowest BPPC corresponding to CNTC 16.}
    \resizebox{\textwidth}{!}{
    \begin{tabular}{c|c|c|c|c}
    \hline
          & Reconstruction mips $0,\ldots, 5$&  mip level & resolution & PSNR (dB)\\ \hline\hline
         \multirow{10}{*}{\rotatebox[origin=c]{90}{diffuse}} & \multirow{10}{*}{\includegraphics[height=3.75cm]{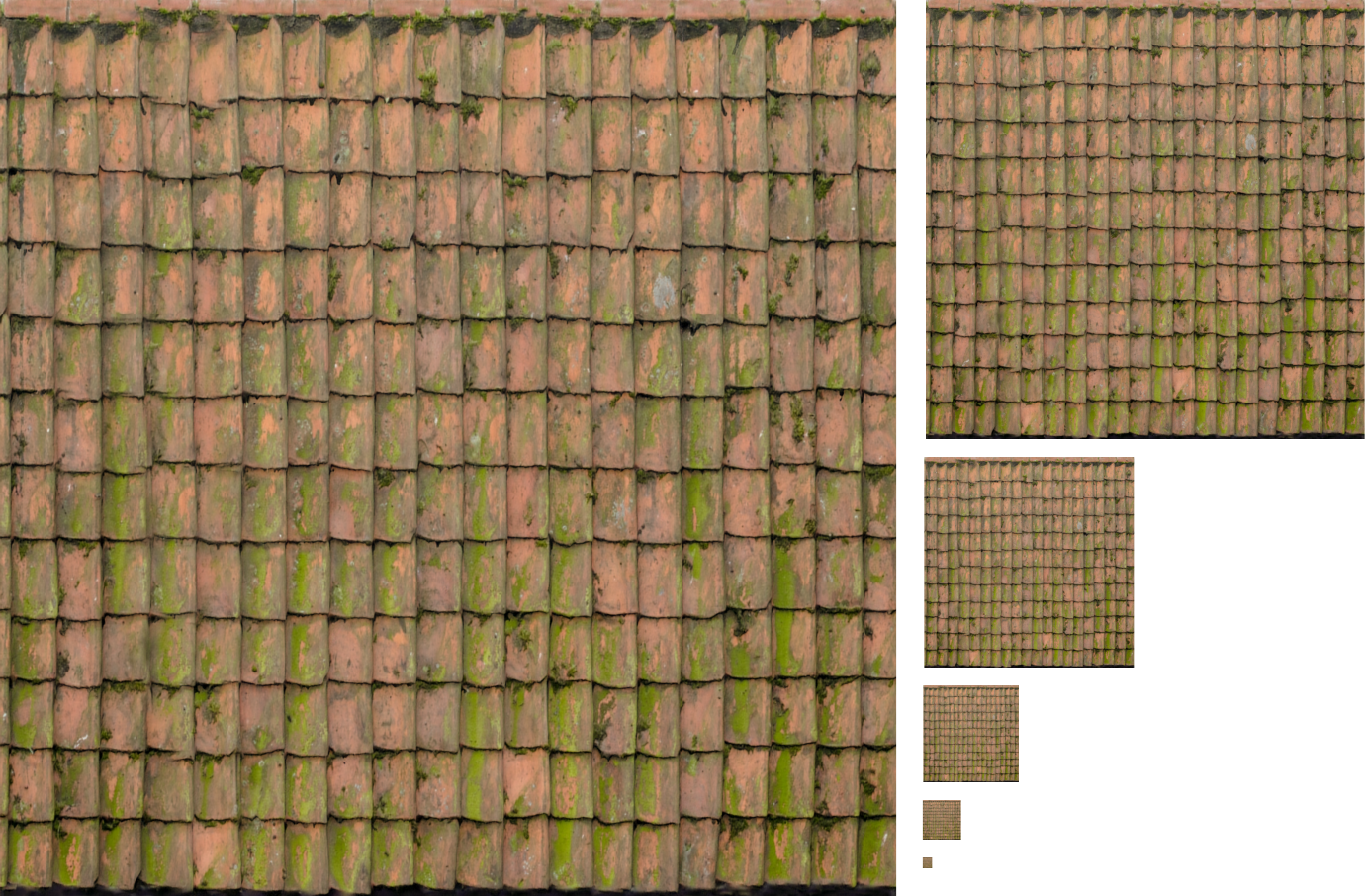}} & 
            0 &  $2048 \times 2048$ & $36.82$\\
            &&1 &  $1024 \times 1024$ & $39.82$\\
            &&2 &  $512 \times 512$ & $39.75$\\
            &&3 &  $256 \times 256$& $35.40$\\
            &&4 &  $128 \times 128$& $34.98$\\
            &&5 &  $64 \times 64$& $36.47$ \\
            &&6 &  $32 \times 32$& $39.06$ \\
            &&7 &  $16 \times 16$& $43.45$\\
            &&8 &  $8 \times 8$& $45.53$\\
            &&9 &  $4 \times 4$& $42.46$\\ 
            \hline
            \multirow{10}{*}{\rotatebox[origin=c]{90}{normal}} & \multirow{10}{*}{\includegraphics[height=3.75cm]{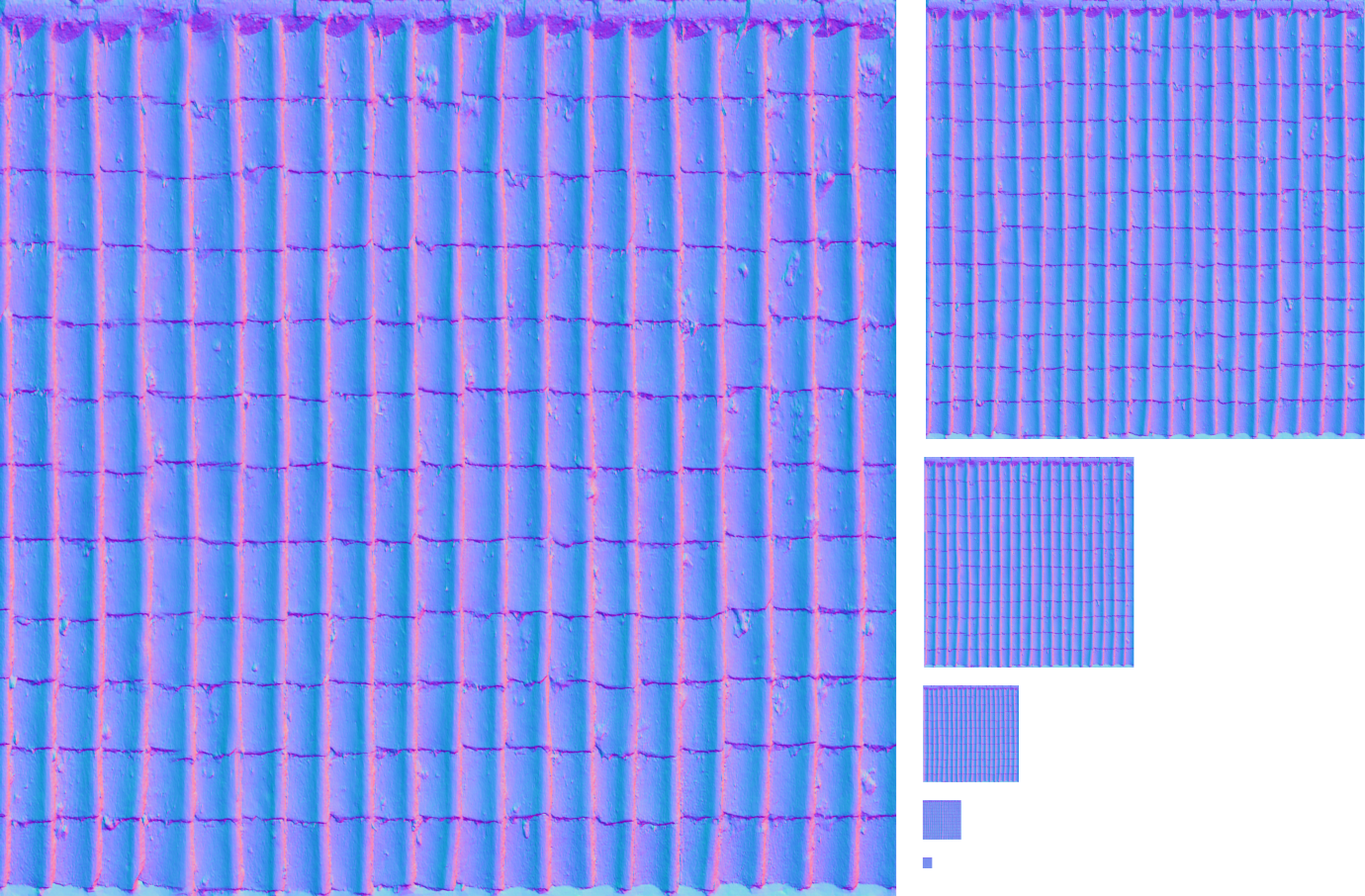}} & 
            0 &  $2048 \times 2048$ & $35.15$\\
            &&1 &  $1024 \times 1024$ & $39.08$\\
            &&2 &  $512 \times 512$ & $39.47$\\
            &&3 &  $256 \times 256$& $36.18$\\
            &&4 &  $128 \times 128$& $34.07$\\
            &&5 &  $64 \times 64$& $36.18$ \\
            &&6 &  $32 \times 32$& $40.16$ \\
            &&7 &  $16 \times 16$& $42.93$\\
            &&8 &  $8 \times 8$& $42.78$\\
            &&9 &  $4 \times 4$& $39.64$\\
            \hline
            \multirow{10}{*}{\rotatebox[origin=c]{90}{displacement}} & \multirow{10}{*}{\includegraphics[height=3.75cm]{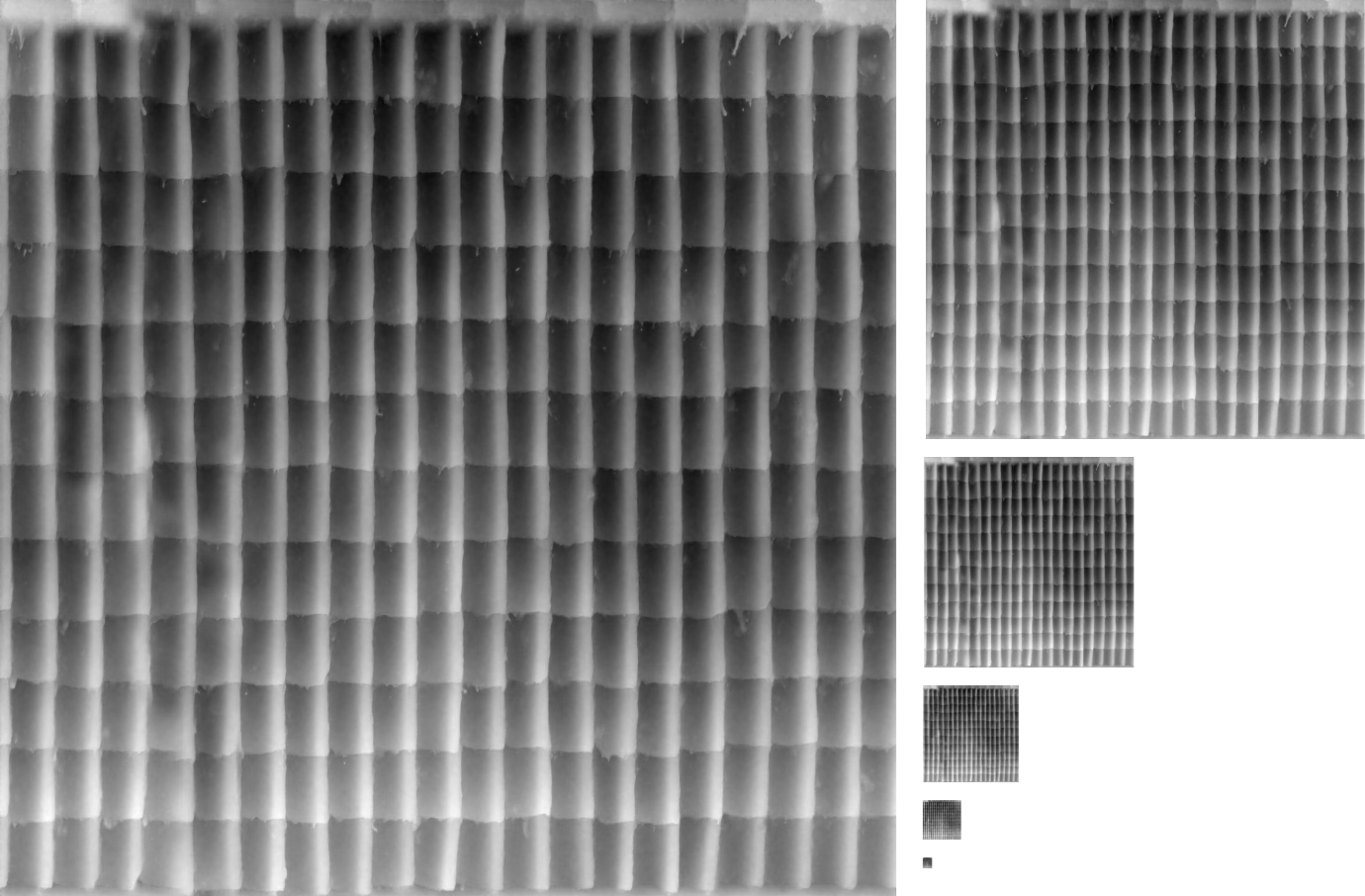}} & 
            0 &  $2048 \times 2048$ & $41.97$\\
            && 1 &  $1024 \times 1024$ & $42.82$\\
            &&2 &  $512 \times 512$ & $42.53$\\
            &&3 &  $256 \times 256$& $39.57$\\
            &&4 &  $128 \times 128$& $34.90$\\
            &&5 &  $64 \times 64$& $35.35$ \\
            &&6 &  $32 \times 32$& $37.40$ \\
            &&7 &  $16 \times 16$& $41.16$\\
            &&8 &  $8 \times 8$& $41.38$\\
            &&9 &  $4 \times 4$& $36.72$\\
            \hline
         \multirow{10}{*}{\rotatebox[origin=c]{90}{ao \& roughness}} & \multirow{10}{*}{\includegraphics[height=3.75cm]{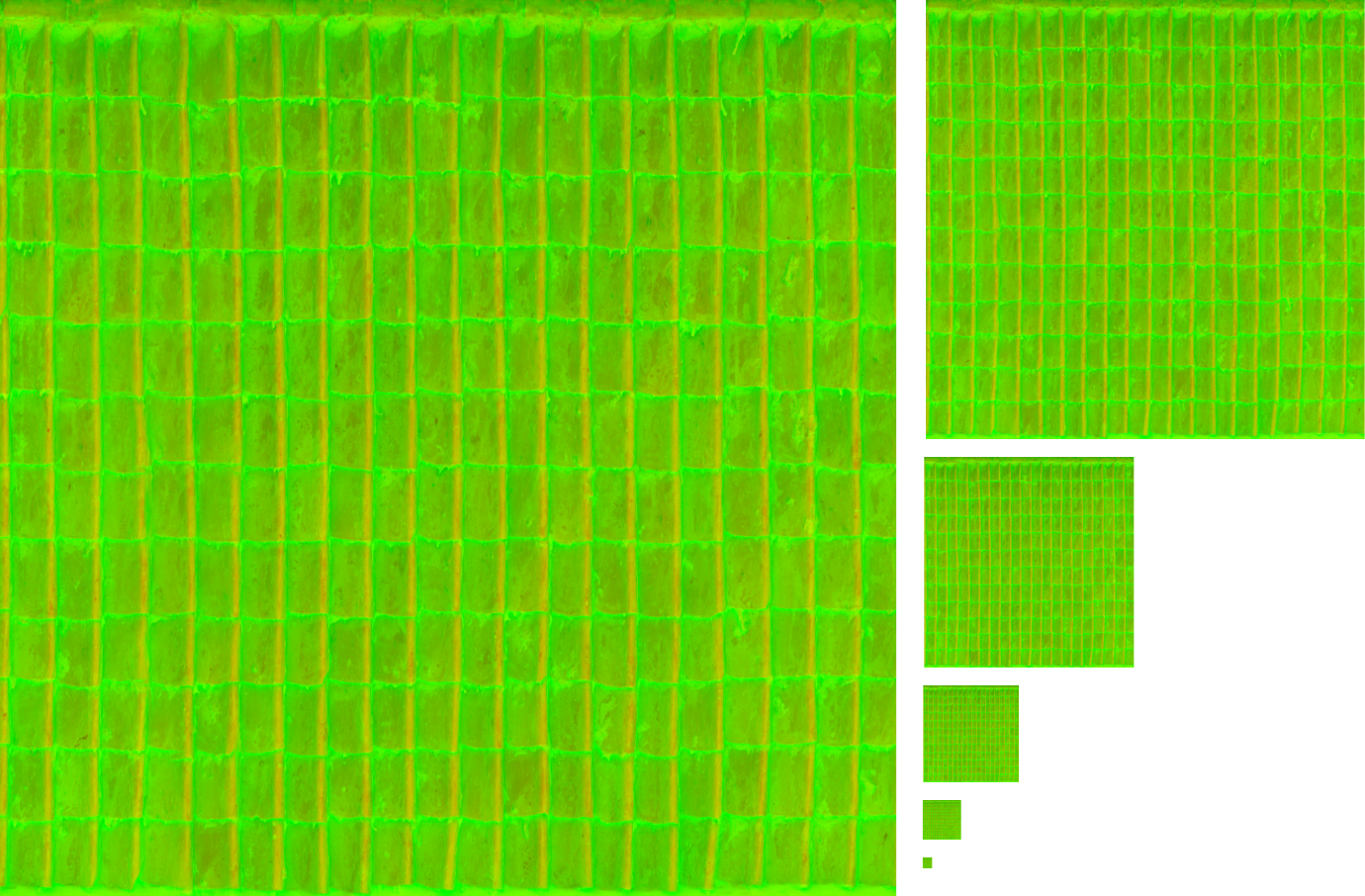}} & 
         0 & $2048\times 2048$ & $37.31$ \\ 
            &&1 &  $1024 \times 1024$ & $40.72$\\
            &&2 &  $512 \times 512$ & $40.99$\\
            &&3 &  $256 \times 256$& $37.51$\\
            &&4 &  $128 \times 128$& $36.18$\\
            &&5 &  $64 \times 64$& $37.89$ \\
            &&6 &  $32 \times 32$& $40.74$ \\
            &&7 &  $16 \times 16$& $45.61$\\
            &&8 & $16 \times 16$& $46.24$\\
            &&9 &  $4 \times 4$& $45.12$\\ 
            \hline
    \end{tabular}
    }
    \label{tab:ceramic_roof_psnr}
\end{table}

\begin{table}[t]
    \centering
    \scriptsize
    \vspace{-0.2cm}
    \caption{``Paving stones 131'' reconstructed mips using our method (CNTC 16).}
    \vspace{-0.2cm}
    \begin{adjustbox}{width=\linewidth}{
    \begin{tabular}{c|c|c|c|c}
    \hline
          & Reconstruction mips $0,\ldots, 5$&  mip level & resolution & PSNR (dB)\\ \hline\hline
         \multirow{10}{*}{\rotatebox[origin=c]{90}{diffuse}} & \multirow{10}{*}{\includegraphics[height=2.75cm]{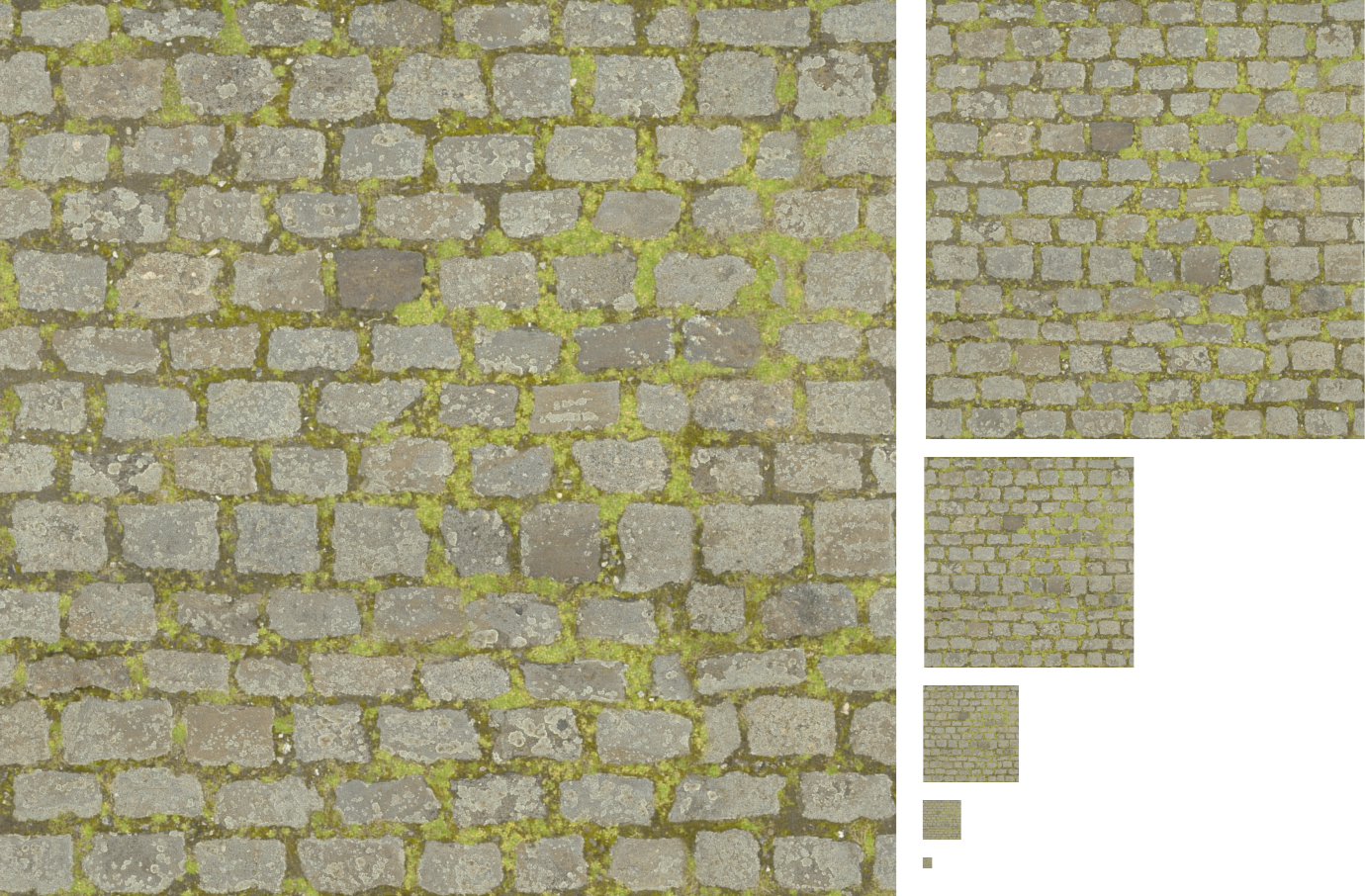}} & 
            0 & $2048 \times  2048$ & 28.54 \\
            && 1 & $1024 \times  1024$ & 34.62 \\
            && 2 & $512 \times  512$ & 36.83 \\
            && 3 & $256 \times  256$ & 33.54 \\
            && 4 & $128 \times  128$ & 32.19 \\
            && 5 & $64 \times  64$ & 32.97 \\
            && 6 & $32 \times  32$ & 36.17 \\
            && 7 & $16 \times  16$ & 41.20 \\
            && 8 & $8 \times  8$ & 47.92 \\
            && 9 & $4 \times  4$ & 51.47 \\
            \hline
            \multirow{10}{*}{\rotatebox[origin=c]{90}{normal}} & \multirow{10}{*}{\includegraphics[height=2.75cm]{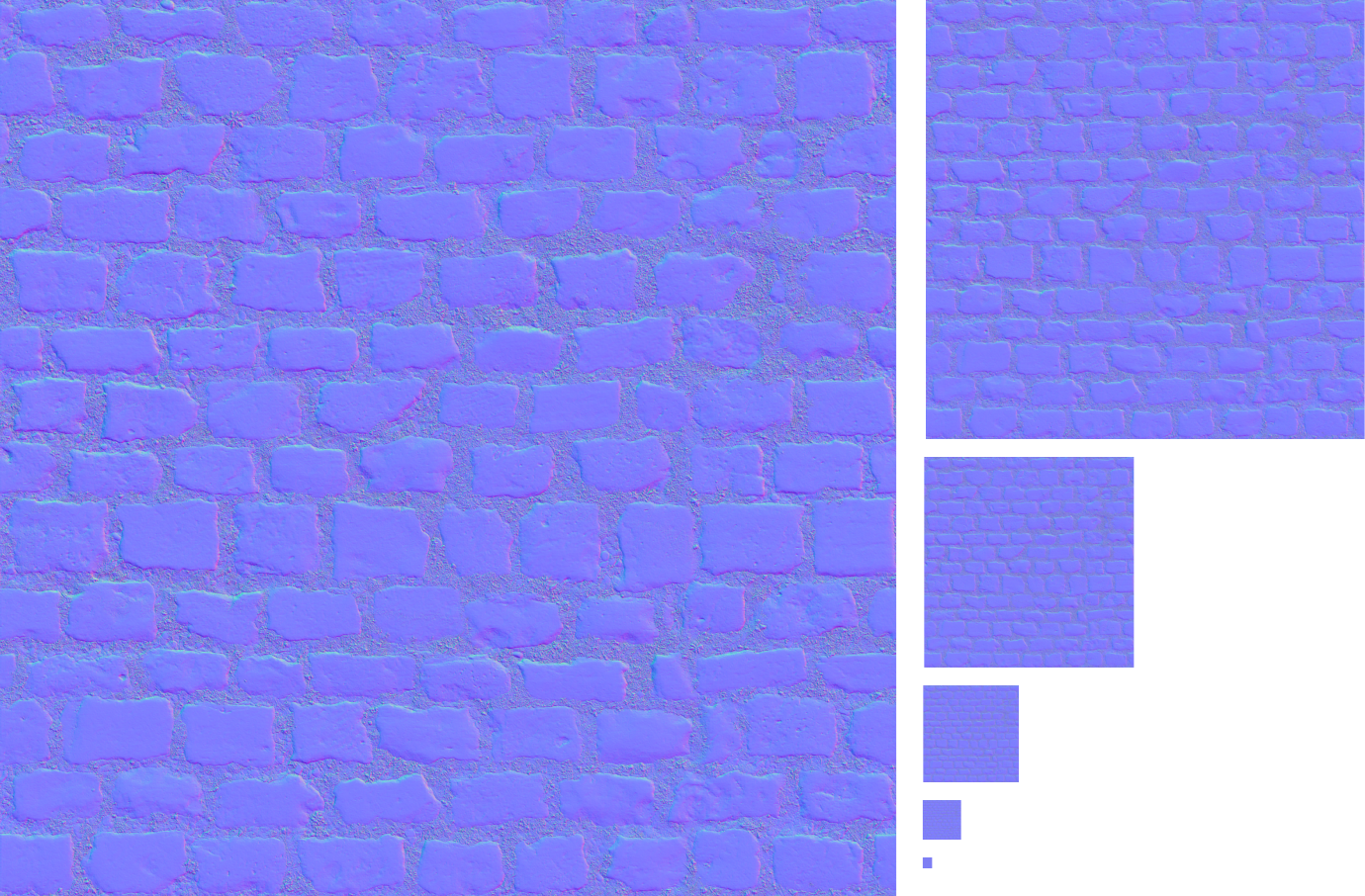}} & 
            0 & $2048 \times  2048$ & 28.98 \\
            && 1 & $1024 \times  1024$ & 35.47 \\
            && 2 & $512 \times  512$ & 39.09 \\
            && 3 & $256 \times  256$ & 37.19 \\
            && 4 & $128 \times  128$ & 37.72 \\
            && 5 & $64 \times  64$ & 38.97 \\
            && 6 & $32 \times  32$ & 41.80 \\
            && 7 & $16 \times  16$ & 46.00 \\
            && 8 & $8 \times  8$ & 51.25 \\
            && 9 & $4 \times  4$ & 55.40 \\
            \hline
            \multirow{10}{*}{\rotatebox[origin=c]{90}{displacement}} & \multirow{10}{*}{\includegraphics[height=2.75cm]{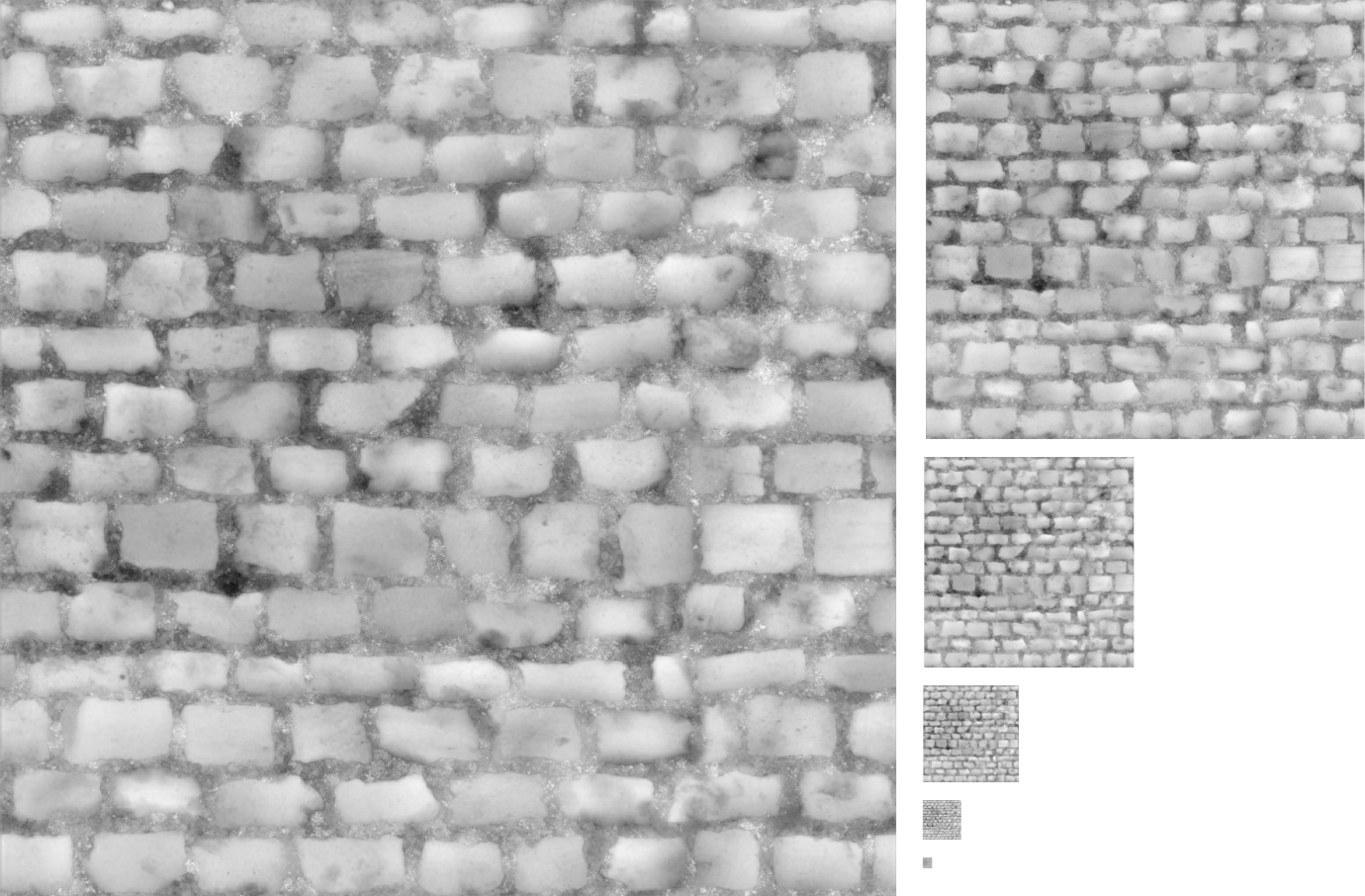}} & 
            0 & $2048 \times  2048$ & 38.19 \\
            && 1 & $1024 \times  1024$ & 39.88 \\
            && 2 & $512 \times  512$ & 40.17 \\
            && 3 & $256 \times  256$ & 36.60 \\
            && 4 & $128 \times  128$ & 32.94 \\
            && 5 & $64 \times  64$ & 32.50 \\
            && 6 & $32 \times  32$ & 33.80 \\
            && 7 & $16 \times  16$ & 37.35 \\
            && 8 & $8 \times  8$ & 43.68 \\
            && 9 & $4 \times  4$ & 43.94 \\
            \hline
         \multirow{10}{*}{\rotatebox[origin=c]{90}{ambient occlusion}} & \multirow{10}{*}{\includegraphics[height=2.75cm]{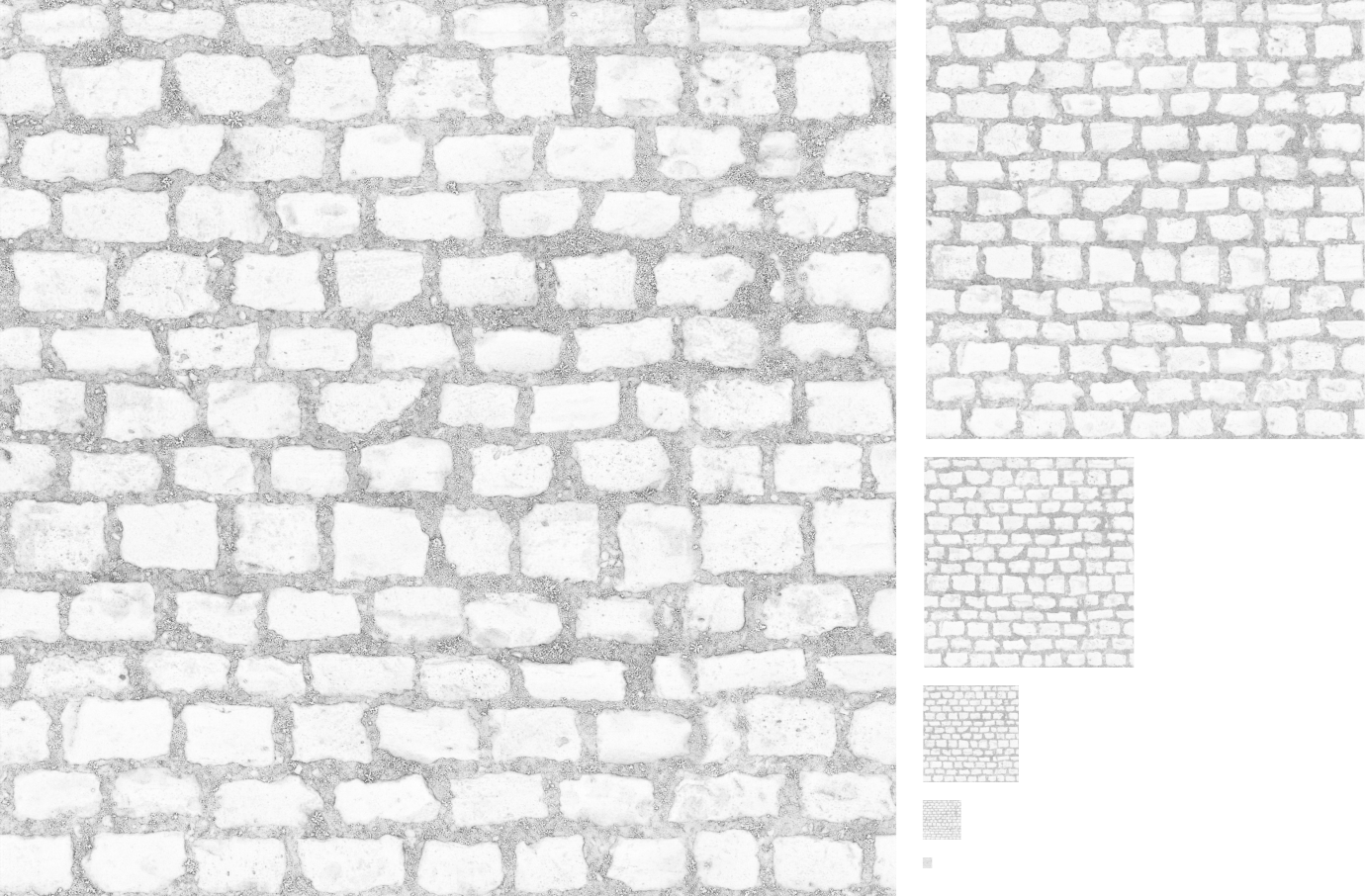}} & 
            0 & $2048 \times  2048$ & 26.68 \\
            && 1 & $1024 \times  1024$ & 34.44 \\
            && 2 & $512 \times  512$ & 37.20 \\
            && 3 & $256 \times  256$ & 35.54 \\
            && 4 & $128 \times  128$ & 33.36 \\
            && 5 & $64 \times  64$ & 33.60 \\
            && 6 & $32 \times  32$ & 35.83 \\
            && 7 & $16 \times  16$ & 40.56 \\
            && 8 & $8 \times  8$ & 46.18 \\
            && 9 & $4 \times  4$ & 51.58 \\
            \hline
             \multirow{10}{*}{\rotatebox[origin=c]{90}{roughness}} & \multirow{10}{*}{\includegraphics[height=2.75cm]{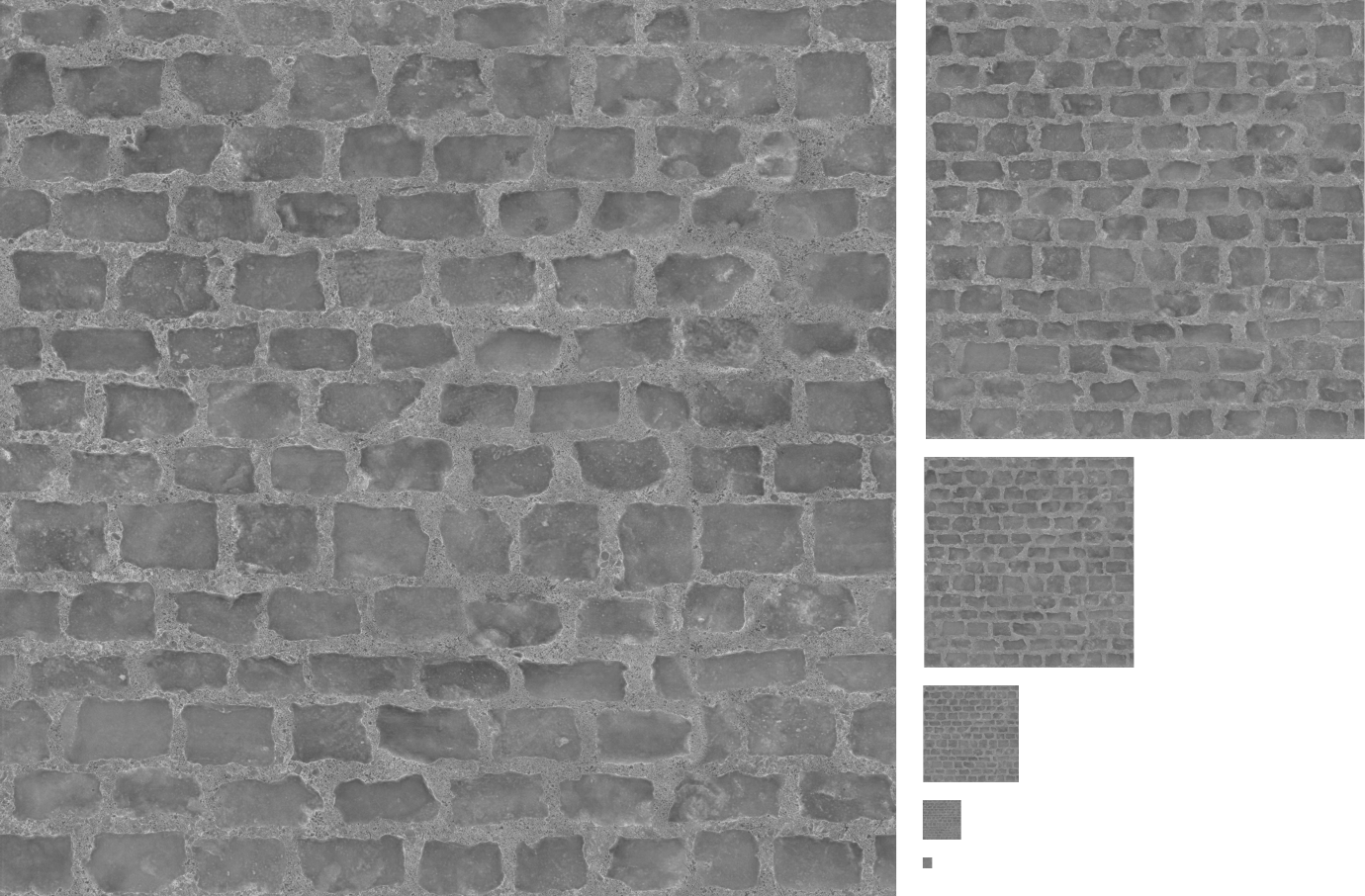}} & 
            0 & $2048 \times  2048$ & 30.64 \\
            && 1 & $1024 \times  1024$ & 36.47 \\
            && 2 & $512 \times  512$ & 39.66 \\
            && 3 & $256 \times  256$ & 37.69 \\
            && 4 & $128 \times  128$ & 36.46 \\
            && 5 & $64 \times  64$ & 36.93 \\
            && 6 & $32 \times  32$ & 38.86 \\
            && 7 & $16 \times  16$ & 40.91 \\
            && 8 & $8 \times  8$ & 45.69 \\
            && 9 & $4 \times  4$ & 51.19 \\
            \hline
    \end{tabular}
    }
        \end{adjustbox}
    \label{tab:paving_stones_psnr}
\end{table}

\end{document}